\pdfoutput=1
\documentclass{article}

\usepackage{PRIMEarxiv}

\usepackage[utf8]{inputenc} 
\usepackage[T1]{fontenc}    
\usepackage{hyperref}       
\usepackage{url}            
\usepackage{booktabs}       
\usepackage{amsfonts}       
\usepackage{nicefrac}       
\usepackage{microtype}      
\usepackage{lipsum}
\usepackage{fancyhdr}       
\usepackage{graphicx}       
\graphicspath{{media/}}     

\usepackage{booktabs}
\usepackage{framed,multirow}
\usepackage{amsmath}
\usepackage{subcaption}
\usepackage{lineno}
\usepackage{amssymb}
\usepackage{hyperref}
\usepackage{latexsym}
\usepackage{url}
\usepackage{xcolor}
\usepackage[utf8]{inputenc}
\usepackage[T1]{fontenc}
\usepackage[british]{babel}
\usepackage[acronym, nonumberlist]{glossaries}
\usepackage{subcaption}
\usepackage{float}
\usepackage{adjustbox}
\usepackage{mathtools}
\usepackage{amsmath}
\usepackage{amssymb}
\usepackage{array}
\newtheorem{definition}{Definition}[section]   
\newcolumntype{L}{>{\centering\arraybackslash}m{5cm}}
\newcolumntype{Y}{>{\centering\arraybackslash}m{3cm}}

\title{Trace Encoding in Process Mining: a survey and benchmarking
}

\author{
  Sylvio Barbon Jr. \\
  University of Trieste \\
  Trieste, Italy\\
  \texttt{sylvio.barbonjunior@units.it} \\
   \And
  Paolo Ceravolo, Rafael S. Oyamada, Gabriel M. Tavares \\
  University of Milan \\
  Milan, Italy \\
  \texttt{\{paolo.ceravolo, rafael.oyamada, gabriel.tavares\}@unimi.it} \\
}

\begin{document}
\maketitle

\begin{abstract}
Encoding methods are employed across several process mining tasks, including predictive process monitoring, anomalous case detection, trace clustering, etc. These methods are usually performed as preprocessing steps and are responsible for transforming complex information into a numerical feature space. Most papers choose existing encoding methods arbitrarily or employ a strategy based on a specific expert knowledge domain. Moreover, existing methods are employed by using their default hyperparameters without evaluating other options. This practice can lead to several drawbacks, such as suboptimal performance and unfair comparisons with the state-of-the-art. Therefore, this work aims at providing a comprehensive survey on event log encoding by comparing 27 methods, from different natures, in terms of expressivity, scalability, correlation, and domain agnosticism. To the best of our knowledge, this is the most comprehensive study so far focusing on trace encoding in process mining. It contributes to maturing awareness about the role of trace encoding in process mining pipelines and sheds light on issues, concerns, and future research directions regarding the use of encoding methods to bridge the gap between machine learning models and process mining. \end{abstract}


\section{Introduction}


Encoding methods are responsible for transforming complex information into a representative feature space. 
In process mining (PM), several tasks (e.g., predictive process monitoring, trace clustering, anomaly detection, etc.) must encode data before feeding specific algorithms.
This step is crucial to account for the goals of a user correctly.
For instance, if a problem demands a solution where interpretability and explainability are needed, the data should be encoded by methods that tend to accomplish those objectives.
On the other hand, if the most essential requirements are space or time complexity, the user should agree to lose part of the previous benefits to match these ones.

In the PM literature, most of the efforts have been dedicated to designing new algorithms and analytical methods but little attention has been given to the impact of encoding methods across the existing tasks.
For instance, in predictive process monitoring \cite{EvermannRF16} used the \textit{word embedding} method to map the cases of an event log into real-valued vectors, whereas \cite{TaxVRD17} used the \emph{one-hot}.
A custom function is adopted by \cite{Hompes2015}, whereas the \emph{count2vec} (occurrence frequencies of activities) is employed by \cite{Appice2016}.
Thus, a researcher interested in comparing the results of these works is in front of a factor she cannot control, as the impact of encoding is not documented and the methods used are different. Moreover, in this work, we emphasize that very few alternative encoding methods have been employed by the community and demonstrate that arbitrarily encoding data might bring suboptimal results and misalignment with the user’s goals. 
We believe that a better understanding of the effect of encoding methods, according to the datasets' characteristics, is decisive in developing more interpretable, explainable, robust, and accurate PM solutions.

Using anomaly detection as a case study, we extend the results of our previous paper~\cite{barbon2020evaluating} by considering several aspects.
First, we increase the number of encoding methods and provide a new taxonomy to classify them according to different dimensions.
Second, we include more datasets, considering more types of anomalies, in order to increase the space of characteristics and achieve a better understanding of how each encoding method behaves according to the data properties.
Third, we employ evaluation criteria that are valuable for PM practitioners and can support the choice of the suitable encoding method according to their goals.
Lastly, we provide a systematic review of encoding methods across popular PM tasks: predictive monitoring, trace clustering, anomaly detection, online process mining, and security and privacy in PM.

More specifically, we first highlight how difficult it is not just to choose a suitable encoding method but also its parameters.
Subsequently, we perform an extensive experimental evaluation of 27 encoding methods with different parameters over 420 synthetic event logs.
We also discuss how current PM literature is limiting their experiments by not considering the impact that encoding methods have in any problem domain.
Thus, we discuss our results and focus the contribution of our work on answering the following research questions:

\begin{enumerate}
    \item How expressive is an encoding method for separating the problems' classes?
     \item What is the demand of time and memory to reach a suitable encoding method? 
     \item Is there any correlation between the encoding method and the performance achieved by algorithms in PM tasks?
     \item How generic encoding methods are, i.e. can an encoding method be applied to any PM task?
\end{enumerate}

We answer these questions by proposing specific evaluation metrics according to different criteria. 
Through an in-depth analysis, we consider the criteria \textit{expressivity}, which aims at capturing patterns across different characteristics of the employed datasets;
\textit{scalability}, which measures the elapsed time and the memory usage of encoding methods;
\textit{correlation power}, which maps the data characteristics to the algorithm performances;
and the \textit{domain agnosticism}, which considers if the encoding method depends or not on the problem domain.
We demonstrate through our extensive experimental evaluation how difficult it might be to choose a suitable encoding method since each of the evaluated metrics has a different best performing method. 
Thus, the main contributions of this work include:

\begin{itemize}
    \item A systematic review of encoding methods in PM and a new taxonomy developed according to such review.
    \item The proposal of new evaluation metrics to measure the quality of encoding methods in PM tasks.
    \item A deep experimental evaluation of several encoding methods never employed before in PM.
    \item A discussion of insights into future research on encoding for PM.
\end{itemize}

We organize the presentation of our work as follows. First, in Section~\ref{sec:problem_definition} we define the problem of choosing the right encoding method and its parameters.
In Section~\ref{sec:background} we provide the necessary background to understand this work. In Section~\ref{sec:encoding_methods} we first present a systematic review of encoding methods in different process mining tasks. Subsequently, we introduce a new taxonomy for encoding event data, organize the encoding methods found by families of algorithms, describe each method, and discuss related works.
Section~\ref{sec:methodology} describes the employed methodology to implement our experimental evaluation and Section~\ref{sec:experiments} presents the carried experiments and results.
In Section~\ref{sec:discussion} we discuss the main insights obtained in this work and provide future directions.
We conclude our discussion in Section~\ref{sec:conclusion}.

\section{Problem Definition}\label{sec:problem_definition}

In this section, we address the problem of how arbitrarily employing encoding methods in PM tasks leads to sub-optimal performance and results in unfair evaluations.
Due to the wide range of encoding methods available nowadays, choosing one given a specific problem is challenging.
This can be seen in the current literature, across different domains, with several automated solutions that have been proposed to decrease human intervention in the design of algorithms and data science pipelines \cite{OlsonM16,KimT18,FeurerKES0H19}.
In PM, we believe this is even more challenging due to the nature of event logs, where events can be described by both numerical and categorical attributes, are aggregated by cases, and are constrained by the control flow of the process. For example, the availability of a given amount of resources may be a precondition to observe an event (e.g., the execution of an activity) with dependencies to other preceding or concurrent events. Condensing all this information into a single encoding method is difficult, and, in practical terms, each method can only capture specific aspects.


Usually, encoding methods for PM are adapted from other domains. 
Simple techniques are often considered, for instance, the \emph{one-hot} encoding scheme~\cite{TaxVRD17} or frequency-based encoding methods~\cite{Francescomarino19}.
To capture the sequential nature of event logs, methods originally proposed in the Natural Language Processing (NLP) community have been employed~\cite{KoninckBW18,TavaresB20}.
However, while we can take into consideration the similarity between the sequential nature of traces and natural language sentences, there are also differences that must be discussed.
For instance, NLP tasks usually handle a very large vocabulary, i.e., a set of unique words or tokens, whereas processes are usually represented by considerably small vocabularies (e.g., the business process activities).
As an attempt of capturing additional complexity, graph neural networks have been recently studied in the literature \cite{VenugopalTFS21}. Convolutional neural networks have also been used for feature extraction~\cite{MehdiyevMLF2022}.
Image-like data engineering methods have been introduced by \cite{Pasquadibisceglie19, SenderovichFM19,VenugopalTFS21}.
More recently, several pipelines have approached domain-specific encoding methods, which we will further describe in Section~\ref{sec:related_works}, that exploit derived features, such as the resource pool discovery algorithm used to encode event resources by~\cite{CamargoDR19}. 

In the context of our work, we stress that adopting the right encoding method and selecting optimal hyperparameters can directly impact the final performance of a given task.
Moreover, evaluating a new algorithm, e.g., a trace clustering algorithm, by comparing it with other solutions but employing different data inputs (i.e., different encoding steps preceding the clustering), produces an unfair evaluation.
\cite{ManeiroVL21} stress this problem, highlighting that a given model cannot be compared with another if their implementations consider different feature spaces.
A brief example illustrating this issue can be found in~\cite{CamargoDR19}, where the authors are approaching the problem of predictive monitoring.
In their evaluation, the authors employ baselines to compare their proposal with existing LSTM architectures, each one based on a different preprocessing procedure.
Regarding other predictive monitoring work, word embedding is employed in \cite{CamargoDR19,Jebrni21,TaymouriRE21} whereas a traditional \emph{one-hot} encoding was used in \cite{PolatoSBL18,MauroAB19,KratschMRS21}, preventing the comparison between these studies. This problem is exacerbated by the fact that the PM community lacks shared benchmarks to be used in algorithm evaluation and comparison. 

\begin{figure}
    \centering
    \includegraphics[width=\linewidth]{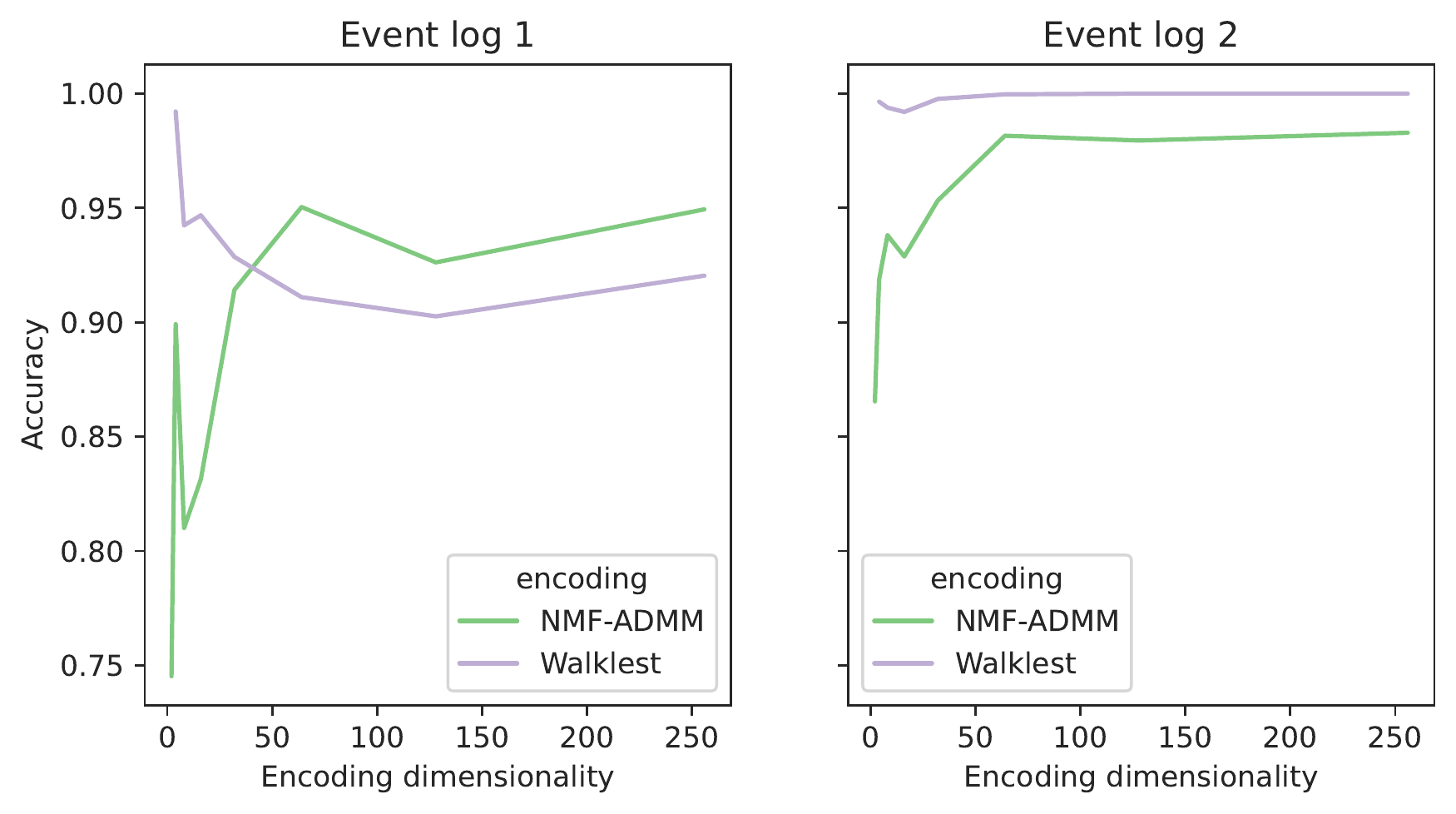}
    \caption{A brief example of performances achieved for two different datasets regarding the anomaly detection problem. For both datasets we fixed two graph-based encoding methods with different parametrizations regarding dimensionality.}
    \label{fig:problem_definition}
\end{figure}

In order to briefly illustrate the impact of arbitrarily encoding an event log, we demonstrate in~\autoref{fig:problem_definition} the following scenario.
We compare two encoding algorithms, vary the parameter of vector dimensionality, apply them to two datasets with different characteristics, and measure the accuracy achieved by a Random Forest classifier regarding the anomaly detection problem\footnote{A detailed description of the material and methods is provided in Section~\ref{sec:experiments}.}.
The datasets have different cardinalities, different types of anomalies, and different rates of anomaly injection. 
The first one has $10k$ traces, $444k$ events, and a 20\% rate of \emph{insertion} anomaly (a random activity is inserted in the trace), whereas the second has $5k$ traces, $221k$ events, and a 15\% rate of \emph{rework} anomaly (an activity is doubled in the trace).
As we can see in the Figure~\ref{fig:problem_definition}, for the first event log (Event log 1), encoding the data employing the \textit{Walklets} method performed better than using the \textit{NMF-ADMM} with low dimensionality and worse for medium and high dimensionality.
In addition, the latter method presented a high accuracy variation for different dimensionalities.
For the second event log (Event log 2), the \textit{Walklets} encoding presented a more stable accuracy, while the \textit{NMF-ADMM} achieved higher accuracy w.r.t. the other dataset but always performed worse than \textit{Walklets}.
This is just a brief example to demonstrate that there is no best encoding method for every dataset or default parameterization to apply. 

Considering the nuances found regarding encoding methods, we address this existing limitation by evaluating encoding methods for PM tasks.
In this work, we attempt to demonstrate in detail how different methods perform on several datasets with distinct characteristics and properties.
We first demonstrate through extensive experimental evaluation that each algorithm has distinct performances across different event logs and pipelines.
We employ several metrics in order to summarize the overall behavior of each method and focus the general evaluation on expressivity, scalability, correlation, and domain agnosticism, which will be further detailed in Section~\ref{sec:methodology}.

\section{Background Notions}\label{sec:background}


PM can be defined as a set of techniques to extract knowledge from event logs \cite{Aalst16}. The goal is to provide analysis that uses event data to extract process-related insights, i.e. creating solutions that are specifically tailored for business processes and their stakeholders.






Thus, let us first consider $\Sigma$ a universe of events, i.e. the set of all possible event identifiers. $\Sigma^{*}$ denotes the set of all sequences over $\Sigma$. 

\begin{definition}[Event, Attribute]\label{def:event}
Events may have various \textit{attributes}, such as {\tt timestamp}, {\tt activity}, {\tt resource}, and others. Let $\mathcal{AN}$ be the set of attribute names. For any event $\textit{e} \in \Sigma$ and an attribute $\textit{n} \in \mathcal{AN}$, then $\#_\textit{n}(\textit{e})$ is the value of attribute \textit{n} for event \textit{e}. Typically, values are restricted to a domain. For example, $\#_\textit{activity} \in A$, where $A$ is the universe of the legal activities of a business process, e.g. $\{a,b,c,d,e\}$.
\end{definition}

With abuse of notation, we refer to the name of the activity of an event $\#_\textit{activity}(\textit{e})$ as the event itself. Thus $\langle a,b,d \rangle$ denotes a trace of three subsequent events. An event can also be denoted by its position in the sequence as $e_{i}$ with $e_{n}$ the last event of this sequence.
A sequence of events composes a \textit{trace} $t \in \Sigma^\ast$ and it can be defined as follows.

\begin{definition}[Trace, Subtrace]\label{def:trace}
In a \textit{trace} each event appears only once and time is non-decreasing, i.e. for $1 \leq \textit{i} < \textit{j} \leq |t|: t(\textit{i}) \ne t(\textit{j})$. 
A trace can also be denoted as a function generating the corresponding event for each position of its sequence: $t(i \rightarrow n) \mapsto \langle e_{i}, ..., e_{n}\rangle$. A subtrace is a sequence $t(i \rightarrow j)$ where $0<i~\leq~j<n$.
\end{definition}

Now let $C$ be the \textit{case universe}, that is, the set of all possible identifiers of a business case execution. $C$ is the domain of an attribute $case \in \mathcal{AN}$.

\begin{definition}[Case, Event Log]\label{def:case}  
We denote a case $c_{i} \in C$ as $\langle a,b,d \rangle_{c_{i}}$, meaning that all events share the same case. For example, for $c_{i}$ we have $\#_{case}(e_{1})$ = $\#_{case}(e_{2})$ =
$\#_{case}(e_{3})$.
An \textit{event log} $L$ is a set of cases $L \subseteq \Sigma^{*}$ where each event appears only once in the log, i.e. for any two different cases the intersection of their events is empty.
\end{definition}

In PM, encoding is a crucial step for several tasks in order to project the information contained in an event log to another feature space before combining it with posterior algorithms such as clustering.
In the context of this work, we approach the anomaly detection problem to benchmark encoding methods.
Thus, let $\mathbf{M}$ be a process model representing the event log and $f$ a test function that indicates if a trace from a log $\mathbf{L}$ is an instance of a model $\mathbf{M}$. Thus, we can define the anomaly detection problem as follows:

\begin{definition}[Anomaly detection]
Let $f: \mathbf{L} \to \{ R, A \}$ be a test function that evaluates whether a trace is regular ($R$) or anomalous ($A$). A trace is considered anomalous if it can not be completely parsed by $\mathbf{M}$. Thus, 
    \begin{equation}
    f(t) =
    \begin{cases}
    Regular,& if~it~can~be~replayed~by~\mathbf{M}\\
    Anomalous,& otherwise
    \end{cases}
    \end{equation}
\end{definition}

Considering the particular granularity of PM data, i.e., traces consisting of events containing numerical, categorical, and time-like values, in this paper, we propose a new taxonomy of methods that handle event data.


\section{Encoding Methods}\label{sec:encoding_methods}

A literature review guided us in proposing a taxonomy of encoding methods, which will be discussed in the following sections. 
To the best of our knowledge, this work is the first in the PM literature to propose a systematic review of encoding methods for PM tasks. 
There are surveys and benchmarks for specific groups of algorithms, for example regarding graph embedding \cite{GoyalF18} or text embedding \cite{RuderVS19}, but they fall outside the scope of PM applications.
In PM, different tasks need to employ an encoding method; we focus our review on trace clustering, predictive monitoring, and anomaly detection tasks. 


\subsection{Systematic Review}


We performed a systematic review by analyzing the methods adopted in the literature.
The online repositories employed are the ACM Digital Library\footnote{https://dl.acm.org/}, the IEEE Xplore\footnote{https://ieeexplore.ieee.org/Xplore/home.jsp}, and the Scopus\footnote{https://www.scopus.com/search}. 
We did not include Google Scholar in order to narrow our search since it usually captures the same papers as the other repositories, plus papers from unknown databases. 
Moreover, we searched only for works from the last 10 years with respect to the date time this review was performed, i.e. from 2012 to 2022. 
A base query was defined and it was partially modified according to each PM task: \emph{"process mining" AND (``encoding'' OR ``encode'') AND $<task>$}, where the keyword \emph{task} might be \emph{``clustering''}, \emph{(``predictive monitoring'' OR ``Process monitoring'')}, \emph{(``anomaly detection'' OR ``conformance-checking'')}, \emph{(``online process mining'' OR ``stream process mining'')}, or \emph{(``security'' AND ``privacy'')}.
Notice that we are including the terms conformance-checking and anomaly detection interchangeably since anomaly detection can be seen as a sub-task of conformance-checking.

\begin{figure}[ht!]
\centering
Publications over the past ten years\par\medskip
\begin{subfigure}{.49\textwidth}
    \centering
    \includegraphics[width=\linewidth]{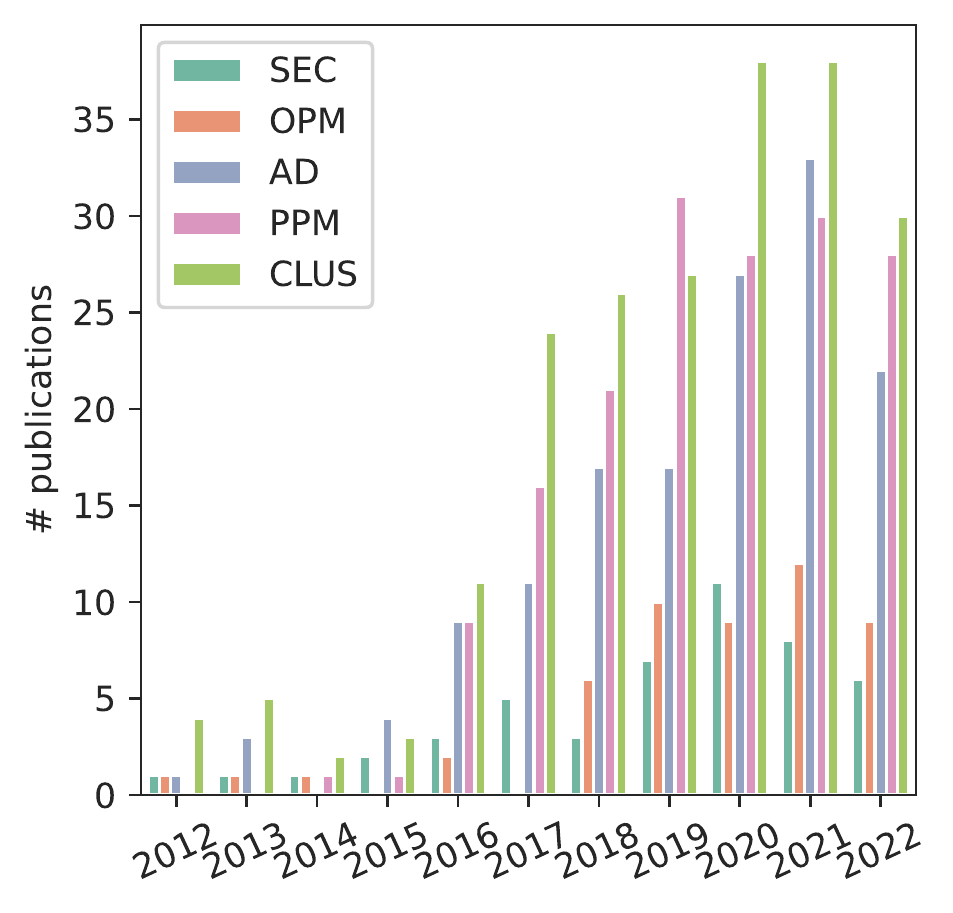}
  \caption{}
\end{subfigure}
\begin{subfigure}{.49\textwidth}
    \centering
    \includegraphics[width=\linewidth]{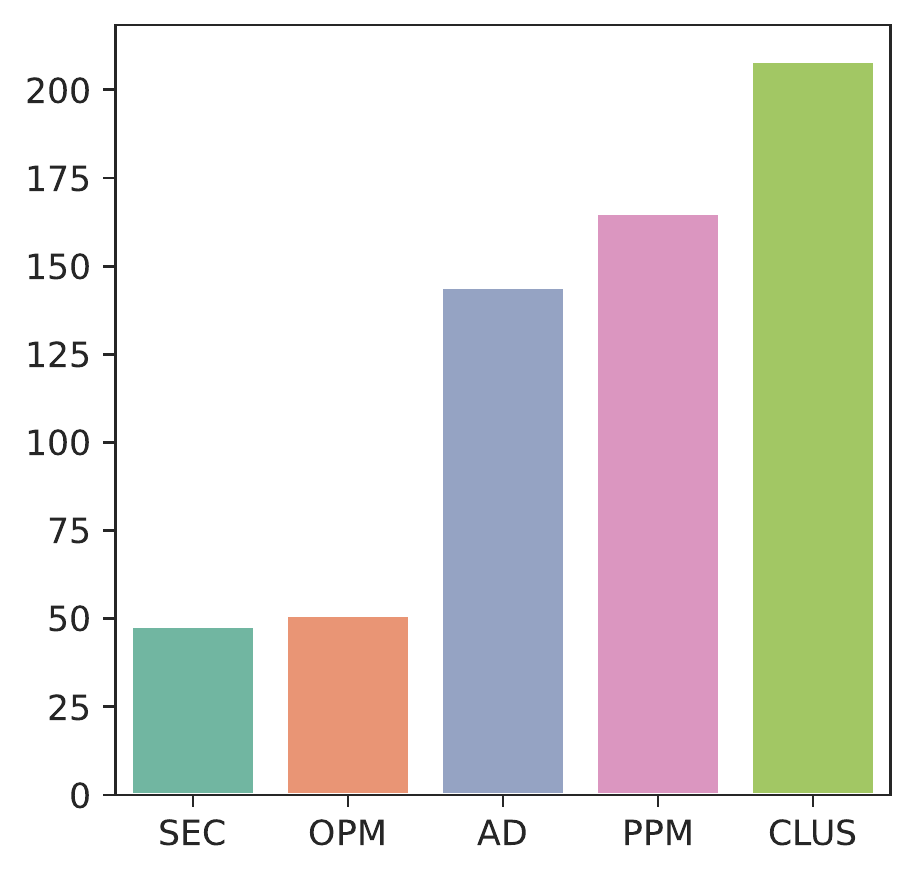}
    \caption{}
\end{subfigure}
\caption{(a) Number of publications each year. (b) The total number of publications over the past ten years.}
\label{fig:publication_timeline}
\end{figure}

After filtering by including only conference and journal papers, and dropping duplicates, we examined the abstracts of each retrieved document to eliminate irrelevant papers. We achieved a total of 616 papers, where 208 are included as clustering (CLUS), 165 as predictive process monitoring (PPM), 144 as anomaly detection (AD), 51 as online process mining (OPM), and 48 as security (SEC).
We illustrate the number of publications for each task and per year in~\autoref{fig:publication_timeline}a and the total publications for each task in~\autoref{fig:publication_timeline}b.
All the retrieved works are public available\footnote{shorturl.at/uwJNW}

\subsection{Taxonomy}    



\begin{figure}[ht!]
\centering
\includegraphics[width=.95\linewidth]{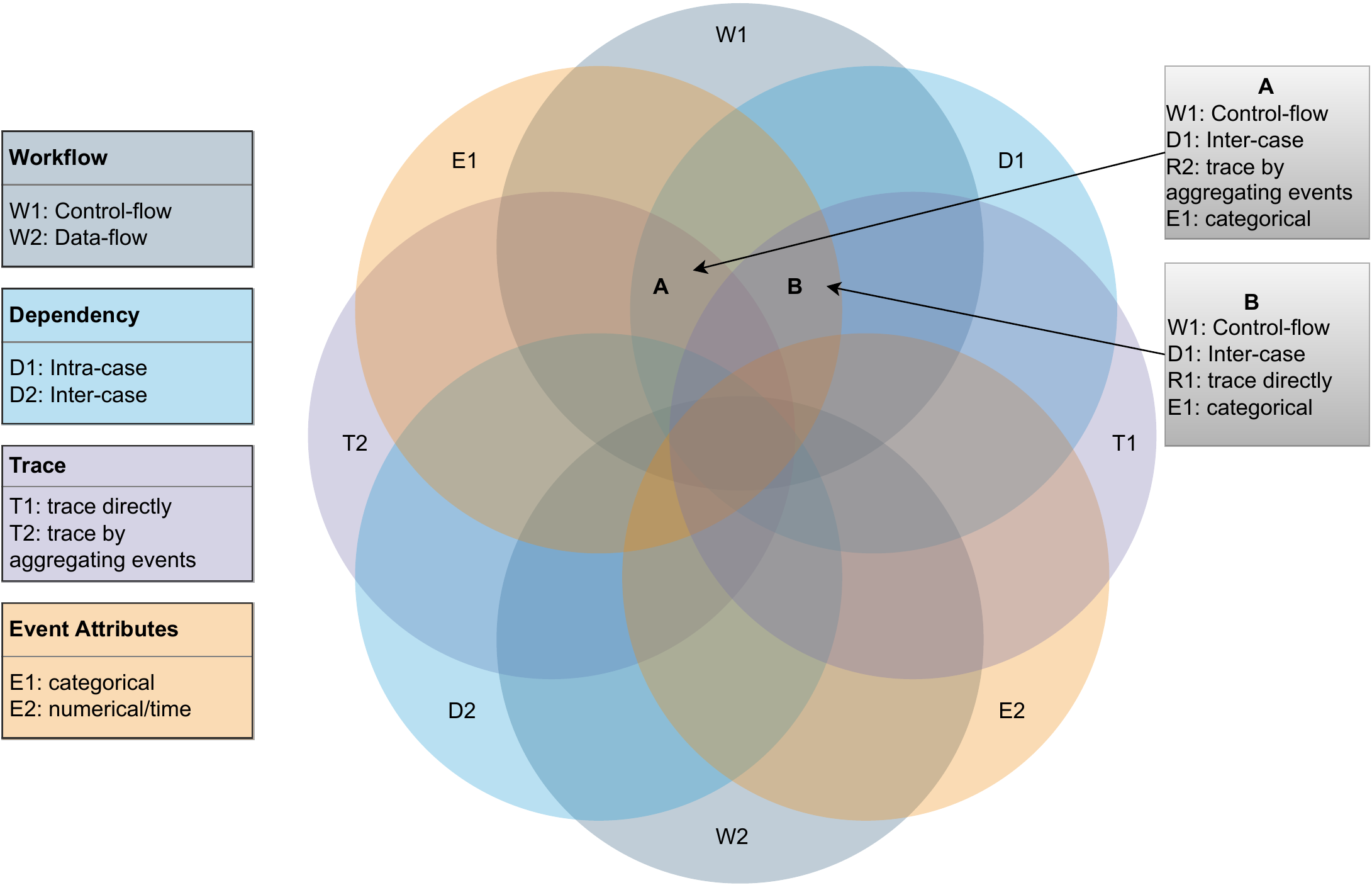}
\caption{General taxonomy for event log encoding. The encoding can be at the event attribute level, where each attribute is encoded independently, or at the trace level, where the event attributes encoded are aggregated to summarize the entire trace.}
\label{fig:encode_taxonomy}
\end{figure}

To guide our discussion, the methods reviewed are organized into a taxonomy, presented in Figure \ref{fig:encode_taxonomy}. 
First, in Figure~\ref{fig:encode_taxonomy}, we illustrate all existing classes on the encoding problem and the intersections among them.
We can think about encoding at the control-flow or data-flow level~\cite{LeoniA13}, where the former considers only the event data whereas the latter analyzes the remaining data from the case \cite{LeoniA13}.
Inter-case and intra-case terminologies have recently been proposed \cite{SenderovichFGJM17} to partially cover this difference by encoding each flow type.
More specifically, the inter-case level aims to capture relationships among cases, e.g. classify case types and similarities between cases. 
One motivation behind this concept is the need to distinguish, for example, two identical sequences of activities (prefixes) with different labels (next activity). 
For intra-case encoding, the focus is to represent past executions by encoding individual activities or completed traces.
Despite being more relevant for real scenarios, the former one is less approached in existing machine learning-based applications since it was recently proposed by \cite{SenderovichFGJM17}.
On the other hand, intra-case encoding is mostly employed across different tasks in PM, and for this reason, we focus our contribution on this encoding level.

Furthermore, at this level, there might be specific targets when encoding data.
For instance, event attributes might need to be encoded individually.
This is a common setting in predictive process monitoring, where each activity is encoded in order to predict the next one.
On the other hand, a specific application (e.g. clustering tasks) might need to encode the complete trace.
In this scenario, the trace can be encoded in a straightforward fashion by the employed algorithm or it can be encoded by aggregating the individual event attributes.

\begin{figure}[ht!]
\centering
\includegraphics[width=.65\linewidth]{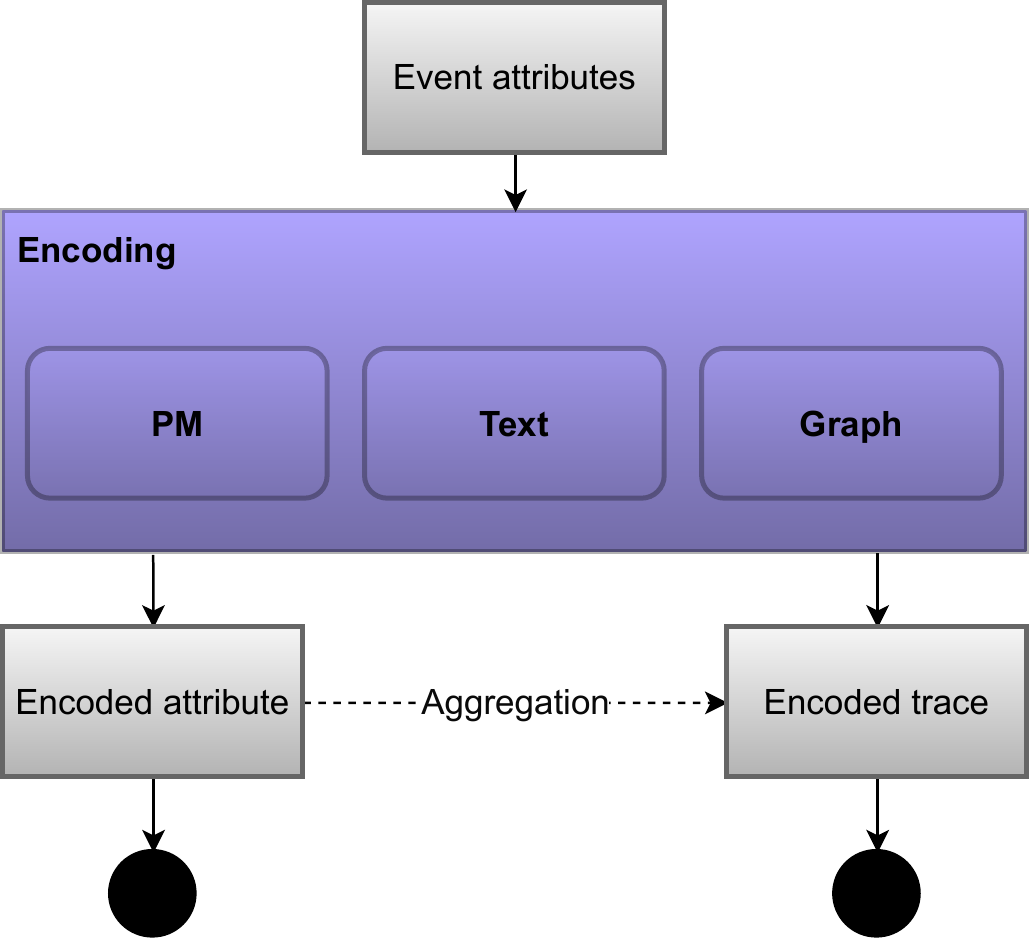}
\caption{Different levels of data encoding. First, event attributes might be individually encoded. Second, the trace can be encoded directly by a certain algorithm or it can be encoded by aggregating the encoded event attributes.}
\label{fig:taxonomy_extension}
\end{figure}

Following our systematic review of encoding methods in process mining, the plethora of alternative encoding methods in the literature, and our proposed diagram of encoding, we are able to wrap the insights from this study and organize the encoding methods into different families.
The taxonomy proposed in Figure~\ref{fig:encode_taxonomy} illustrates the intersections among each possible scenario.
In the figure, we point to only two intersections (\textit{A} and \textit{B}) since those are the most common in the literature, although we expect that future works might fill the other ones. 
Some of the intersections, if not logically impossible are hard to be achieved, for example, a method using both control- and data-flow or intra- and inter-case is of complex design. 
In any case the fact only two intersections are covering the entire set of encoding methods we surveyed is significant of the potential for new methods to experiment in PM. For example, none of the encoding methods we surveyed is exploiting the temporal dimension of cases.
By leveraging our systematic review, we can extend the pointed intersections to group the found encoding methods into families as illustrated in Figure~\ref{fig:taxonomy_extension}.
Thus, these families can be based on or inspired by techniques derived from the following research fields: process mining, text mining, and graph embedding.
Moreover, Figure~\ref{fig:taxonomy_extension} also presents the difference between both intersections, i.e. how the trace encoding should be performed: if it must be encoded directly by a given algorithm or by aggregating previous encoded event attributes.

Taking into consideration the families of encoding methods found in the literature, in the following subsections, we describe each encoding method according to its respective families.
For this work, most of the methods employed have never been used in PM tasks to the best of our knowledge.
Furthermore, to motivate researchers and practitioners to consider these alternative methods more often, we only include methods that have open-source implementations in this study.

\subsection{PM-based Encoding}

Given an event log, we retrieve its respective process model and perform conformance-checking techniques to measure its adherence to the model. Each trace in the event log is evaluated. The results produced are employed as the encoded representation of the trace. The methods considered in our survey are illustrated below.

\textbf{Trace-replay}: given a process model, traces are replayed in it to obtain values that measure its conformance~\cite{berti2019reviving}. More specifically, the values accumulated at each step are the number of tokens correctly consumed ($c$), the number of tokens correctly produced ($p$), the number of missing tokens to execute the event in the next step ($m$), and the number of unconsumed tokens after the last event execution ($r$). Thus, the final measure defined by the trace-replay metric is given by $fitness=\frac{1}{2} (1 - \frac{m}{c}) + \frac{1}{2} (1 - \frac{r}{p})$. All the values produced, $\langle c, p, m, r, fitness \rangle$, are used as the feature vector of a given trace.

\textbf{Trace alignment}: performs a comparison between the process model and a trace and relates the trace to valid execution sequences, i.e., allowed by the model~\cite{CarmonaDSW18}. An alignment is a sequence of moves that can be synchronous, model-dependent, or log-dependent. It is also important to note that more than one alignment between the log and model is possible, and techniques aim at finding the optimal one. The final feature vector is composed of the cost of the alignment, the number of visited states, the number of queued states, the number of traversed arcs, and the fitness value produced. 

\textbf{Log skeleton}: this technique aims at summarizing activity traces by capturing a set of constraints that apply to activities throughout the log~\cite{VerbeekC18}. 
For example, the $R_L^{eq}$ captures the equivalence relation between two activities, which exists if both activities have the same frequency of occurrence in every trace. 
On the other hand, the $C_L^{df}$ counts the number of directly-follows occurrences for every pair of activities. Other examples of measures to capture relations include the always-after and never-together; examples of countermeasures include the sum of occurrences of a given activity in the entire log and the min and max numbers of occurrences of an activity in any trace. In the implementation used for this paper, six different constraints are used.

\textbf{Position profile}: this technique represents an event log through a matrix, where each position refers to the $activity \times position$ regarding all traces~\cite{CeravoloDTB17}. It can be formally defined as a triple $apf = (a, p, f) \in E$, where $a$ is the activity, $p$ is the position of the activity, $f$ is the frequency of occurrence of the given activity, and $E$ is the universe of events.

\subsection{Text-inspired Encoding}

Many solutions used for trace encoding in PM are adapted from methods used in NLP. Exploiting the fact that words in sentences are ordered in sequence and are constrained by dependencies, encoding methods applied to text capture that information. Because traces are composed of sequences of activities the same information appears relevant to characterize them. In particular, in our survey, we consider the following methods.

\textbf{N-grams} \cite{GasparettoMZA22}: this method represents a given sequence of elements through sub-sequences of $n$ items. Thus, considering a sequence $\textbf{s} =\{s_1, ..., s_i\}$, the \emph{n-grams} representation of these sequences is given by $\textit{n-grams}=\{(s_1, ..., s_n),\\ (s_2, ..., s_{n+1}), ... (s_{i-n}, ..., s_{i})\}$.

\textbf{One-hot} \cite{WeissIZ15}: given a variable containing $n$ different values, the variable is transformed into an array where each unique value is represented as a binary vector with the $\textit{i-th}$ position set to one and the rest set to zero. Clearly, the dimension of the vector depends on the size $n$ of the unique values in the vector space, easily reaching high dimensional spaces. 

\textbf{CountVectorizer (count2vec)} \cite{WeissIZ15}: given a collection of categorical documents, this method produces a matrix of token occurrences, where each line in the matrix represents a document and each column a token. The size of the vector space depends on the $n$ unique values in the vector space.

\textbf{HashVectorizer (hash2vec)} \cite{WeissIZ15}: it does the same as \emph{count2vec}. However, instead of storing tokens, it directly maps each token to a column position in the matrix of occurrences. It is mainly useful for large datasets, and unlike \emph{one-hot} and \emph{count2vec}, which have the same dimensionality as the vocabulary length, this method has the flexibility to hash tokens in any dimensionality.

\textbf{TF-IDF} \cite{5392672}: the term frequency (TF) captures the frequency of a particular token w.r.t. to a given document, whereas the inverse document frequency (IDF) measures how common the token is in the corpus. TF can be simply the number of times the token appears and the IDF is calculated as follows: $idf(t, D) = log(\frac{N}{count(d \in D:t \in d)})$, where $t$ is the token and $N$ is the number of documents $d$ in the corpus $D$. Thus, the \emph{TF-IDF} is obtained by multiplying both $\text{TF-IDF}(t, d, D)=tf(t, d) \times idf(t, D)$.

\textbf{Word2vec} \cite{MikolovSCCD13a,MikolovSCCD13b}: the main contribution behind \emph{word2vec} was learning distributed representations of words and reducing the computational cost compared to the state of the art at the time. Although there are two original model architectures for learning the word vectors, Continuous Bag-of-Words (\emph{CBOW}) and Continuous Skip-gram Model (\emph{skip-gram}), the core characteristic of \emph{word2vec} is the removal of the hidden layer of a simple Neural Net Language Model. \emph{CBOW} predicts the current word based on the $t$ words around it, i.e., it predicts $w_t$ given $(w_{t-i}, ..., w_t-1, w_{t+1}, ... w_{t+i})$. On the other hand, given $w_t$, the \emph{skip-gram} predicts the surrounding words $(w_{t-i}, ..., w_t-1, w_{t+1}, ... w_{t+i})$. The parameter $i$ in both cases is a parameter representing a range surrounding the current word $w_t$.

\textbf{Doc2vec} \cite{Le2014}: this algorithm is an extension of \emph{word2vec} and learns the embeddings of documents (sentence, paragraph, essay, etc.). The difference w.r.t. \emph{word2vec} is given by the learning which is performed via the distributed memory and distributed bag of words models and by adding another vector (document ID) to the input.

\textbf{GloVe} \cite{Pennington2014}: this is an unsupervised learning algorithm for obtaining vector representations for words. The main intuition behind this model is the capturing ratios of word-word co-occurrence probabilities in order to capture both local and global dependencies. This is expressed by $F(w_i, w_j, \overset{\text{-}}{w}_k) = \frac{P_{ik}}{P_{jk}}$, where $P_{ik}$ and $P_{jk}$ are the probabilities that the word $k$ appears in the context of words $i$ and $j$ respectively. 

\subsection{Graph-based Encoding}

The intuition behind graph embedding methods is to represent nodes of a graph as low dimensional vectors, where such vectors are representative enough to keep its original relations (edges) intact.
We can formally define the general idea as follows.
A graph can be described as $G=(V, E)$, where $V={\{v_1, ..., v_n\}}$ is a set of vertices (nodes) and $E$ is a set of edges $e=(u,v)$ that connect a pair of vertices $u, v \in V$.
Given a graph $G$, a graph embedding is a mapping function $f: v_i \to y_i \in R^d$, such that $d << |v|$ and $f$ preserves the original structure of their local neighborhood and minimizes the information loss.
In this section, we describe graph embedding methods for event log encoding.


\textbf{DeepWalk} \cite{PerozziAS14}: it can be seen as a two-stage algorithm. First, a discovery of the local structure is performed through random walks. There are two parameters here, the number of random walks $\alpha$ and the number of vertices to visit $t$ for each random walk. Second, similar to the \emph{word2vec}, the \emph{skip-gram} is performed to learn the embeddings. The intuition behind this algorithm is learning embeddings close to each other if they often occur in a similar structural context. 

\textbf{Node2vec} \cite{Grover2016}: this algorithm is similar to \emph{DeepWalk}, where the difference is a biased-random walk that aims at employing a trade-off between breadth-first and depth-first searches. In practice, such balance is capable of providing more informative embeddings than \emph{DeepWalk}.

\textbf{Walklets} \cite{Perozzi2017}: while \emph{DeepWalk} and \emph{node2vec} implicitly capture a certain level of dependencies by generating multiple random walks through $A_{ij}^k$, this algorithm does explicitly by combining factorization approaches with random walks. It preserves dependencies by sub-sampling short random walks on the vertices and by skipping over steps in each random walk. This results in paths of fixed lengths composing sets of pairs of vertices. Thus, these sets are used to learn the latent representations.

\textbf{role2vec} \cite{Ahmed2018}: this is a framework that uses random walks to approximate the pointwise mutual information matrix, which is obtained by multiplying a matrix of structural features with the pooled adjacency power matrix.

\textbf{Laplacian Eigenmaps} \cite{Belkin2001}: this algorithm intuitively keeps the embedding of two nodes close when the weight $W_{ij}$ is high. Given a graph $G$, this algorithm computes eigenvalues and eigenvectors $Ly = \lambda D y$, where $D$ is a diagonal weight matrix $D_{ii} = \sum_j W_{ji}$, and $W$ is the weight matrix. Thus, $L=D-W$ is the Laplacian matrix that can be used to minimize the function $\rho(Y) = \frac{1}{2}\sum{|Y_i - Y_j|^2 W_{ij}} = tr(Y^TLY)$.

\textbf{GraRep} \cite{Cao2015}: this algorithms learns the latent representation $W \in \mathbf{R}^{|V| \times d}$ of the vertices of the weighted graphs. It leverages global structural information to capture long-distance connections. The overall idea is first to calculate the transition probability matrix $A=D^{-1}S$ for each $k$, where $1 <= k <= K$ is the maximum transition step. Subsequently, obtain each $k$-step representation by factorizing the log probability matrix using singular value decomposition. Finally, the $k$-step representations for each vertex on the graph are concatenated and used as latent representations. 

\textbf{Hope} \cite{OuCPZ016}: this embedding algorithm is similar to \emph{GraRep}, but instead of using the transition probability matrix, it employs a similarity matrix $S$. Thus, $S$ can be obtained by using different similarity measures and consequently preserves higher-order dependencies.

\textbf{BoostNE} \cite{Li0GL019}: this algorithm performs a non-negative matrix factorization to calculate the residuals generated by previous embedding models. It assumes the same idea as the gradient boosting method in ensemble learning, where multiple weak learners lead to a better one when aggregated. Given a connectivity matrix obtained through the adjacency matrix of the graph, the algorithm calculates $k$ residual matrices and uses each one as input to the next one using the following equation:

    \begin{equation}
    R_i =
    \begin{cases}
    X,& \text{if } i=1 \\
    max(R_{i-1} - U_{i-1}V_{i-1}, 0),& \text{if } i\geq2
    \end{cases}
    \end{equation}

where $U_i \in R_+^{n \times d_s}$ and $V_i \in R_+^{n \times d_s}$ intuitively act like the embedding representation of the center node and the context node in the $i-th$ level, respectively.
Assuming the defined residual matrix, the embedding representation at the $i-th$ level is obtained by minimizing the loss function $L= \min_{U_i,V_i,\geq 0} || R_i - U_iV_i ||_F^2$, for $1 <= i <= k$.

\textbf{Diff2vec} \cite{RozemberczkiS2020}: the overall idea of this algorithm is sub-sampling diffusion graphs for each node in a graph and generating sequences of vertices through an Euler tour. Given a graph $G$, a graph $G'$ of $l$ vertices is sub-sampled in a diffusion-like random process. Then, from $G'$, sequences of vertices are generated by performing an Euler walk. In this process, $G'$ is first converted to a multi-graph by doubling each edge. Thus, the Euler walk is employed instead of the random walk since this algorithm can capture a more complete view in graphs with this characteristic. The generated sequences of vertices are then used to create the graph embedding.

\textbf{GLEE} \cite{TorresCE20}: unlike most graph embedding algorithms that expect similar nodes to have their embeddings close to each other, this algorithm uses the Laplacian matrix of a given graph to find an embedding with geometric properties. Examples of such properties are dot product (angle), length (area or volume) of a line segment (or polygon), the convex hull of a set of vectors, etc. Thus, given a graph $G$ and its Laplacian matrix $L$, this procedure extracts eigenvectors corresponding to the largest eigenvalues in $L$. These vectors are used as node embeddings.

\textbf{NetMF} \cite{Qiu2018}: this method is built on a theoretical analysis that shows the equivalence of different graph embedding algorithms based on \emph{DeepWalk}. In the original paper, the authors show that methods that use negative samplings, such as \emph{DeepWalk} and \emph{node2vec}, implicitly perform matrix factorization. Thus, the framework \emph{NetMF} is proposed to unify existing methods and perform an explicit factorization.

\textbf{NMF-ADMM} \cite{Sun2014}: given an adjacency matrix, the \emph{NMF-ADMM} algorithm learns the embeddings by using the alternating direction method of multipliers to solve the negative matrix factorization problem.

\textbf{GraphWave} \cite{Donnat2018}: given an undirected graph $G=(V,E)$, an adjacency matrix $A$ (binary or weighted), and a degree matrix $D_{ii}=\sum_j A_{ij}$, this method learns a structural embedding of every vertex $v \in V$. The learning is performed in an unsupervised fashion based on spectral graph wavelets. \emph{GraphWave} is given by $\textit{GraphWave} = U\ Diag(g_s(\lambda_1), ..., g_s(\lambda_n))U^T \alpha_v$, where $\alpha_v$ is the one-hot vector for the vertex $v$, $U$ the decomposition of the eigenvector from $A$, and $g_s$ is a kernel that modulates the eigenspectra.

\textbf{NodeSketch} \cite{Yang2019}: this method recursively generates $k$-order node embeddings in a recursive manner. These embeddings are categorized into low-order $(k = 2)$ and high-order $(k > 2)$. At each step $k$, a Self-Loop-Augmented (SLA) adjacency matrix is generated to obtain the embeddings. Low-order SLA is obtained by simply adding the identity matrix to the original adjacency matrix $M' = M + I$. On the other hand, high-order embeddings first sketch an approximate $k$-order SLA adjacency of the current nodes and merge it with the $(k-1)$-order SLA adjacency matrix in a weighted manner. 

\subsection{Related Works}\label{sec:related_works}

Several tasks in PM, such as predictive monitoring, trace clustering, and anomaly detection, need to encode data to feed algorithms that are applied down the pipeline.
Although transforming event data into a reasonable feature space is a sensible task, i.e., it might drastically impact algorithms' performances, very little attention has been given to encoding methods.
Regarding the literature of other problem domains, there are surveys and benchmarks trying to standardize and better investigate the behaviors of encoding methods according to different problems' characteristics.
For instance, \cite{GoyalF18} surveyed several graph-based embedding methods on different datasets and discussed the main challenges for future research in the field.
Regardless of the approached task (e.g. link prediction, node classification, etc.), the authors demonstrate the difficulty of choosing not only the right algorithm but also the right set of parameters (mainly the dimensionality).
Several trade-offs must always be taken into consideration, for instance increasing the memory usage to achieve more precision or decreasing the dimensionality to decrease the computation time. 
On the other hand, \cite{Goldberg16} covered a wide range of methods to encode textual information.
The work focuses on methods based on encoding methods to feed neural network architectures regarding different tasks and also provides historical notes for each category of task.

The aforementioned works usually focus on representational learning, which employs neural networks to learn a high-quality representation (encoding) of data.
In the natural language literature, the \textit{word2vec} \cite{MikolovSCCD13a,MikolovSCCD13b} can be seen as one of the most important methods for this purpose, which has two architectures variants, one using the \emph{CBOW} algorithm and another one using the \emph{skip-gram} model. 
From this perspective, several methods derived from it, for instance, \cite{Le2014,KoninckBW18}.
The resulting feature vectors representing the original data are also called \textit{embeddings}.

Recently, representational learning has been applied in PM as well. \cite{KoninckBW18} proposes the \textit{act2vec}, \textit{trace2vec}, \textit{log2vec}, and \textit{model2vec}.
Each approach derives from existing encoding methods in the literature and leverages the previous level information to enrich the learning.
That is, the first level is \textit{act2vec}, which extends the \textit{word2vec} architecture to learn the representation of activities.
Subsequently, the \textit{trace2vec} adopts the \textit{doc2vec} concept and jointly learns the representation of activities and traces.
The \textit{log2vec} architecture derives from the same idea as \textit{trace2vec} where the log representation is included in the architecture to be jointly learned.
Finally, for \textit{model2vec}, the authors extend graph representation learning techniques to represent a process model discovered from the event log.
The final architecture also includes all the previous representations to be learned jointly.

In the literature, we also find ``\emph{hand-crafted}'' methods, which are usually developed by following some expertise domain knowledge.
\cite{VenugopalTFS21} proposes using graph convolutional networks for predictive monitoring. In their approach, the authors first perform a feature engineering step to handle time features and then transforms each activity in an event log into a matrix $num\_unique\_activities \times num\_time\_features$. 
\cite{CamargoDR19} employs a PM algorithm to encode resources in event logs.
In a nutshell, the applied algorithm is able to automatically discover resource pools and, hence, reduce the dimensionality of categorical values by grouping them. 
In \cite{Pasquadibisceglie19}, the authors propose the use of convolutional neural networks to perform predictive monitoring.
Thus, they transform the data into an image-like structure in order to be able to train the neural network. 
\cite{ChiorriniDGPP22} presents a method for feature extraction that can be seen as an encoding method, where seven different features are extracted from each activity given a Petri net.
These features aim at capturing local information for the activity with respect to its current case.

Although recent works in predictive process monitoring have explored more alternatives, it is noticeable that works in PM often use a minimal variety of encoding methods.
Most papers use naive techniques like \emph{one-hot} encoding.
Moreover, other encoding approaches are usually combined with results from feature engineer procedures that handle numerical and time-related information.
In recent works, the most common encoding methods for different tasks include the \emph{one-hot} \cite{TaxVRD17,KratschMRS21,Francescomarino17,MauroAB19,TaymouriRE21}, counting the frequencies of categorical data \cite{Francescomarino19}, some type of embedding network \cite{Jebrni21,LinWW19,CamargoDR19,TaymouriRE21}, or hand-crafted representations \cite{CamargoDR19,VenugopalTFS21,Pasquadibisceglie19}.
Therefore, we motivate our work in order to fulfill this limitation by exploring a wider range of encoding methods.

\section{Methodology}\label{sec:methodology}

In this section, we describe the experimental analysis carried out to evaluate encoding methods. We provide details on the software and materials and on the metrics used in order to assess the quality of the surveyed encoding methods. 

\subsection{Implementation Overview}

Our implementation can be organized into three steps: (i) dataset preparation, (ii) encoding generation, and (iii) evaluation of the encoding methods from multiple perspectives. The source code is available online in this repository\footnote{https://github.com/gbrltv/business\_process\_encoding}.

First, we generated synthetic logs using the PLG2 tool~\cite{burattin2015plg2}.
Subsequently, the encoding of the generated logs was performed using open-source libraries in Python as described in Table~\ref{tab:encoding_methods}, which include Sklearn\footnote{https://github.com/scikit-learn/scikit-learn}, Karate Club\footnote{https://github.com/benedekrozemberczki/karateclub}, PM4PY\footnote{https://github.com/pm4py/pm4py-core}, NLTK\footnote{https://github.com/nltk/nltk}, Gensim\footnote{https://github.com/RaRe-Technologies/gensim}, GloVe\footnote{https://github.com/maciejkula/glove-python}, and the \emph{position profile} implementation on github\footnote{https://github.com/gbrltv/meta\_trace\_clustering/blob/main/clustering.py\#L64}.
We organize each method according to the proposed taxonomy and provide the respective references for original papers and online implementations.
Moreover, we set as baselines the methods \emph{count2vec}, \emph{one-hot}, \emph{n-grams}, and \emph{position profile}, which implement the most simple transformations. 
In the case of event-level encoding, the procedure was first performed at the activity level and then the results were aggregated to obtain the trace representation (trace-level encoding).
This aggregation takes the resulting encoded information of each activity and averages it into a unique feature vector.
For graph-based methods, this aggregation was obtained in two different ways: from edges or from nodes.

\begin{table}[!ht]
    \centering
    \begin{tabular}{llll}
        \toprule
        Algorithm &  Year & Family & Implementation \\
        \midrule
        count2vec \cite{WeissIZ15} & - & Baseline &        Sklearn \\
        n-grams \cite{GasparettoMZA22} & - & Baseline & NLTK \\
        position profile \cite{CeravoloDTB17} & 2017 & Baseline & GitHub \\
        one-hot \cite{WeissIZ15} & - & Baseline & Sklearn \\
        GraphWave \cite{Donnat2018} &  2018 & Graph & Karate Club \\
        Laplacian Eigenmaps \cite{Belkin2001} & 2001 & Graph & Karate Club \\
        NMF-ADMM \cite{Sun2014} &  2014 &    Graph &    Karate Club \\
        DeepWalk \cite{PerozziAS14} &  2014 &    Graph &    Karate Club \\
        GraRep \cite{Cao2015} &  2015 &    Graph &    Karate Club \\
        node2vec \cite{Grover2016} &  2016 &    Graph &    Karate Club \\
        Walklets \cite{Perozzi2017} &  2017 &    Graph &    Karate Club \\
        role2vec \cite{Ahmed2018} &  2018 &    Graph &    Karate Club \\
        NetMF \cite{Qiu2018} &  2018 &    Graph &    Karate Club \\
        NodeSketch \cite{Yang2019} &  2019 &    Graph &    Karate Club \\
        BoostNE \cite{Li0GL019} &  2019 &    Graph &    Karate Club \\
        GLEE \cite{TorresCE20} &  2020 &    Graph &    Karate Club \\
        Hope \cite{OuCPZ016} &  2016 &    Graph &   Karate Club \\
        diff2vec \cite{RozemberczkiS2020} & 2018 & Graph & Karate Club \\
        Log skeleton \cite{VerbeekC18} &  2018 & PM & PM4PY \\
        token-replay \cite{Aalst16} & 2016 & PM & PM4PY \\
        alignment \cite{Aalst16} & 2016 & PM & PM4PY \\
        word2vec (skip-gram) \cite{MikolovSCCD13a} &  2013 & Text & Gensim \\
        hash2vec \cite{WeissIZ15} & - & Text & Sklearn \\
        GloVe \cite{Pennington2014} &  2014 & Text & GloVe \\
        doc2vec \cite{Le2014} & 2014 & Text & Gensim \\
        word2vec (CBOW) \cite{MikolovSCCD13b} &  2013 & Text &         Gensim \\
        TF-IDF \cite{5392672} & 1958 & Text & Sklearn \\
    \bottomrule
    \end{tabular}
    \caption{Encoding methods and related details.}
    \label{tab:encoding_methods}
\end{table}

\subsection{Evaluation Metrics}
    
Assuming that encoding methods are used to map the original problem space into a different vector space, we observed the quality of the new space based on several criteria. Moreover, each encoding method has particularities regarding performance delivered, descriptive capability, computational cost, and complexity of hyperparameter space. Thus, to be effective, an encoding method should meet the following criteria:

\begin{itemize}
    \item Expressivity: the relative capacity of an encoding method to affect the complexity of the mapped space regarding the original problem space. An encoding method should be able to map the event logs of varying complexity, in which a straightforward representation regards a simple process and an intricate representation concerns a complex process.
    \item Scalability: the property related to increasing or decreasing the encoding computational cost in response to changes in the event log size. The encoding method should be able to map the event log quickly, without compromising the PM pipeline run time.
    \item Correlation power: the capacity of an encoding method to improve the original problem space. The new feature vector needs to be highly correlated to the PM task goal, i.e., the encoded feature vector should enhance the performance of PM tasks.
    \item Domain agnosticism: refers to how well a given encoding method maps data from different domains. Encoding methods that are non-agnostic can be used only in specific applications.
\end{itemize}

There are different strategies and metrics to assess encoding methods considering the presented criteria. In this work, we exploit the followings. We exploited Principal Component Analysis (PCA)~\cite{daffertshofer2004pca} to verify how well a vector space can be compressed. Classification complexity metrics~\cite{lorena2019complex} to measure how well samples, i.e., encoded traces, are distributed within classes. The F1-score~\cite{sasaki2007truth} to observe the impact of encoding methods on accuracy. Time and space complexity to assess the computational performances. Table \ref{tab:benchmarking} summarizes the contribution of each measure we exploited.

\begin{table}[ht!]
\footnotesize
\centering
\begin{tabular}{@{}lYlL@{}}
\toprule
Criteria                      & Analysis                                                 & Acronym & Description                                                    \\ \midrule
\multirow{5}{*}{Expressivity} & Principal Component Analysis                         & PCA      & Using the 2D projection of a PCA space it is possible to observe
patterns across scenarios from different complexities, ranging from very low, low, average, high and very high expressivity.\\\cline{2-4}                              
                              &Ratio of the PCA dimension to  the original dimension & T4       & This measure is related to the proportion of relevant dimensions that the coded feature vector is composed of. A larger T4 value means more encoded features are needed to describe data variability. \\
 \hline
\multirow{3}{*}{Scalability}  & Encoding Time                                        & Time     & Accumulated time in seconds during the encoding task \\\cline{2-4}
                              & Encoding Memory                                      & Mem      & Accumulated memory in megabytes during the encoding task       \\
 \hline
\multirow{5}{*}{Correlation power}  & Ratio of intra/extra class near neighbor distance     & N2       & This measure is sensitive to how data are distributed within classes and labeling noise in the data. Low values are indicative of simple problems.\\\cline{2-4}
                              & F1-score                                             & F1       & Average of F1-score obtained from the PM classification task, representing the predictive performance delivered by an encoding method. \\
 \hline
Domain agnosticism                 &     General usage of algorithm        &     DA     & This is a binary evaluation (Agnostic or Non-Agnostic) considering agnosticism regarding the PM domain. \\ \bottomrule
\end{tabular}
\caption{List of criteria,  strategies, and metrics to evaluate encoding methods in process mining.}
\label{tab:benchmarking}
\end{table}

\subsection{Experimental Design}


Our experimental design relies on labeled data for ground truth evaluation of the compared encoding methods, as an extension of \cite{barbon2020evaluating}. Synthetic event logs were generated based on standard PM research practices and anomalies were injected into the generated traces, representing an anomaly detection PM task. Afterward, traces were labeled as anomalous or normal, making our data set suitable for supervised learning. Our dataset was made more realistic by adding heterogeneous behaviors to the event logs.

PLG2~\cite{burattin2015plg2} was used to create five different process models by performing a random generation of a process capable of capturing several behaviors, such as sequential, parallel, and iterative control-flow. The rationale of PLG2 is based on the combination of traditional control-flow patterns~\cite{5312cc11e11d466ab9a40bee0fd9d25d}, e.g., sequence, parallel split, and synchronization. In order to simulate real-world scenarios, the patterns are progressively combined according to predetermined rules. Each of the five generated process models defines five different base scenarios based on the activities and gateways included in the scenario.

Creating the log required simulating the process model. For that, we applied the ProM plug-in\footnote{\url{http://www.promtools.org/doku.php}} for the simulation of a stochastic Petri net. We went through 10 thousand simulated cases, with a case arrival rate of about 30 minutes, and kept the default values of the others hyperparameters. We injected anomalies, following~\cite{BEZERRA201333}, by perturbing regular traces as proposed by~\cite{NOLLE2019101458}, as in Table~\ref{tab:anomalies}.

\begin{table}[ht!]
    \centering
    \begin{tabular}{@{}cL@{}}
         \toprule
         Anomaly&Description\\ \toprule
         skip & A sequence of 3 or fewer necessary events are skipped \\ \hline
         insert & 3 or fewer random activities are inserted in the case\\ \hline
         rework & A sequence of 3 or fewer necessary events is executed twice\\ \hline
         early& A sequence of 2 or fewer events executed too early, which is then skipped later in the case\\ \hline
         late& A sequence of 2 or fewer events executed too late\\ \hline
         all & Scenario where the event log is affected by all anomalies listed above\\\hline
    \end{tabular}
    \caption{Anomalies used to simulate the real-life event logs.}
    \label{tab:anomalies}
\end{table}

For each scenario, we injected different percentages of anomalies (5\%, 10\%, 15\%, and 20\%) by replacing normal traces. A total of 420 event logs were generated given five process models, six scenarios of anomalies, and four anomaly percentages using labels and descriptions as additional attributes. Labels regard a normal execution or an anomalous one. The description attribute describes the anomaly and its impact on the case. A general overview is in Table~\ref{table:event_logs}. It is important to note the different scenarios were created with increasing complexity and trace lengths and log sizes (1k, 5k, and 10k cases).

\begin{table}
\scriptsize
\centering
\begin{tabular}{ccccccc}
\hline
\textbf{log}                & \textbf{\#gw}       & \textbf{trace size}    & \textbf{\#acts}      & \textbf{\#cases ($10^3$)} & \textbf{\#evts ($10^3$)} & \textbf{\#vars ($10^3$)} \\ \hline
\multirow{3}{*}{scenario 1} & \multirow{3}{*}{8}  & \multirow{3}{*}{9-13}  & \multirow{3}{*}{22}  & 1                         & $75 \pm 1$               & $1 \pm 0$                \\
                            &                     &                        &                      & 5                         & $378 \pm 5$              & $2 \pm 0$                \\
                            &                     &                        &                      & 10                        & $755 \pm 9$              & $2 \pm 1$                \\
\multirow{3}{*}{scenario 2} & \multirow{3}{*}{12} & \multirow{3}{*}{26-30} & \multirow{3}{*}{41}  & 1                         & $186 \pm 1$              & $1 \pm 0$                \\
                            &                     &                        &                      & 5                         & $929 \pm 5$              & $5 \pm 0$                \\
                            &                     &                        &                      & 10                        & $1857 \pm 10$            & $10 \pm 0$               \\
\multirow{3}{*}{scenario 3} & \multirow{3}{*}{22} & \multirow{3}{*}{42-50} & \multirow{3}{*}{64}  & 1                         & $308 \pm 1$              & $1 \pm 0$                \\
                            &                     &                        &                      & 5                         & $1538 \pm 5$             & $5 \pm 0$                \\
                            &                     &                        &                      & 10                        & $3077 \pm 10$            & $10 \pm 0$               \\
\multirow{3}{*}{scenario 4} & \multirow{3}{*}{30} & \multirow{3}{*}{3-30}  & \multirow{3}{*}{83}  & 1                         & $89 \pm 3$               & $0 \pm 0$                \\
                            &                     &                        &                      & 5                         & $439 \pm 6$              & $2 \pm 0$                \\
                            &                     &                        &                      & 10                        & $879 \pm 12$             & $4 \pm 0$                \\
\multirow{3}{*}{scenario 5} & \multirow{3}{*}{34} & \multirow{3}{*}{4-37}  & \multirow{3}{*}{103} & 1                         & $133 \pm 3$              & $1 \pm 0$                \\
                            &                     &                        &                      & 5                         & $659 \pm 7$              & $3 \pm 0$                \\
                            &                     &                        &                      & 10                        & $1318 \pm 13$            & $6 \pm 0$                \\ \hline
\end{tabular}
\caption{Overview of five process models. For each scenario, we generated event logs by combining three different cardinalities, injecting seven different anomalies, at four different rates of injection. This resulted in 84 event logs for each scenario and 420 event logs in total. \textit{gw}, \textit{acts}, \textit{evts}, and \textit{vars} stand for the number of gateways, activities, events, and variants, respectively.}
\label{table:event_logs}
\end{table}

\section{Benchmarking Process Mining Encoding}\label{sec:experiments}

In this section, we report on the results achieved in our experiments for each family of encoding methods (Baseline, Graph, Text, and Process Mining).
    
\subsection{Expressivity}

In this work, the expressivity of encoding methods is based on PCA and T4 analysis. PCA models were calculated using the encoded vector of all event logs for each encoding method, using a 2D sub-space projection of the first and the second principal component to identify how complex is the mapped space. Since the original problem (i.e., the event logs) is the same, differences in the distribution and density of the encoded events can lead to interpretations about the mapping quality offered by each encoding method. In the PCA, each point represents a feature vector, and each color represents a scenario. The depiction highlights the encoding capacity of generating feature vectors preserving inter- and intra-traces similarities. This is demonstrated by the co-location of samples, i.e., encoded traces, of similar scenarios, i.e., the same color points near to same color points in each 2D projection. A high level of expressivity relies on non-overlapped clusters of samples from the same scenario with clusters sorted by scenarios' complexity occupying the whole sub-space projected. A low level of expressivity is associated with occluded samples with mixed sparse distributions or dense overlapped allocation. An average expressivity is identified when the projected distribution matches partially high and low characteristics. Very high and very low expressivity are obtained when an encoding method completely matches the mentioned characteristics, positively or negatively.

\begin{figure}[ht!]
    \centering
    \begin{tabular}{cccc}
    \multicolumn{4}{c}{
      \includegraphics[clip, trim=0cm 9cm 0cm 0cm, width = 3.0in]{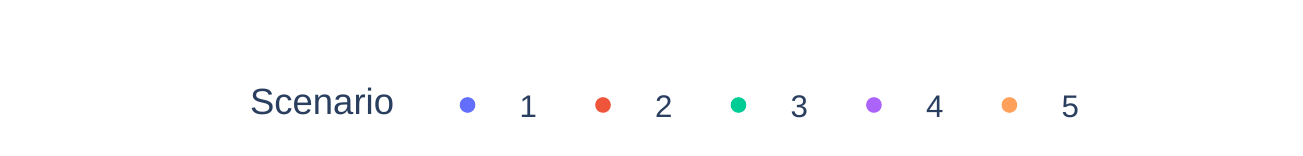}
      }
      
    \\
    \subfloat[count2vec]{\includegraphics[clip, trim=1.5cm 1.5cm 2cm 1cm, width = 1.0in]{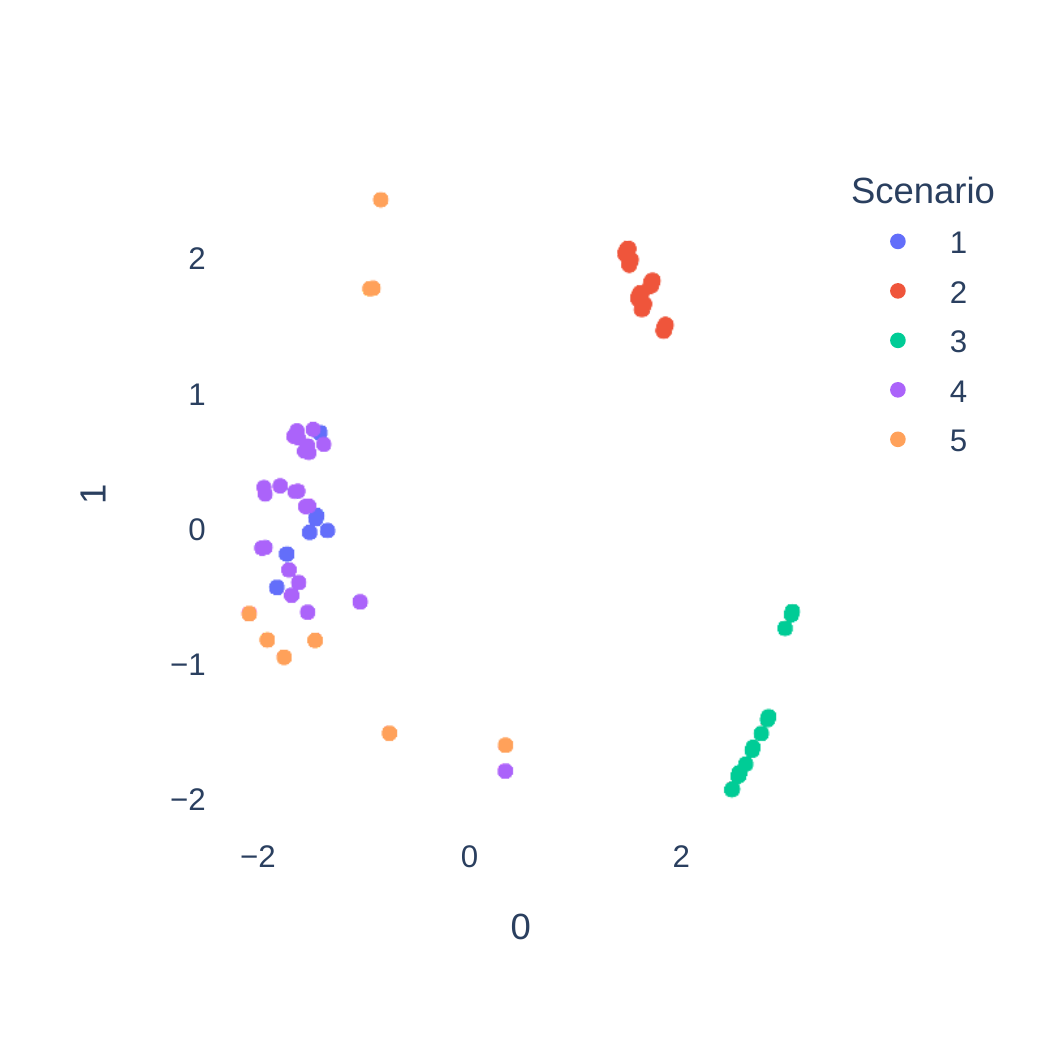}} &
    \subfloat[one-hot]{\includegraphics[clip, trim=1.5cm 1.5cm 2cm 1cm, width = 1.0in]{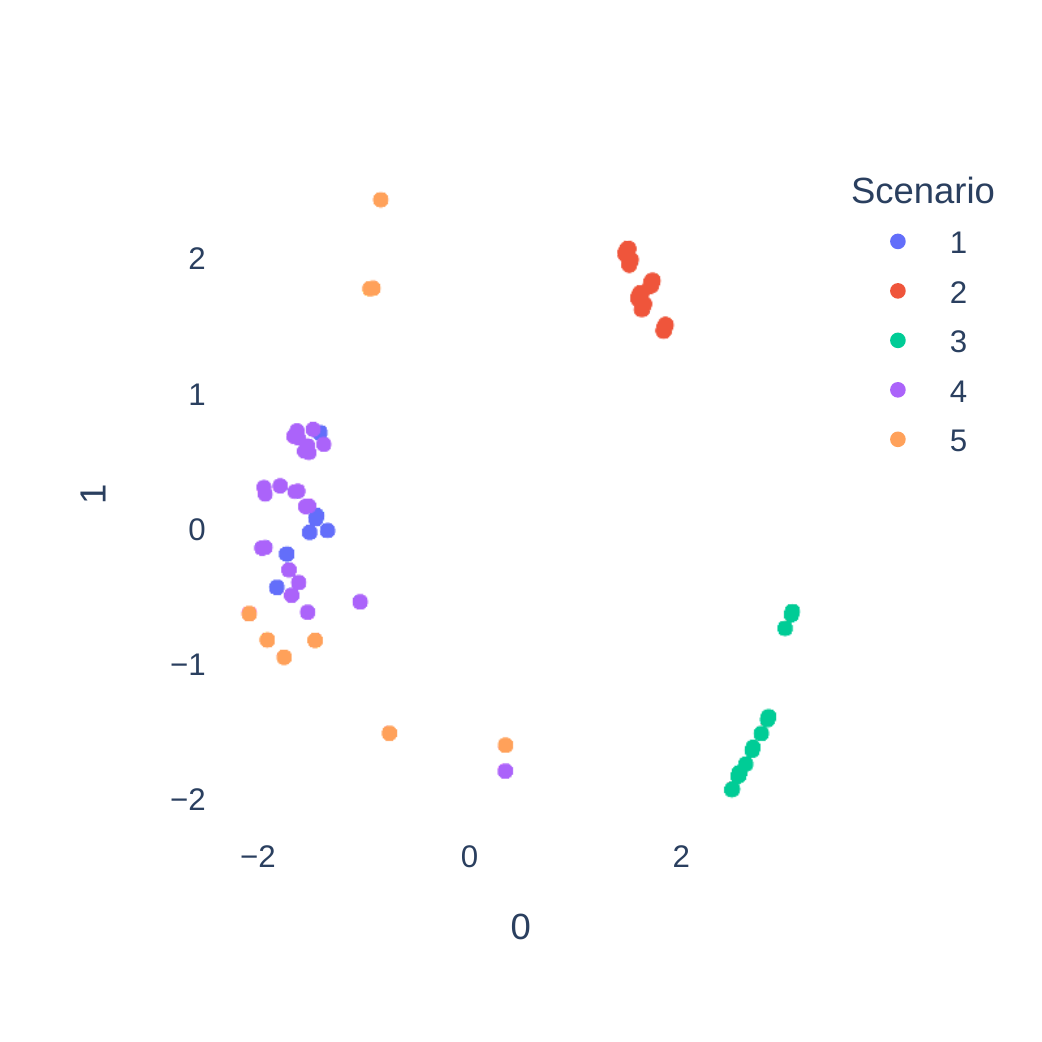}}&
    \subfloat[n-grams]{\includegraphics[clip, trim=1.5cm 1.5cm 2cm 1cm, width = 1.0in]{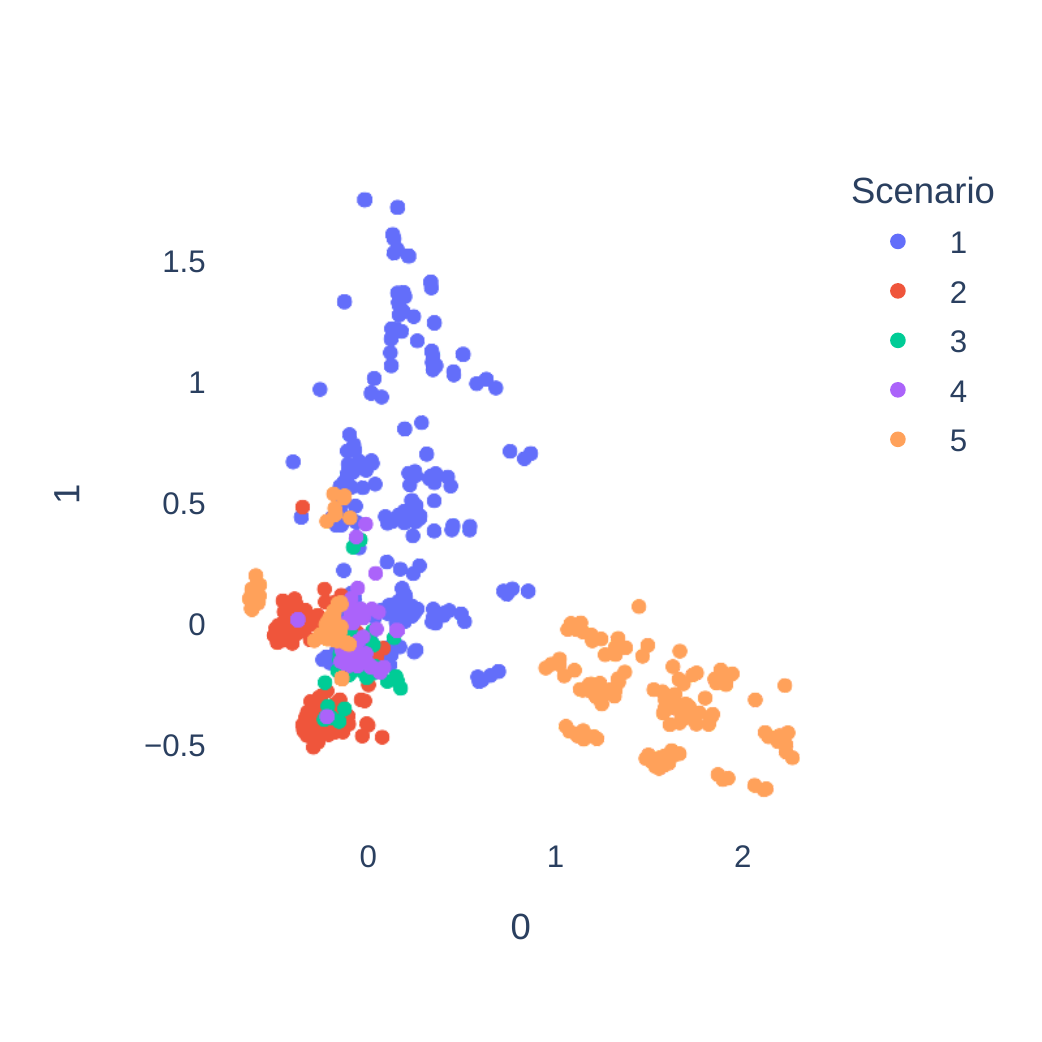}} &    \subfloat[position profile]{\includegraphics[clip, trim=1.5cm 1.5cm 2cm 1cm, width = 1.0in]{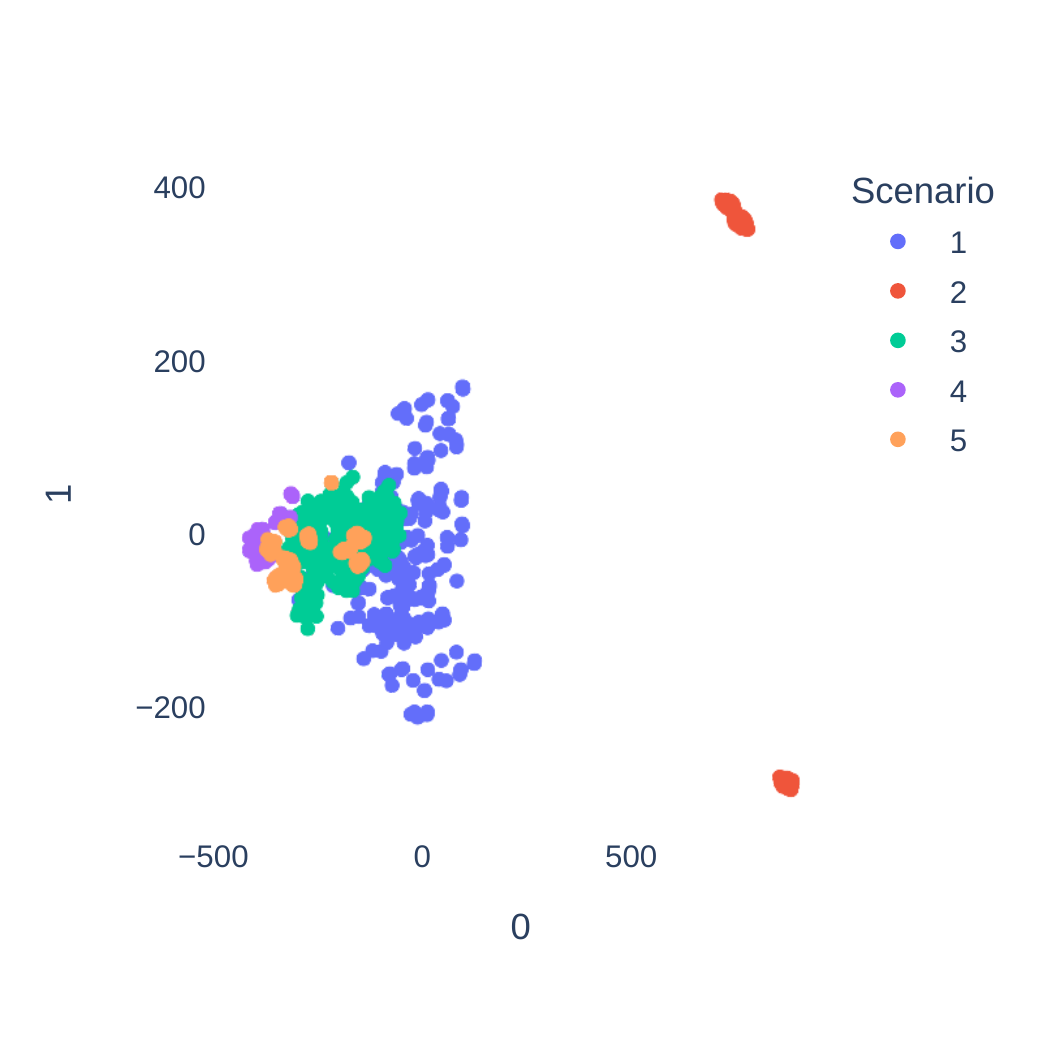}}
     \end{tabular}
    \caption{2D projections based on PCA transformation of the feature vectors of Baseline Family encoding methods of event logs from five different scenarios (1, 2, 3, 4, and 5). Each projection regards an encoding method, each point an encoded trace and each color represents a scenario.}
    \label{fig:PCA_baselines}
\end{figure}

Figure~\ref{fig:PCA_baselines} shows the PCA projections of Baseline methods (\emph{count2vec}, \emph{n-grams}, \emph{position profile} , and \emph{one-hot}). Figure~\ref{fig:PCA_baselines}.a and Figure~\ref{fig:PCA_baselines}.b, regarding \emph{count2vec} and \emph{one-hot}, look similar, assigning more or less the same areas to the same scenarios, but keeping uncovered a significant part of the space. This is due to the methods being very related, with the difference being that \emph{count2vec} accounts for frequencies. Since most traces do not have a high number of repeated activities, when reducing the dimensionality using PCA, the distances in the low-dimensional space are almost the same. On the other hand, \emph{n-grams} (Figure~\ref{fig:PCA_baselines}.c) and \emph{position profile} (Figure~\ref{fig:PCA_baselines}.d) have an average expressivity as different scenarios overlap on the same areas.

\begin{figure}[ht!]
    \centering
    \begin{tabular}{ccc}
    \multicolumn{3}{c}{
      \includegraphics[clip, trim=0cm 9cm 0cm 0cm, width = 3.0in]{PCA/LEGENDA_PCA_walklets_1000_skip_0.1_normal.pdf}}
    \\
    \subfloat[alignment]{\includegraphics[clip, trim=1.5cm 1.5cm 2cm 1cm, width = 1.0in]{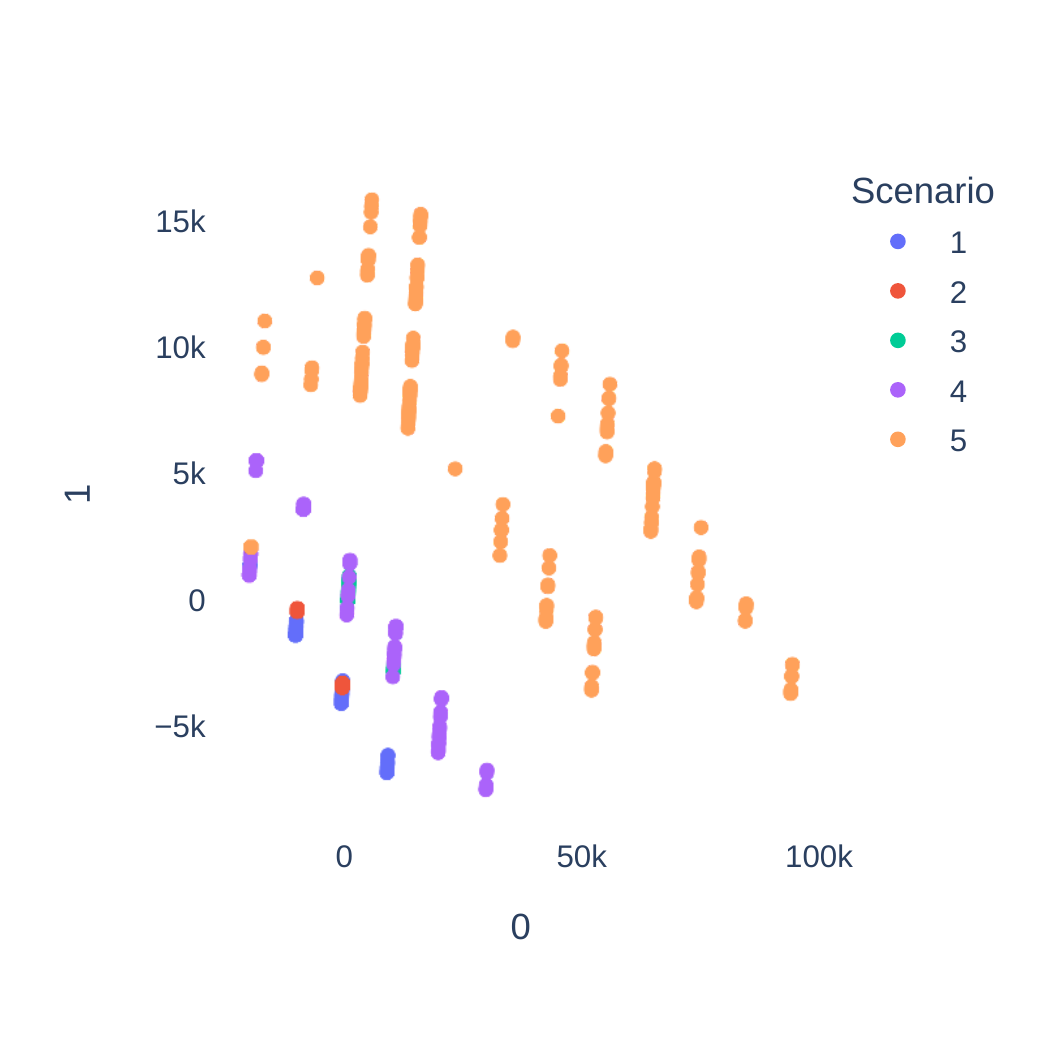}} &
    \subfloat[token-replay]{\includegraphics[clip, trim=1.5cm 1.5cm 2cm 1cm, width = 1.0in]{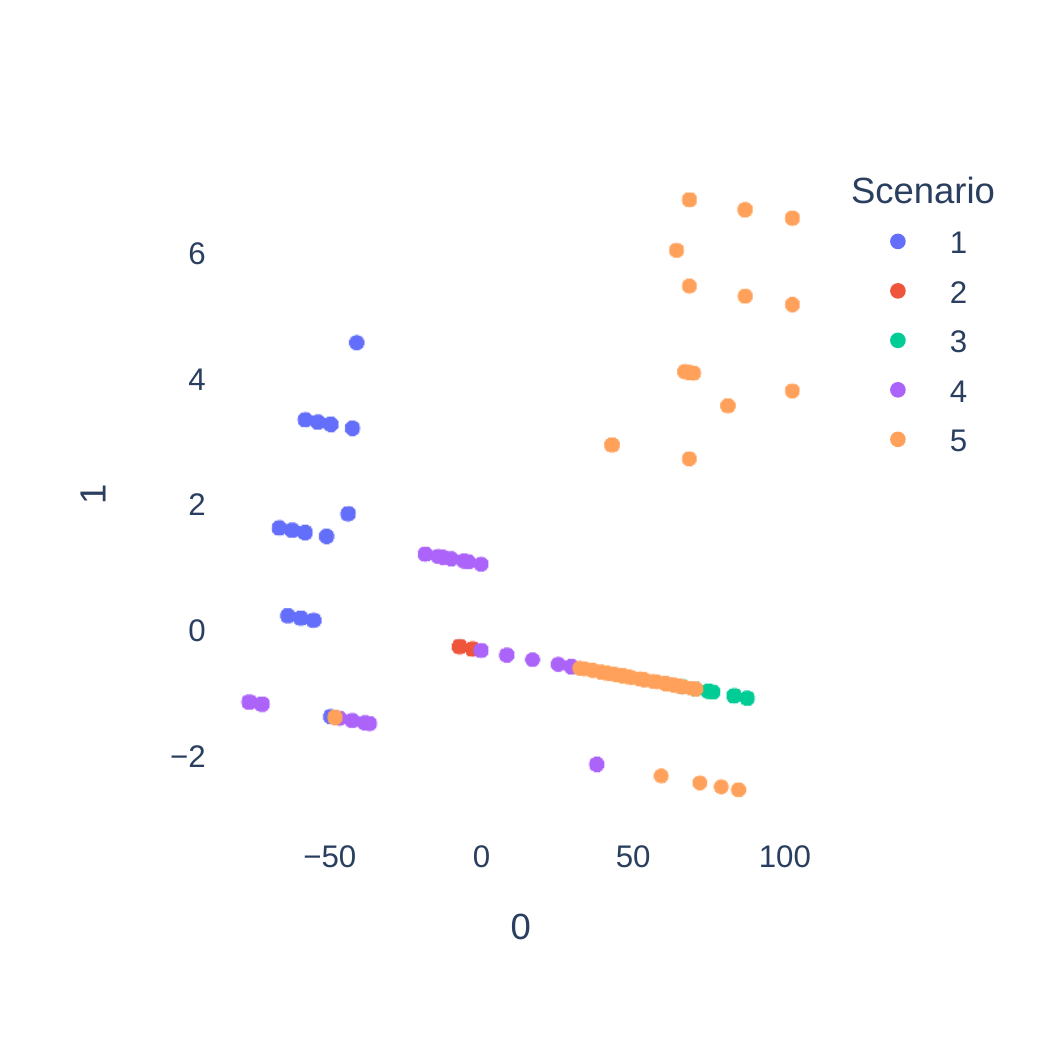}}&
    \subfloat[log skeleton]{\includegraphics[clip, trim=1.5cm 1.5cm 2cm 1cm, width = 1.0in]{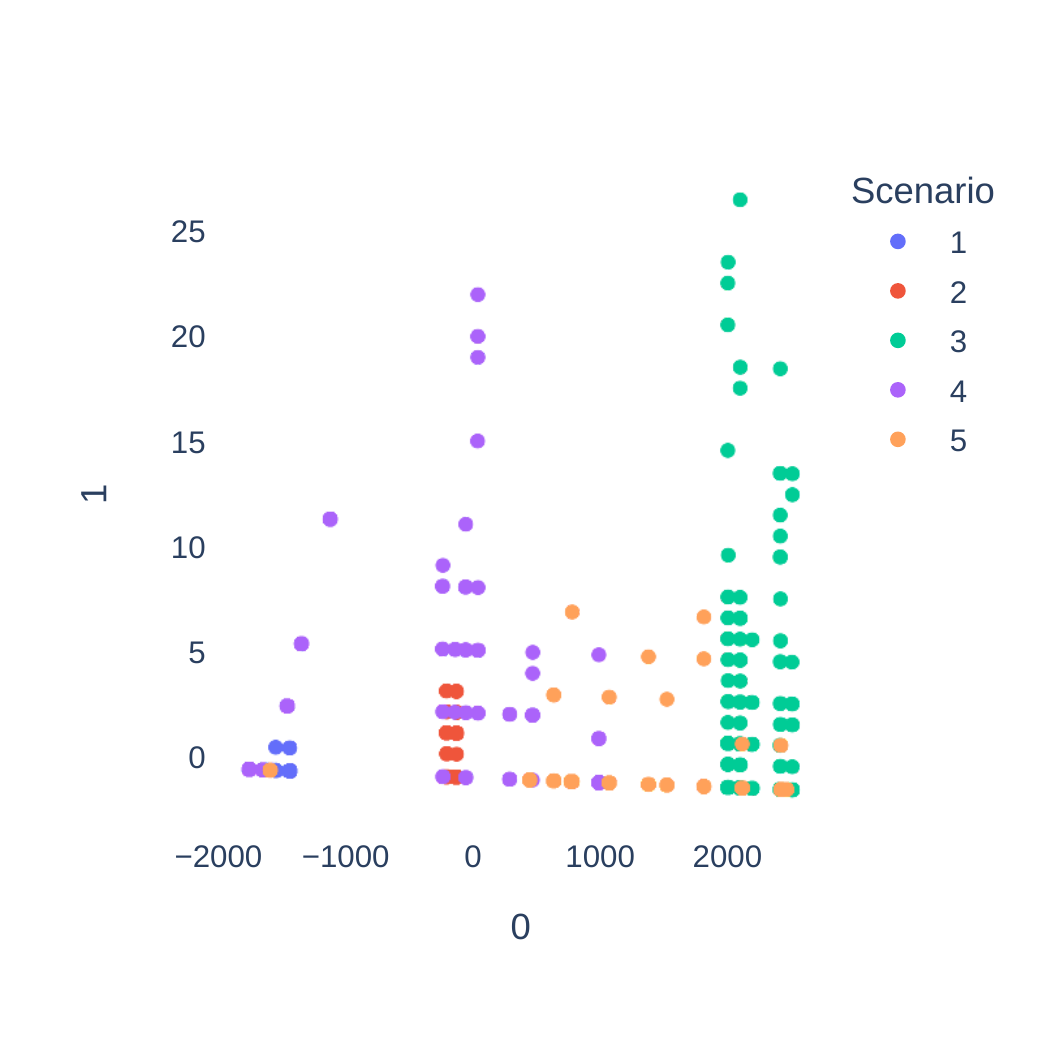}}
     \end{tabular}
    \caption{2D projections based on PCA transformation of the feature vectors of Process Mining encoding family methods of event logs from five different scenarios (1, 2, 3, 4, and 5). Each projection regards an encoding method, each point an encoded event log and each color represents a scenario. }
    \label{fig:PCA_PM}
\end{figure}

Figure~\ref{fig:PCA_PM} shows the projections of PM-based encoding methods. They can be assessed to very high and high expressivity levels, respectively \emph{alignment}  (Figure~\ref{fig:PCA_PM}.a) with the highest level, followed by \emph{token-replay}  (Figure~\ref{fig:PCA_PM}.b) and \emph{Log skeleton}  (Figure~\ref{fig:PCA_PM}.c). It is worth observing that the distribution in the \emph{alignment} projection follows the complexity of scenarios.

\begin{figure}[ht!]
    \centering
    \begin{tabular}{cccc}
    \multicolumn{4}{c}{
      \includegraphics[clip, trim=0cm 9cm 0cm 0cm, width = 3.0in]{PCA/LEGENDA_PCA_walklets_1000_skip_0.1_normal.pdf}}
    \\
    \subfloat[GloVe]{\includegraphics[clip, trim=1.5cm 1.5cm 2cm 1cm, width = 1.0in]{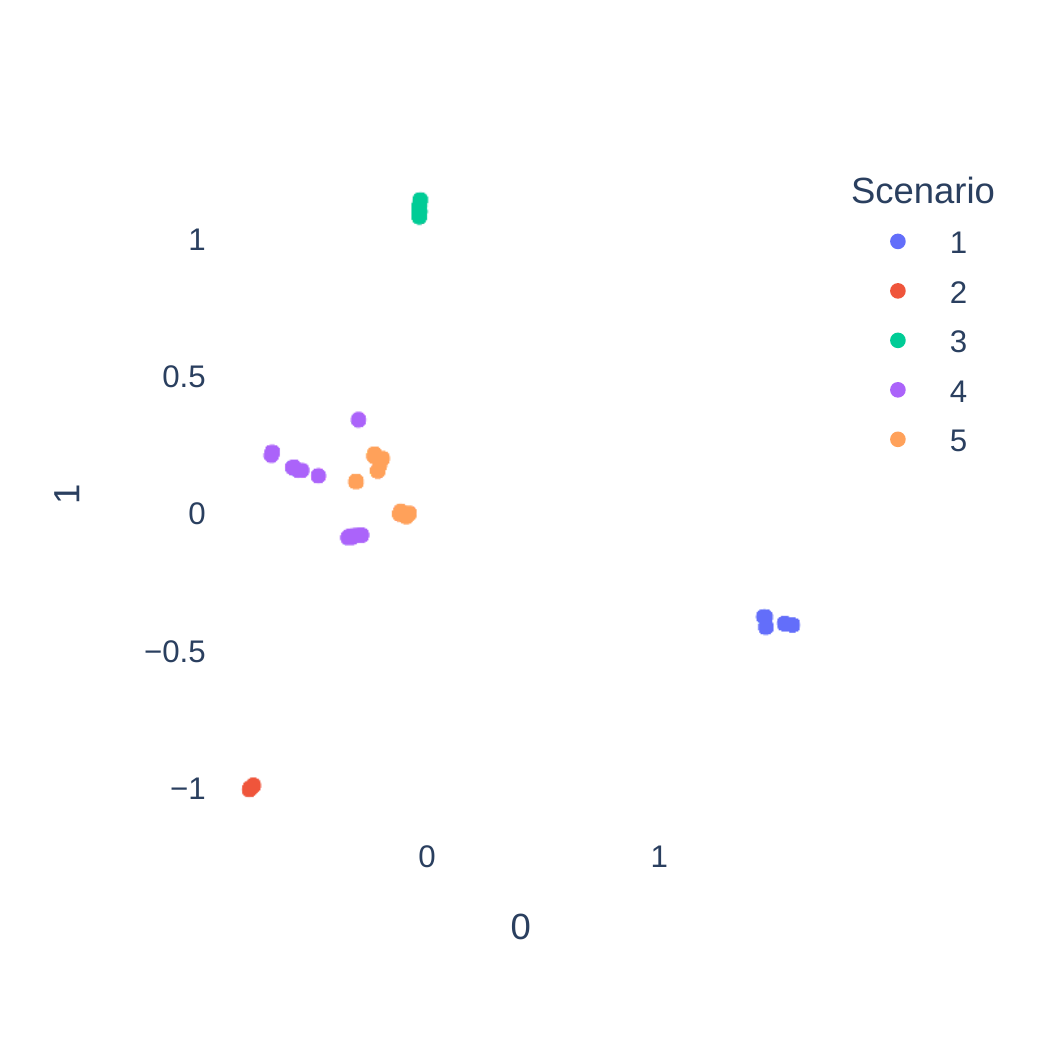}}&
   \subfloat[hash2vec]{\includegraphics[clip, trim=1.5cm 1.5cm 2cm 1cm, width = 1.0in]{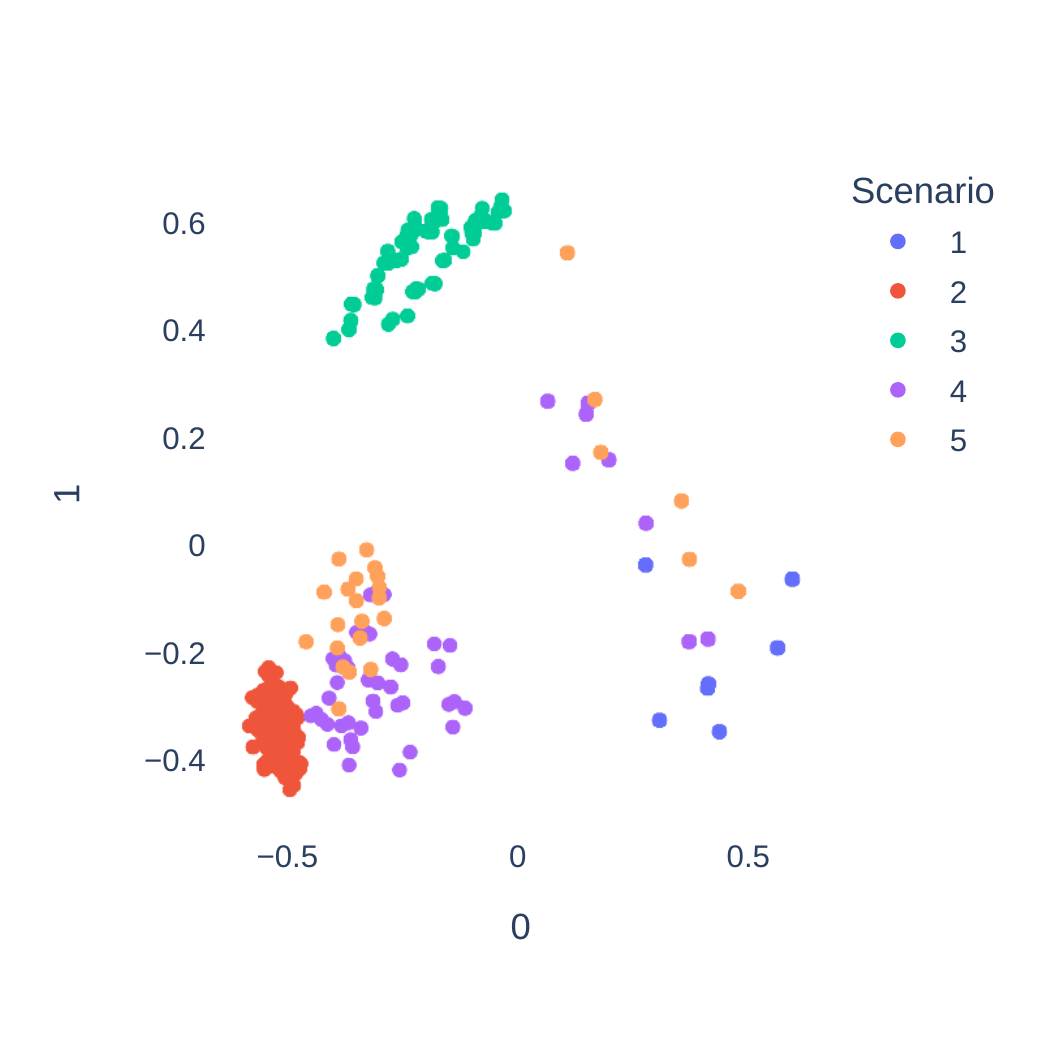}} &
    \subfloat[TF-IDF]{\includegraphics[clip, trim=1.5cm 1.5cm 2cm 1cm, width = 1.0in]{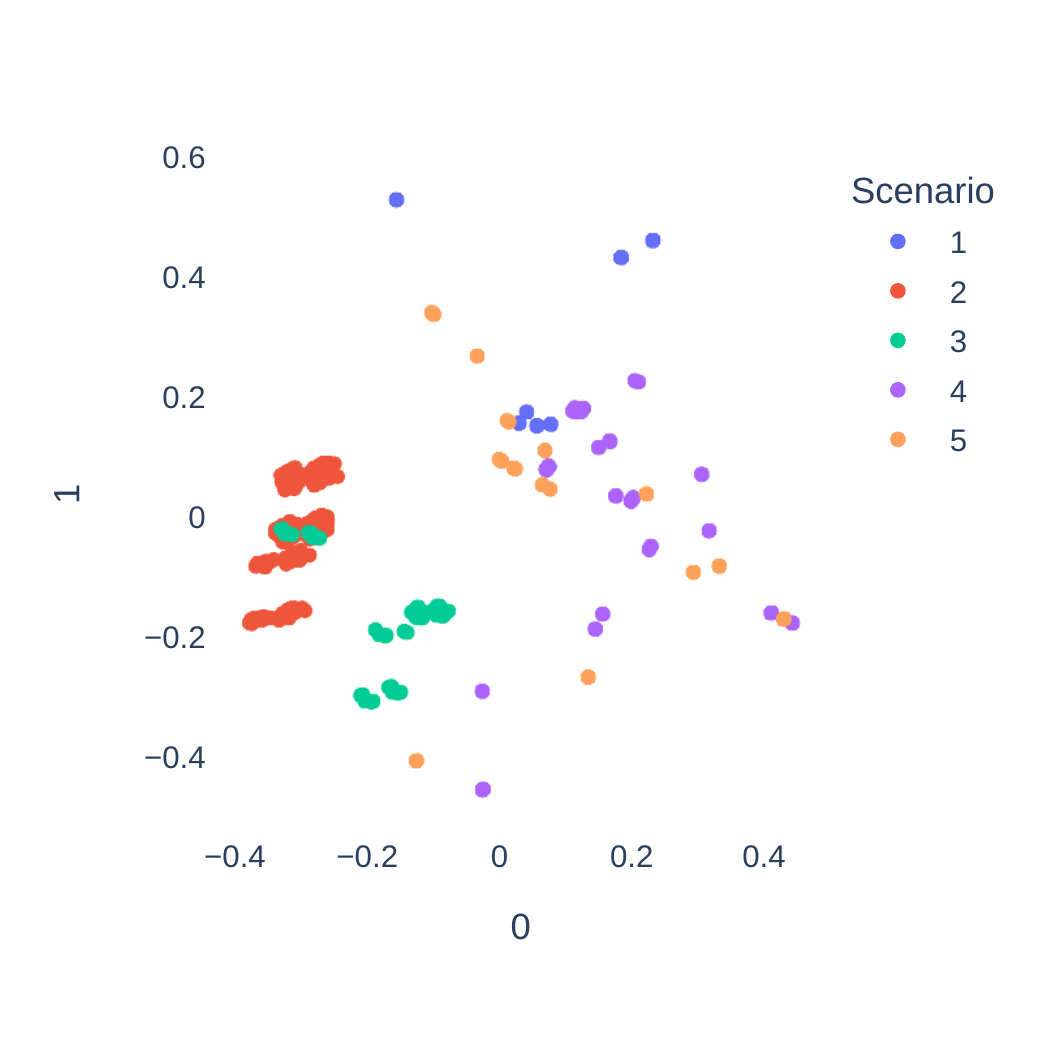}} &
        
    \subfloat[CBOW]{\includegraphics[clip, trim=1.5cm 1.5cm 2cm 1cm, width = 1.0in]{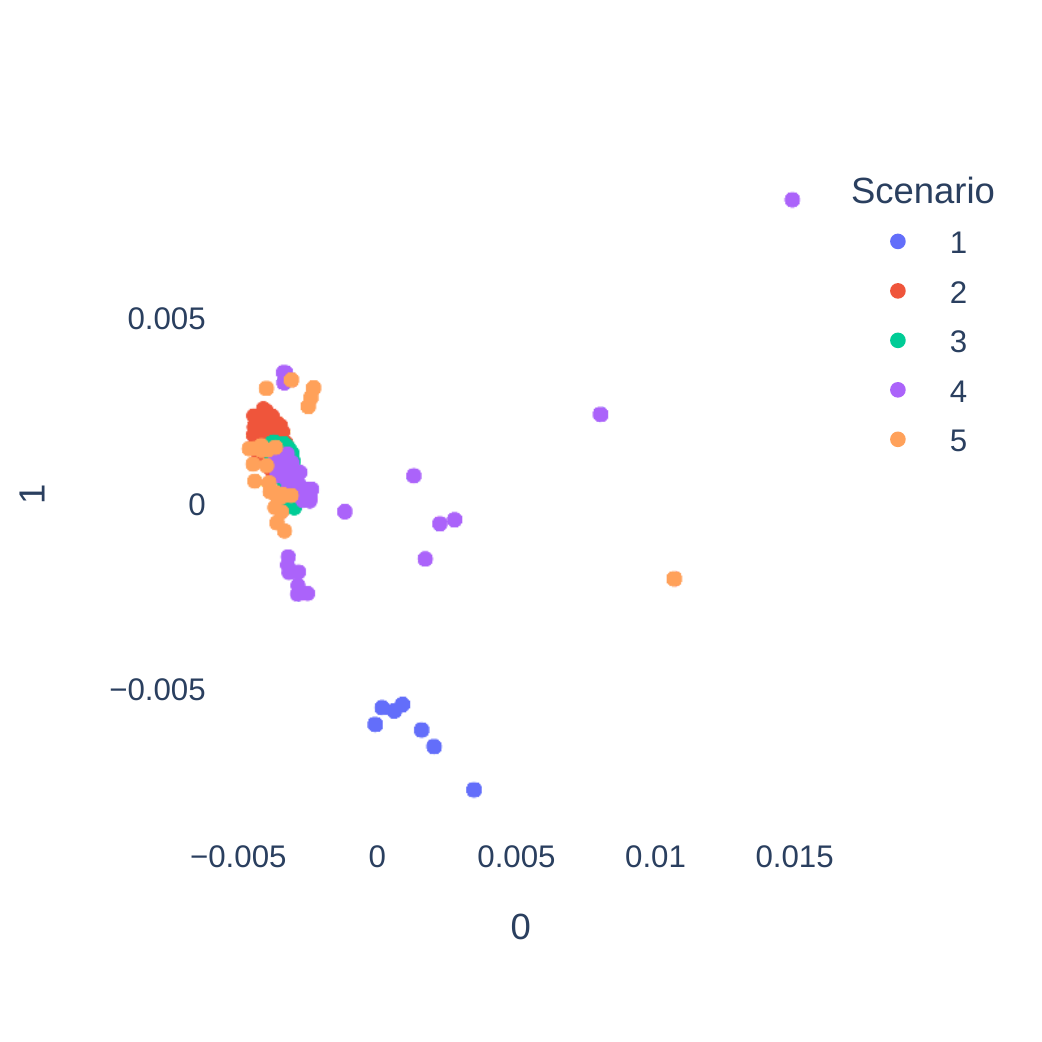}} \\
    \subfloat[skip-gram]{\includegraphics[clip, trim=1.5cm 1.5cm 2cm 1cm, width = 1.05in]{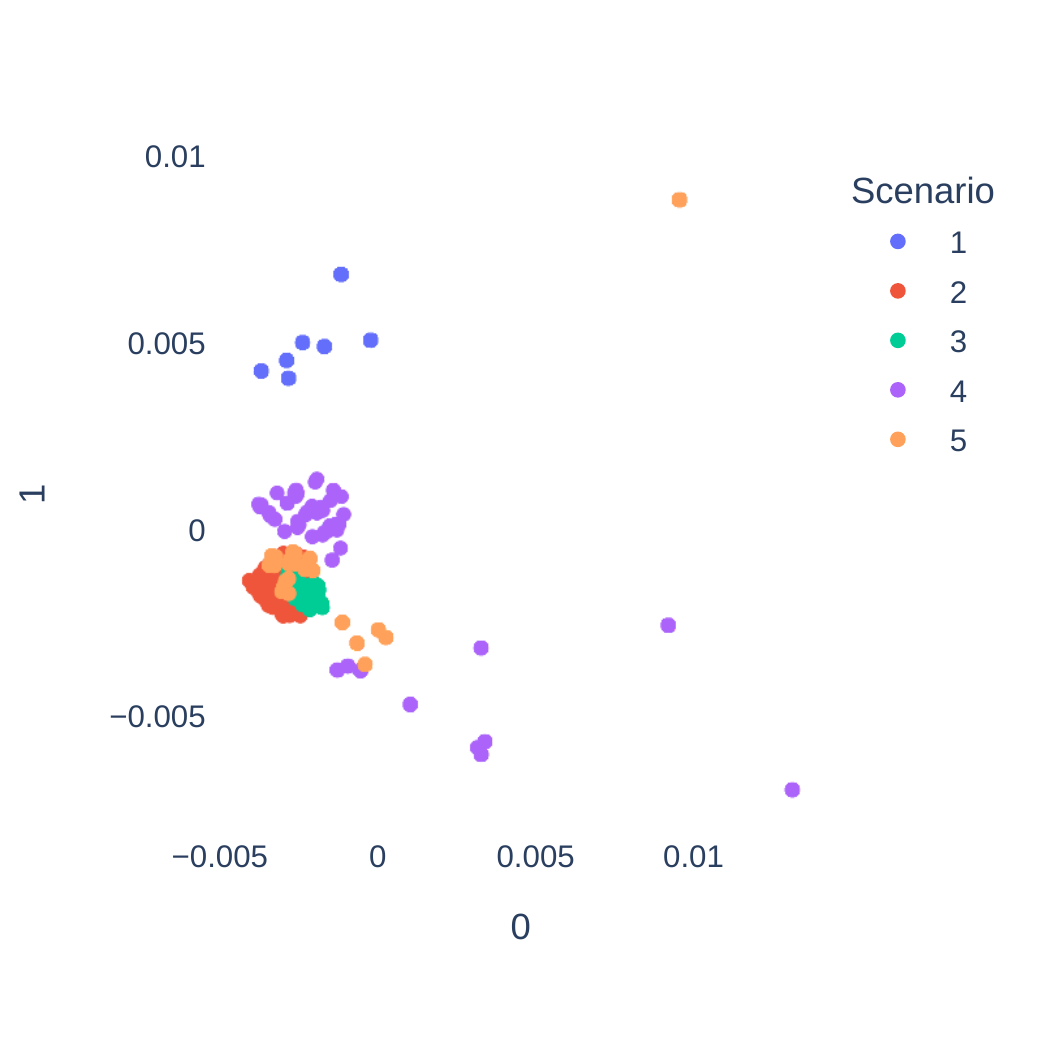}} &
    \subfloat[doc2vec]{\includegraphics[clip, trim=1.5cm 1.5cm 2cm 1cm, width = 1.0in]{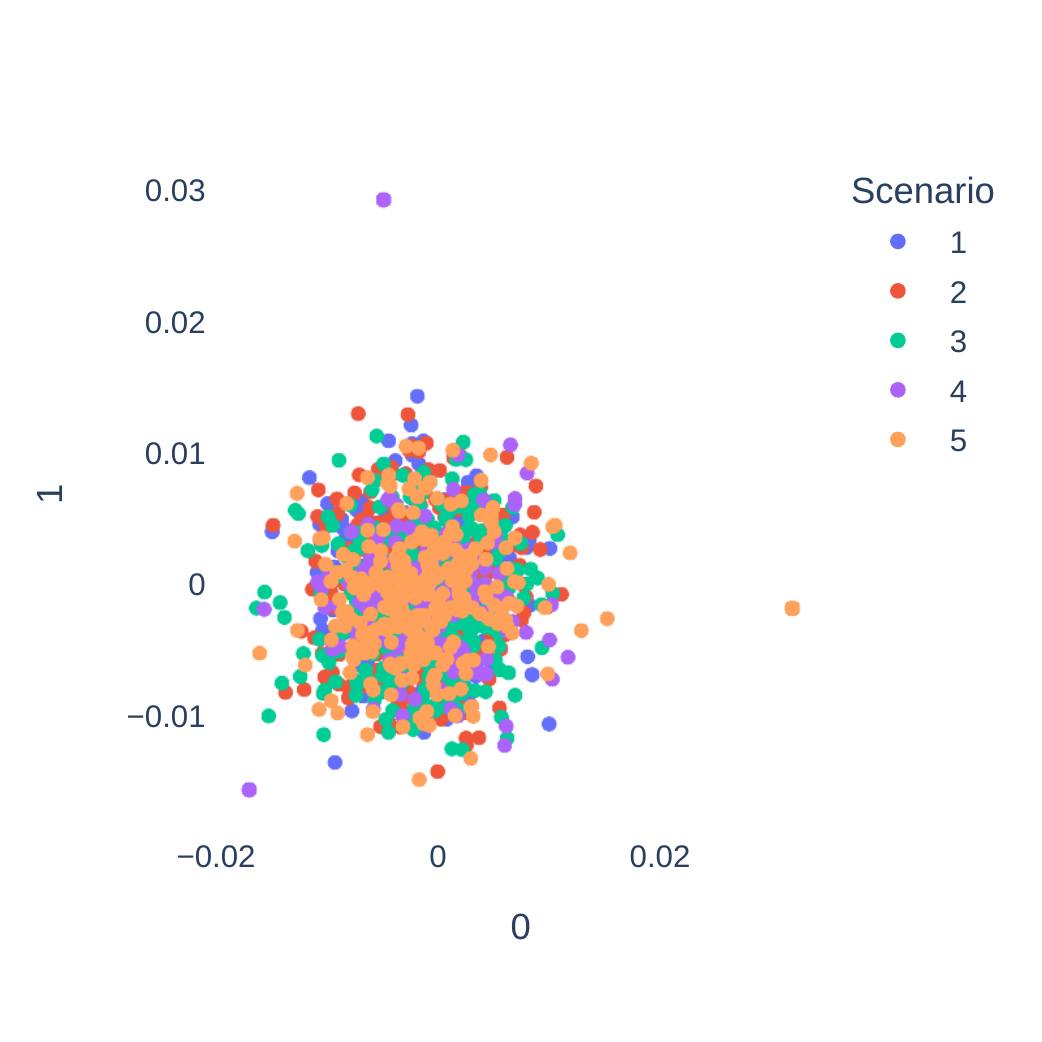}} &
        
     \end{tabular}
    \caption{The figure illustrates the 2D projections based on PCA transformation of the feature vectors of Text encoding family methods of event logs from five different scenarios (1, 2, 3, 4, and 5). Each projection regards an encoding method, each point an encoded event log and each color represents a scenario.}
     \label{fig:PCA_Text}
\end{figure}

The Text encoding family presented average, low, and very low levels of expressivity, as shown in Figure~\ref{fig:PCA_Text}. \emph{GloVe} and \emph{hash2vec} are from average level, as observed in Figure~\ref{fig:PCA_Text}.a and Figure~\ref{fig:PCA_Text}.b. Low levels of expressivity were obtained by \emph{TF-IDF} (Figure~\ref{fig:PCA_Text}.c), \emph{CBOW} (Figure~\ref{fig:PCA_Text}.d) and \emph{skip-gram} (Figure~\ref{fig:PCA_Text}.e). The worst expressivity level, very low, was obtained by \emph{doc2vec} (Figure~\ref{fig:PCA_Text}.f).

\begin{figure}[ht!]
    \centering
    \begin{tabular}{cccc}
    \multicolumn{4}{c}{
      \includegraphics[clip, trim=0cm 9cm 0cm 0cm, width = 3.0in]{PCA/LEGENDA_PCA_walklets_1000_skip_0.1_normal.pdf}}
    \\
    \subfloat[BoostNE]{\includegraphics[clip, trim=1.5cm 1.5cm 2cm 1cm, width = 1.0in]{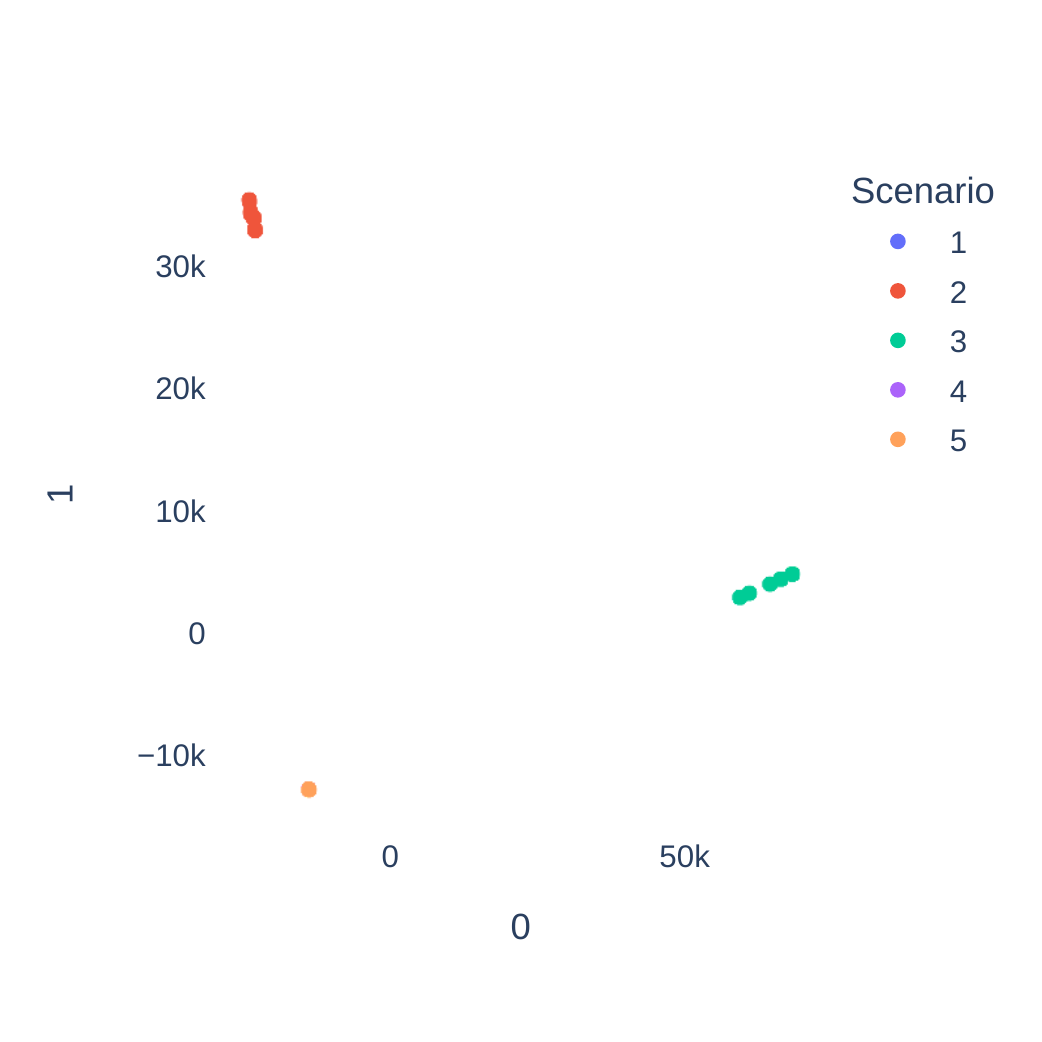}} &
    \subfloat[DeepWalk]{\includegraphics[clip, trim=1.5cm 1.5cm 2cm 1cm, width = 1.0in]{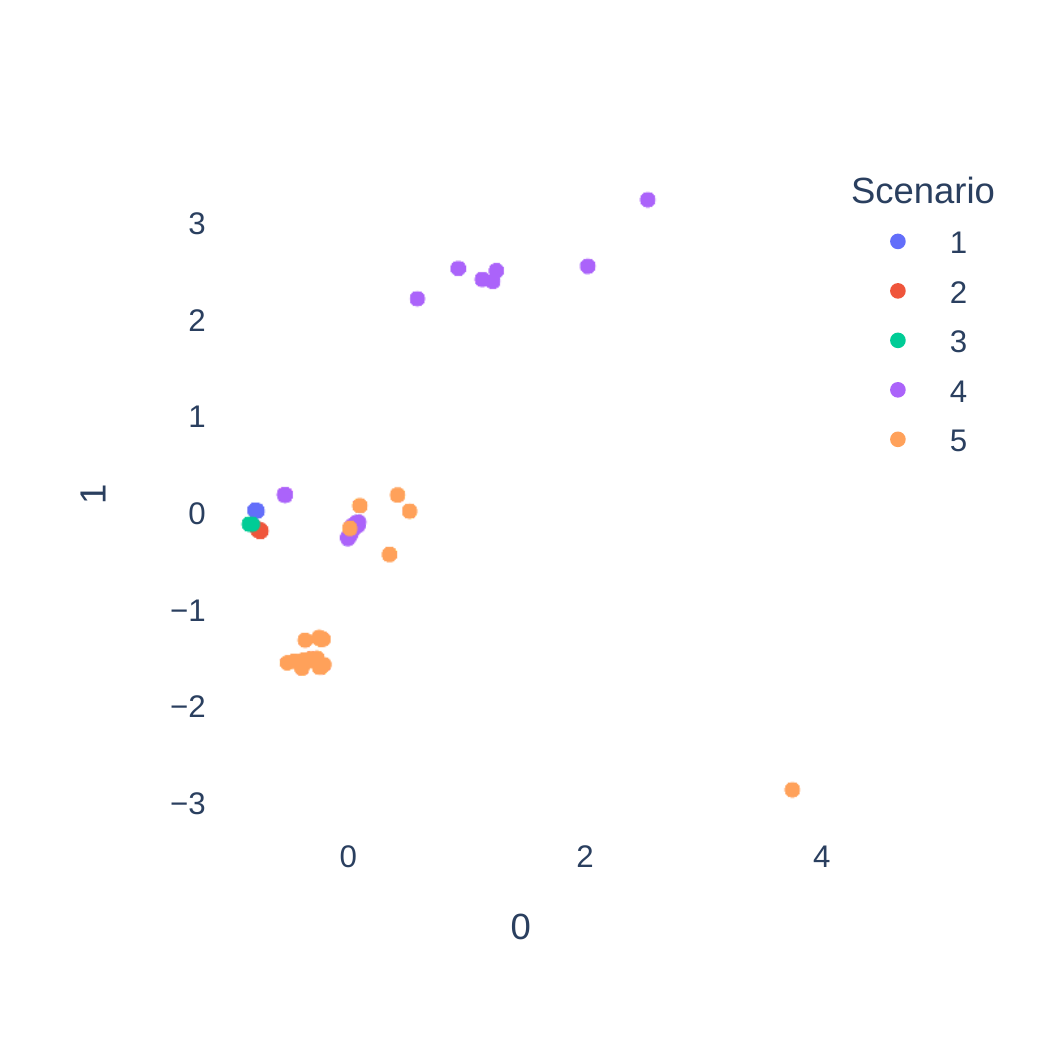}} &
    \subfloat[diff2vec]{\includegraphics[clip, trim=1.5cm 1.5cm 2cm 1cm, width = 1.0in]{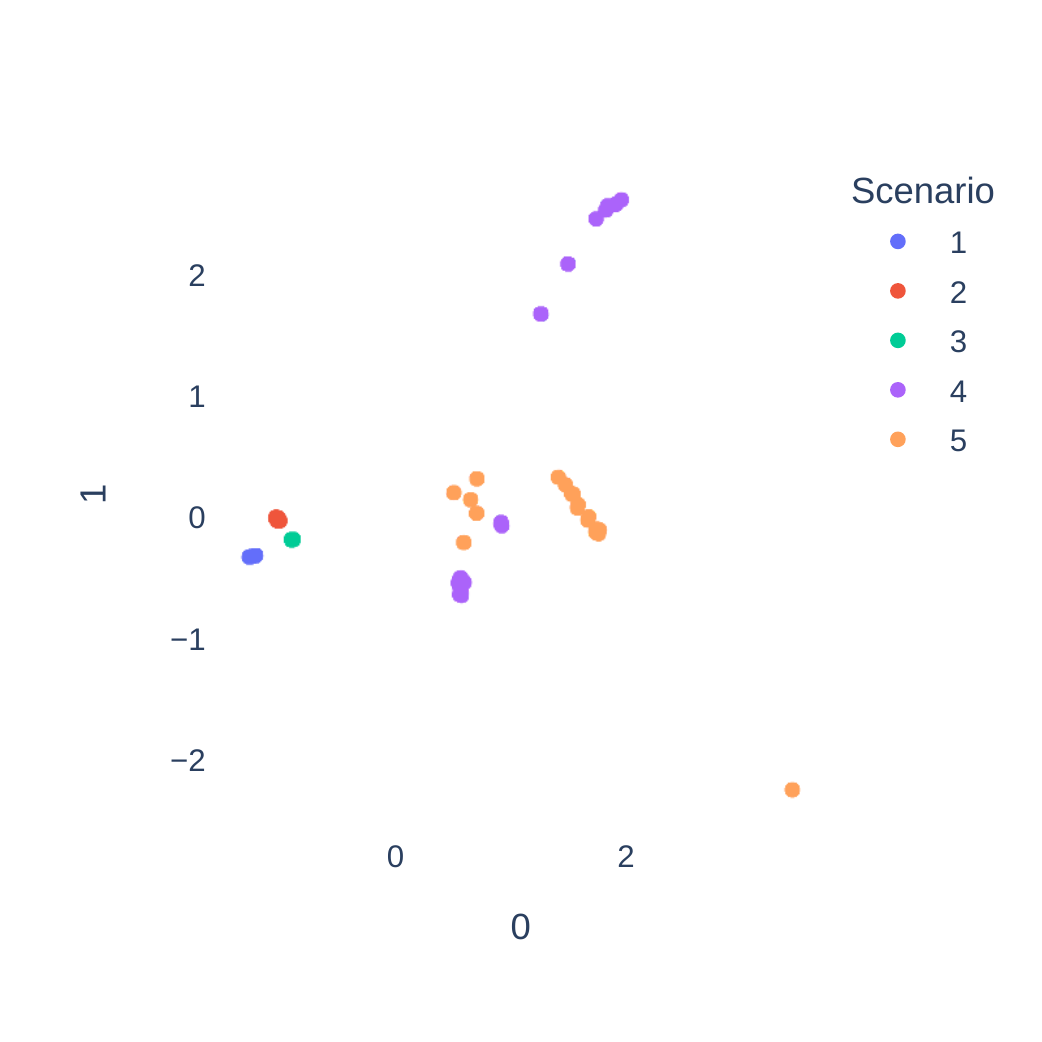}} &
    \subfloat[GLEE]{\includegraphics[clip, trim=1.5cm 1.5cm 2cm 1cm, width = 1.0in]{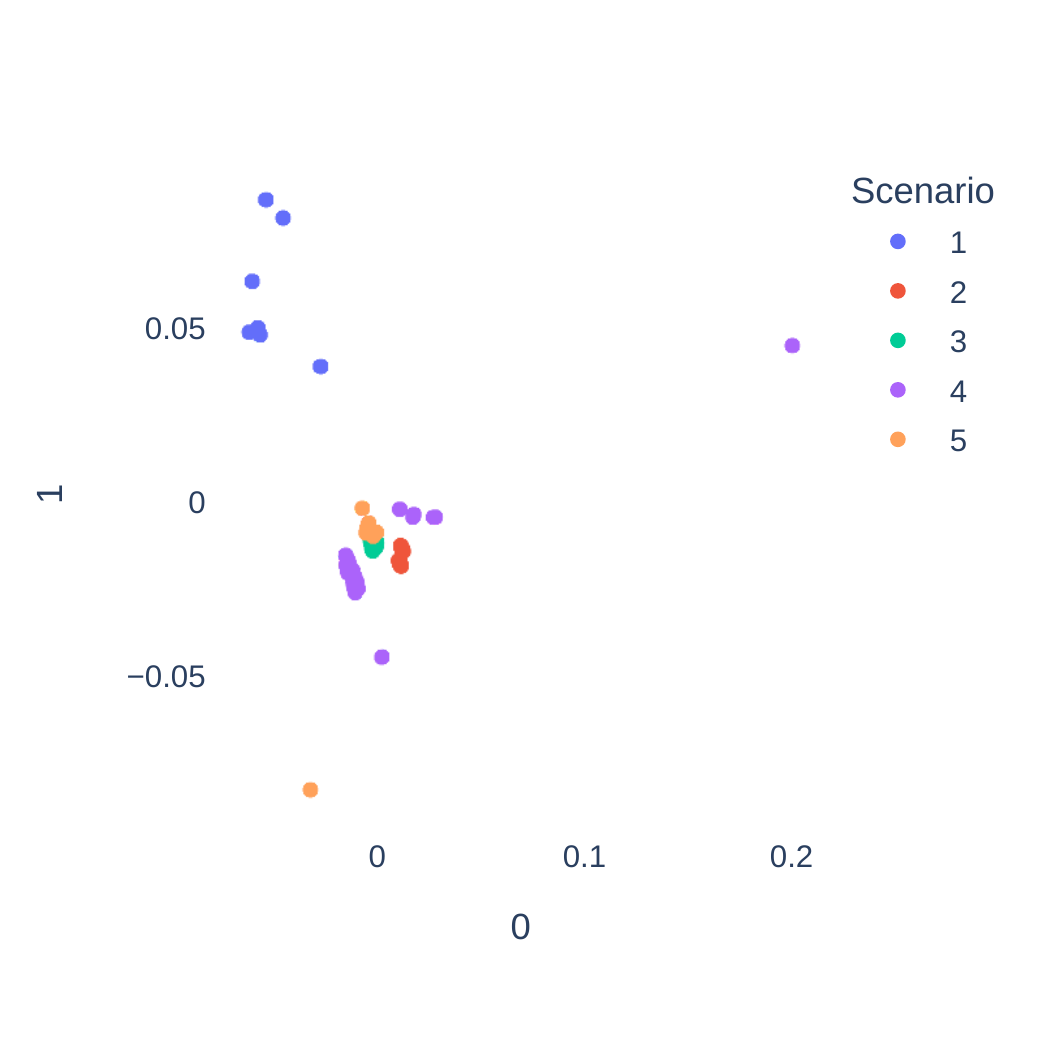}} \\
    
     \subfloat[GraphWave]{\includegraphics[clip, trim=1.5cm 1.5cm 2cm 1cm, width = 1.0in]{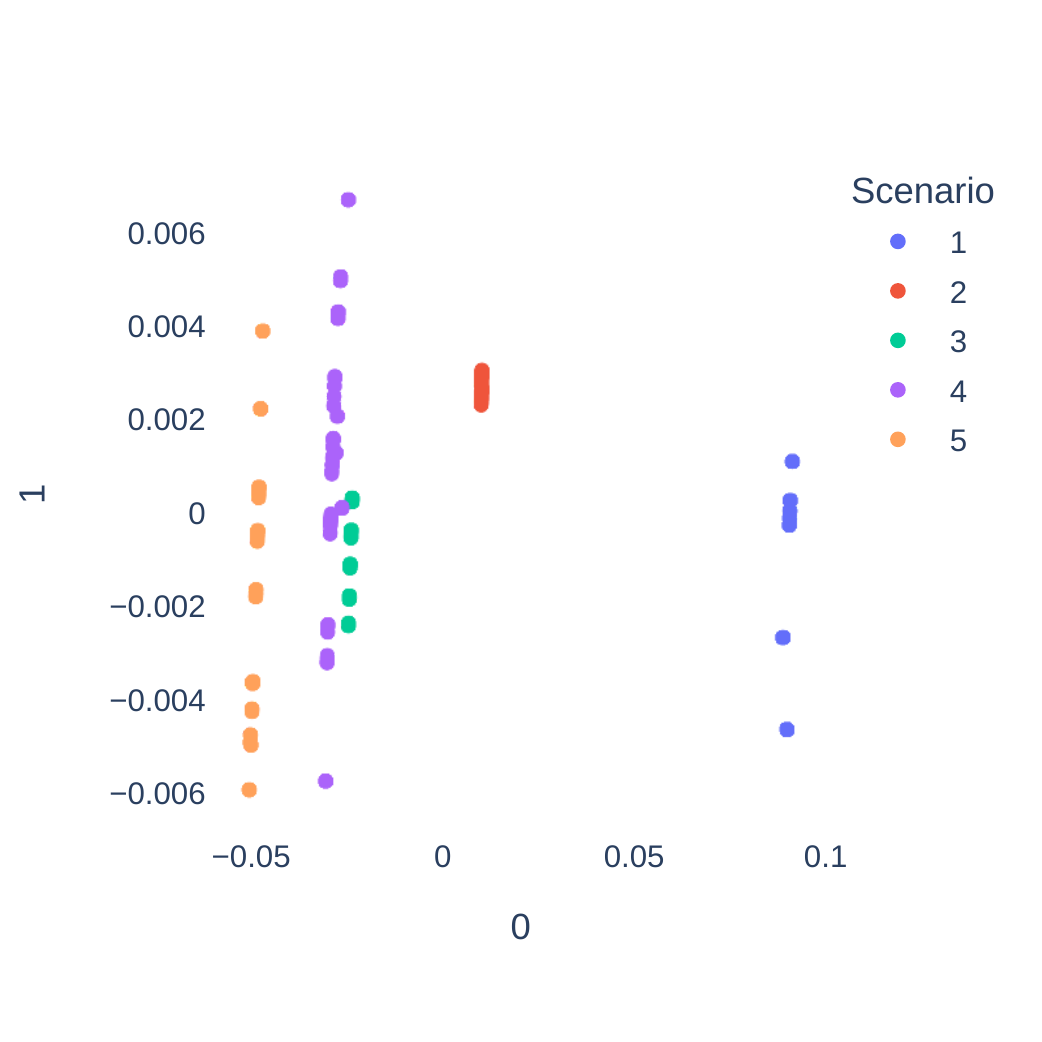}} &
    \subfloat[GraRep]{\includegraphics[clip, trim=1.5cm 1.5cm 2cm 1cm, width = 1.0in]{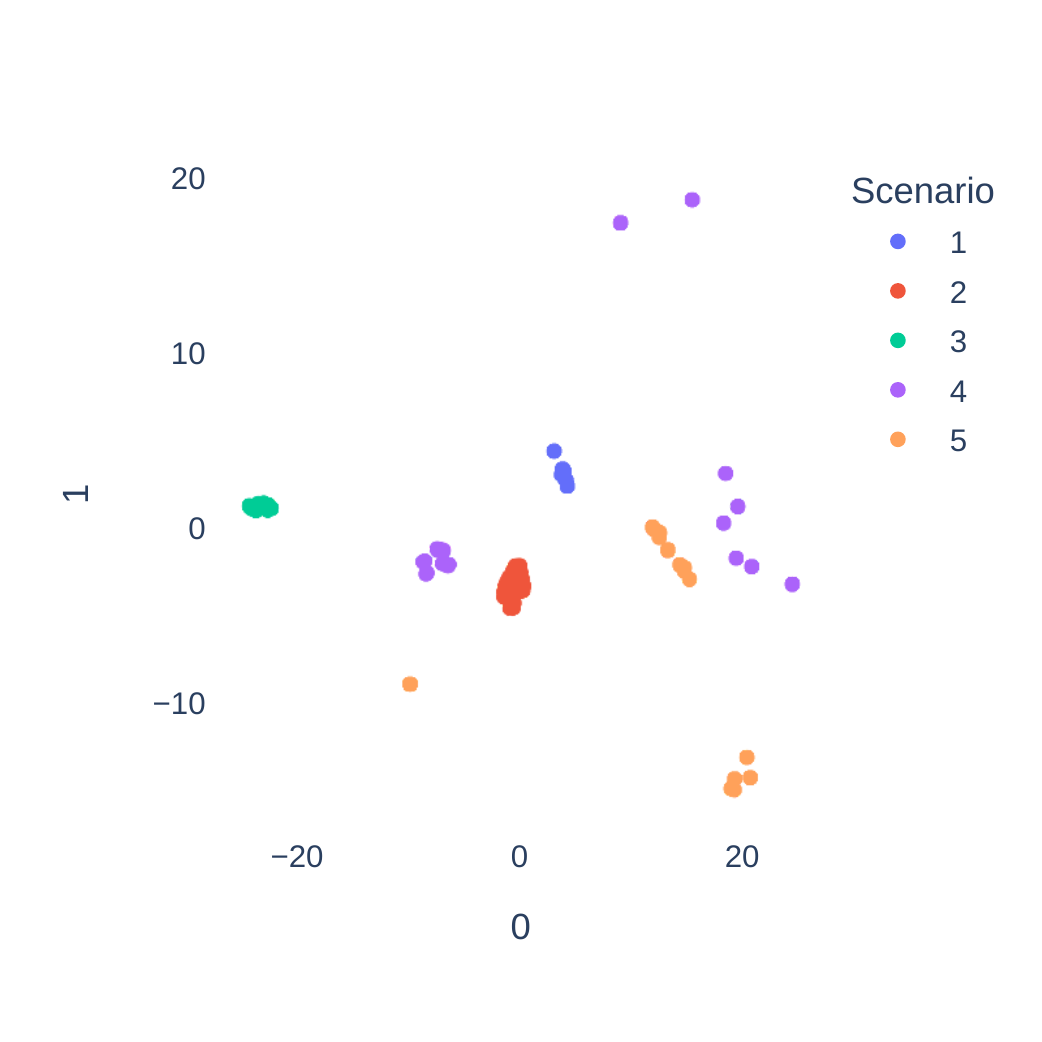}} &
    \subfloat[Hope]{\includegraphics[clip, trim=1.5cm 1.5cm 2cm 1cm, width = 1.0in]{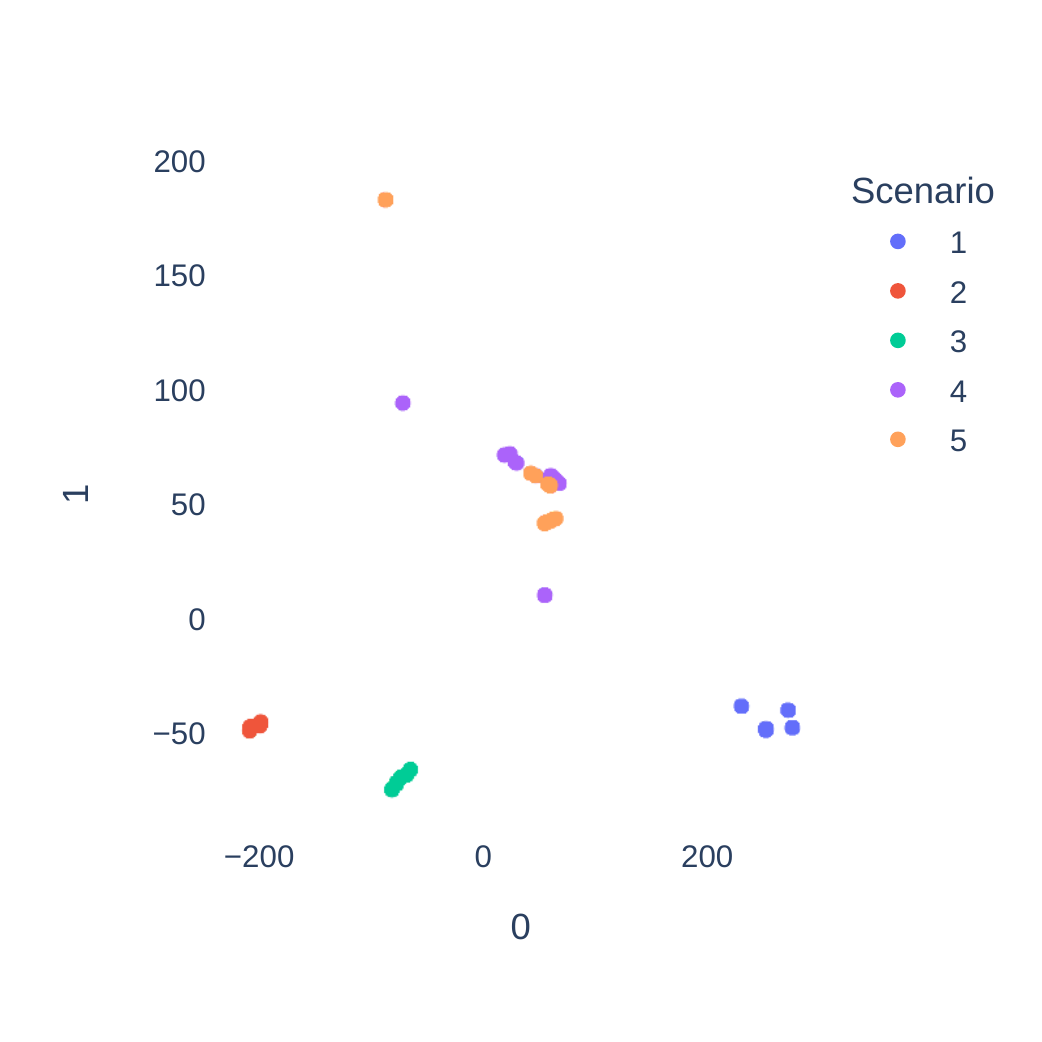}}&    
    \subfloat[Laplacian]{\includegraphics[clip, trim=1.5cm 1.5cm 2cm 1cm, width = 1.0in]{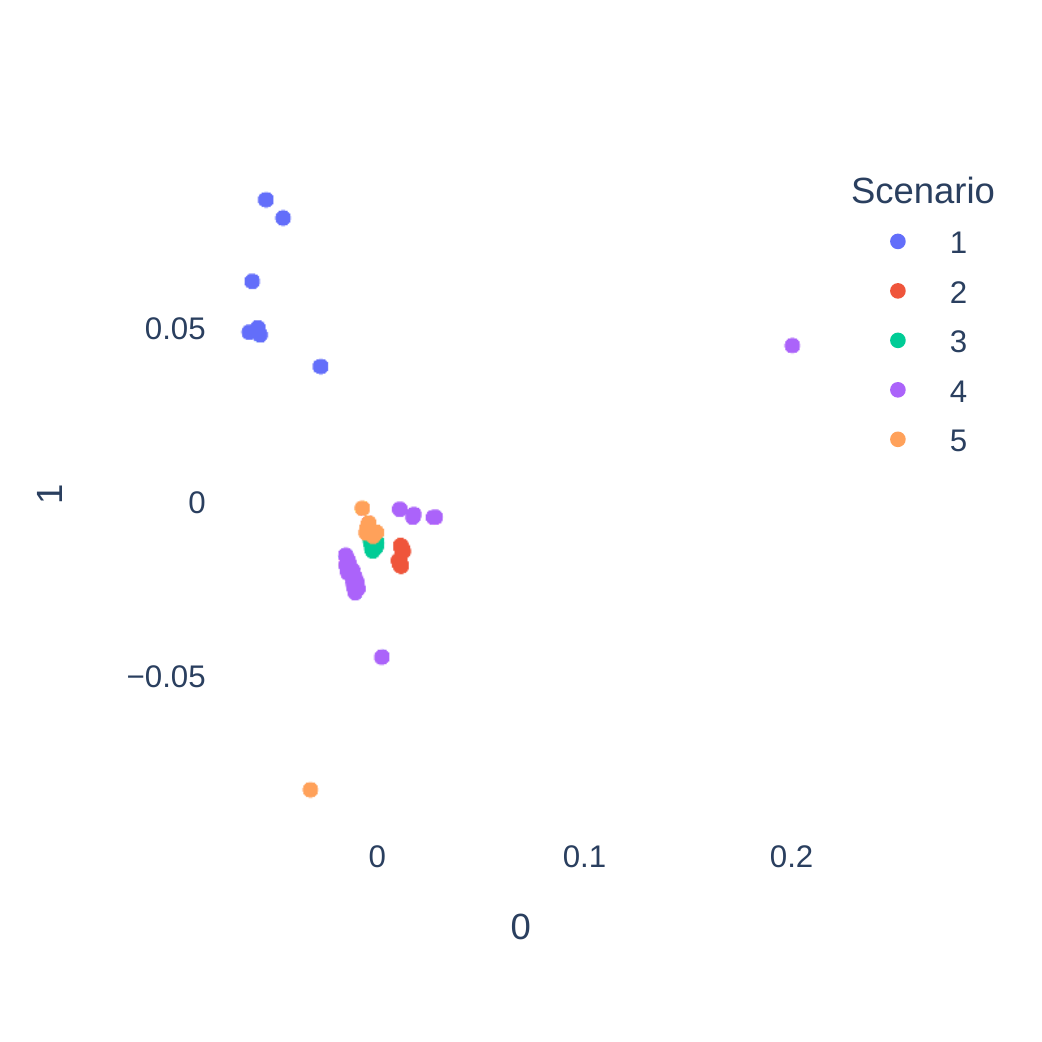}}\\

    \subfloat[NetMF]{\includegraphics[clip, trim=1.5cm 1.5cm 2cm 1cm, width = 1.0in]{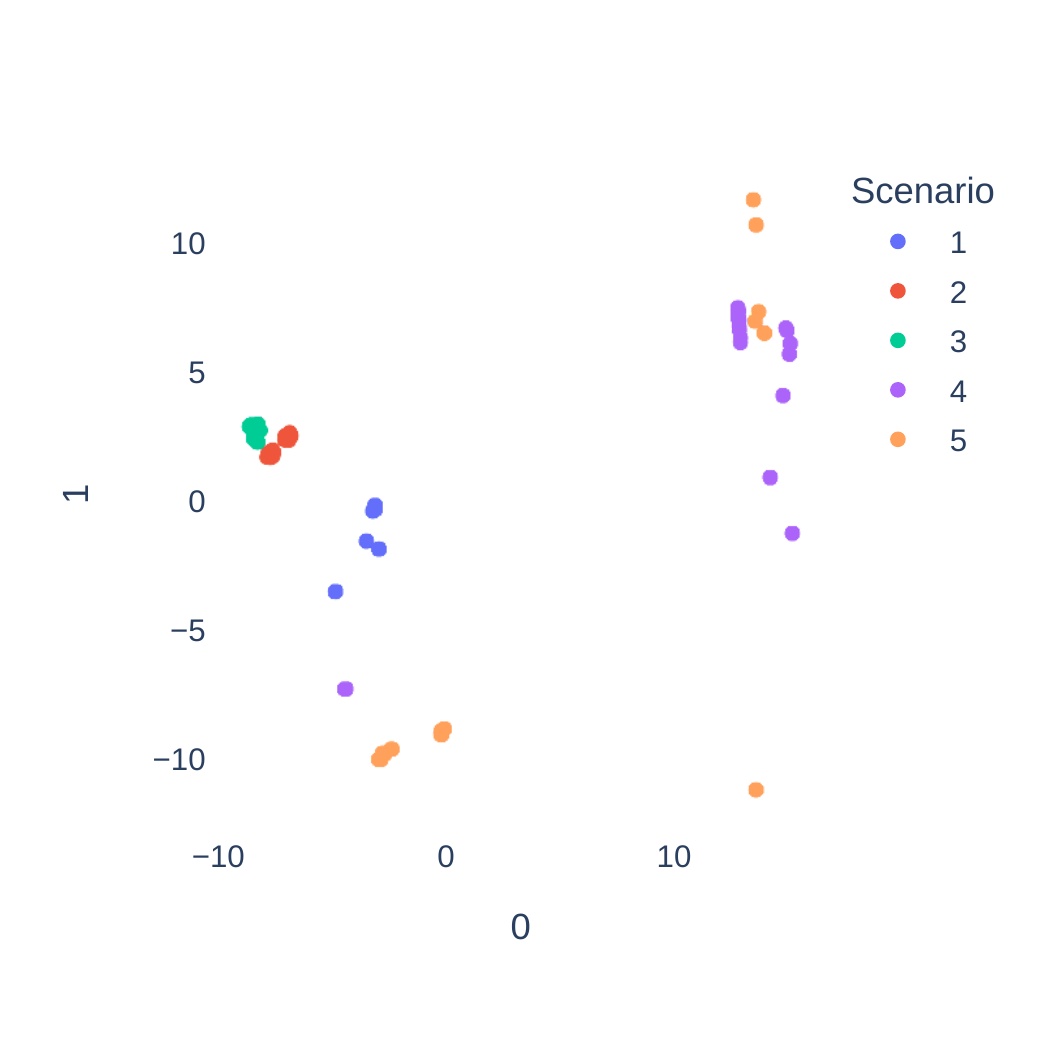}} &
    \subfloat[NMF-ADMM]{\includegraphics[clip, trim=1.5cm 1.5cm 2cm 1cm, width = 1.0in]{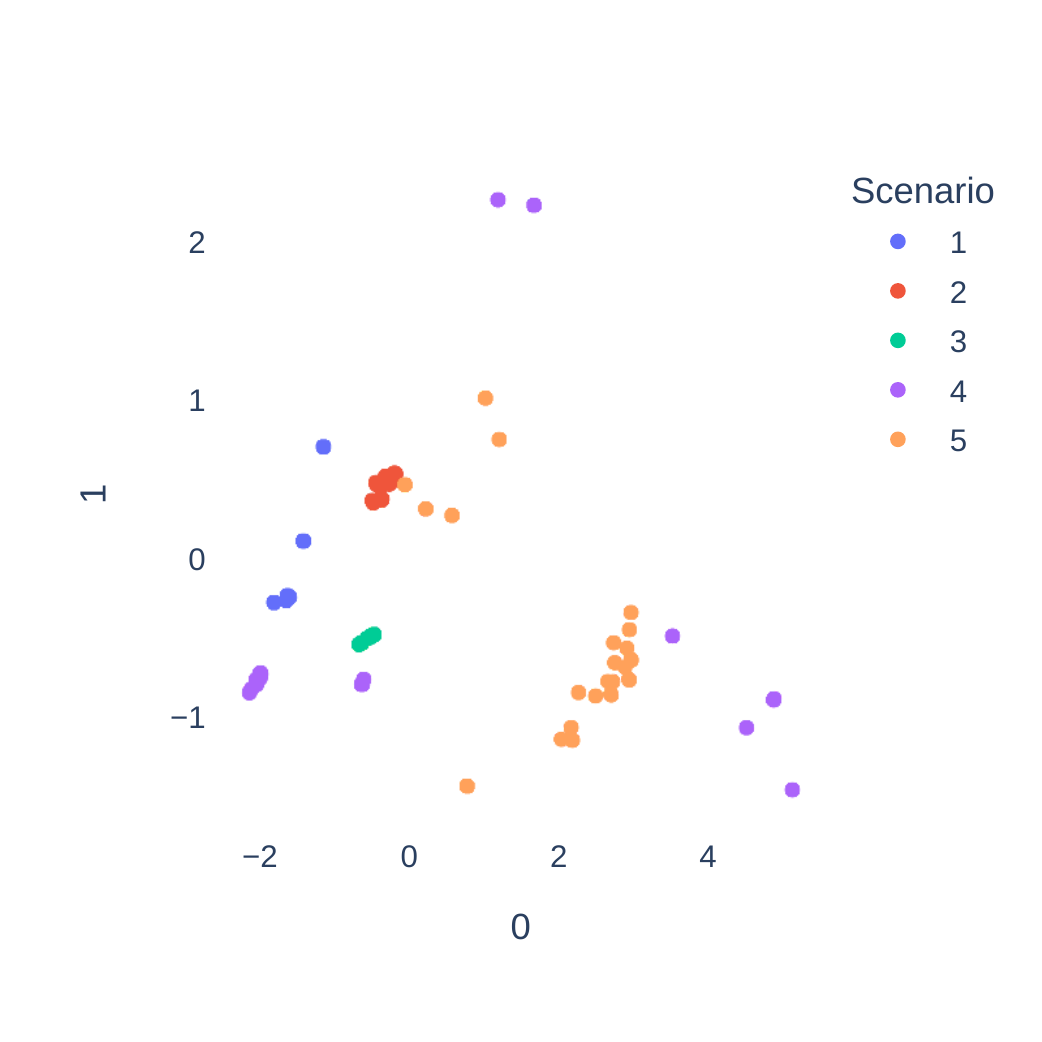}}&
    \subfloat[node2vec]{\includegraphics[clip, trim=1.5cm 1.5cm 2cm 1cm, width = 1.0in]{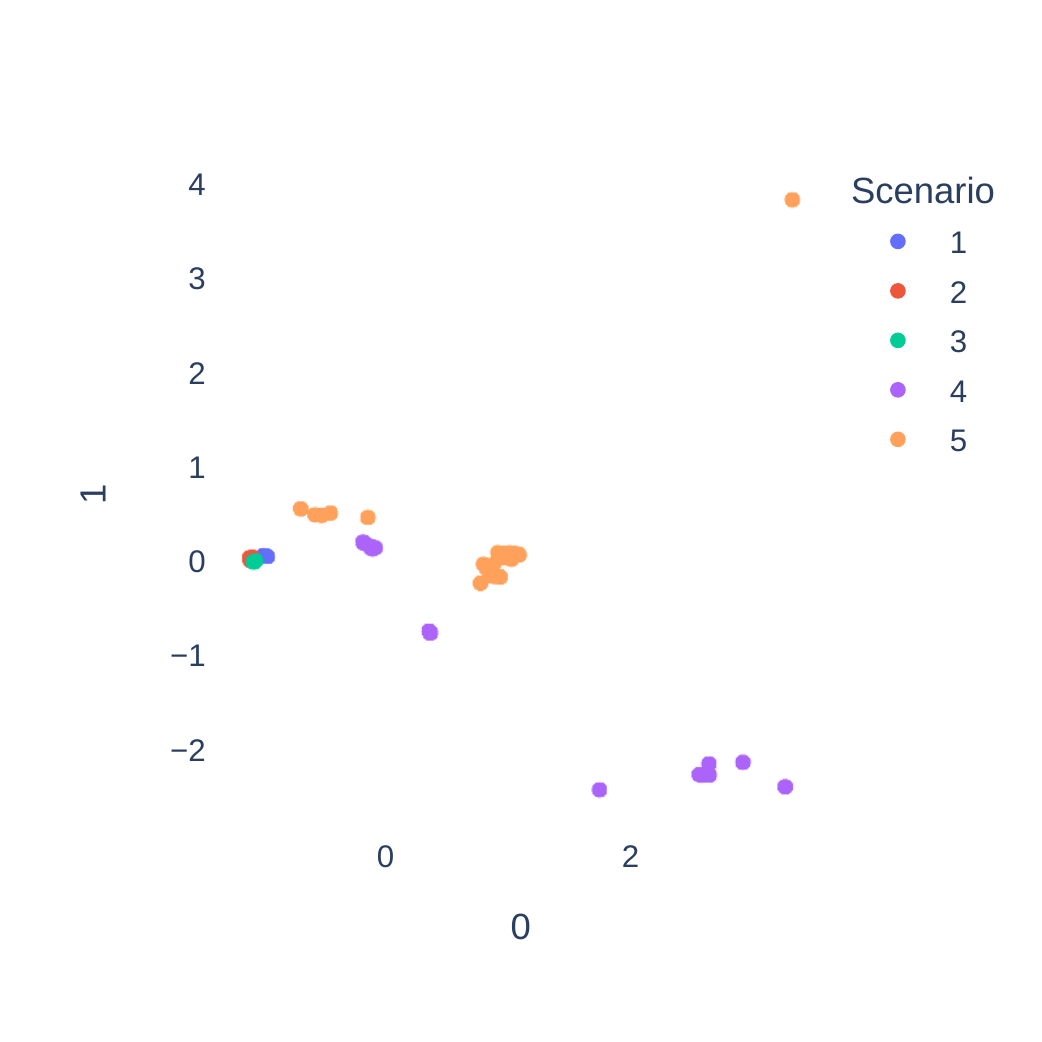}} &
    \subfloat[NodeSketch]{\includegraphics[clip, trim=1.5cm 1.5cm 2cm 1cm, width = 1.0in]{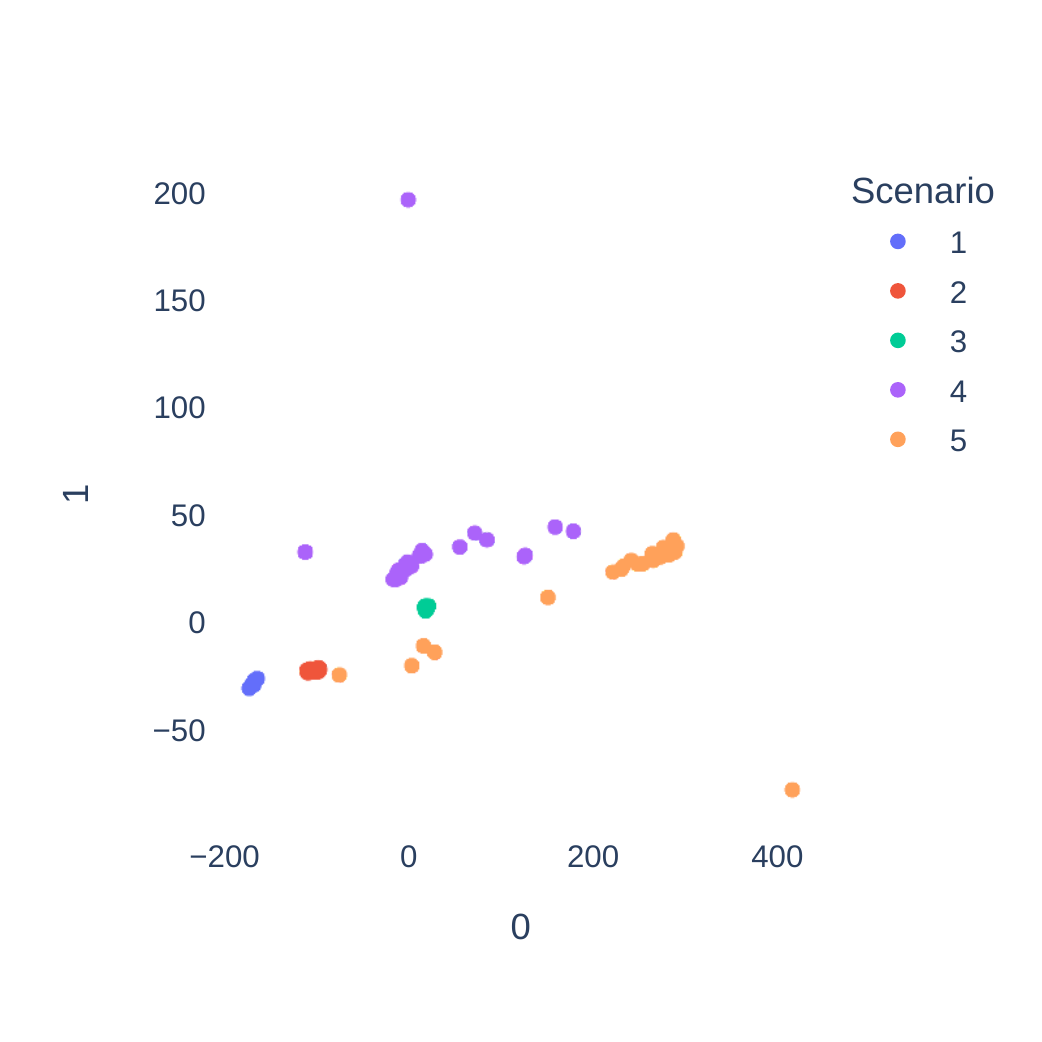}} \\
    
    \subfloat[role2vec]{\includegraphics[clip, trim=1.5cm 1.5cm 2cm 1cm, width = 1.0in]{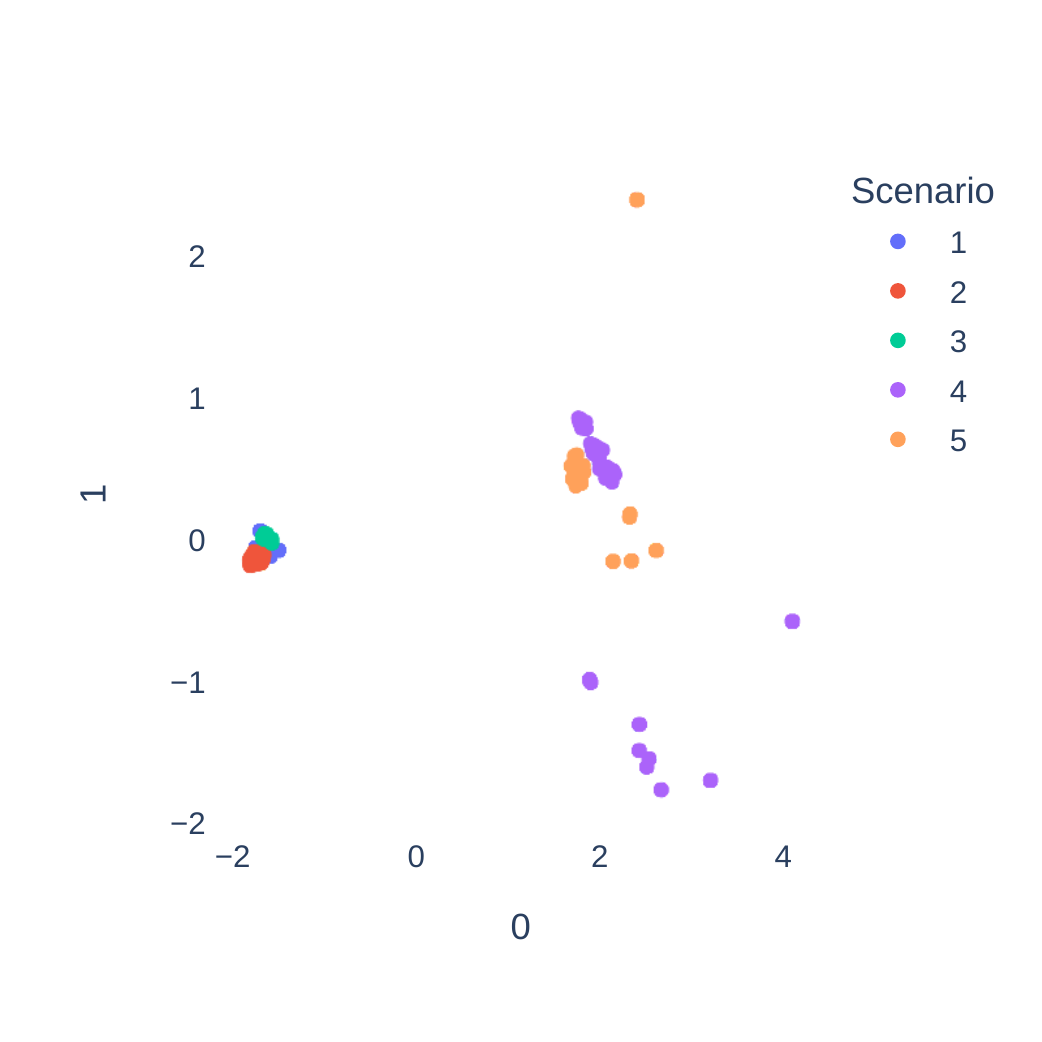}} &
    \subfloat[Walklets]{\includegraphics[clip, trim=1.5cm 1.5cm 2cm 1cm, width = 1.0in]{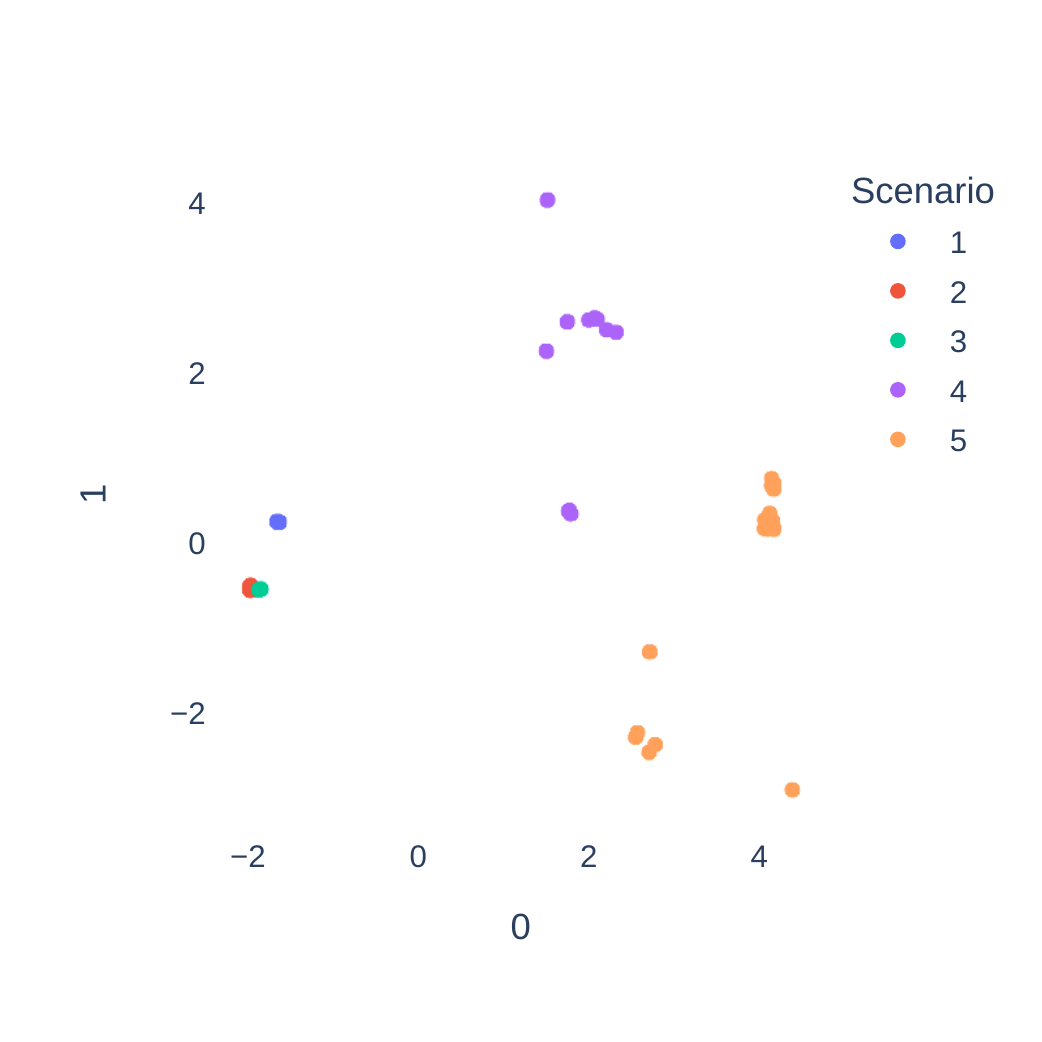}} &
    \end{tabular}
    \caption{2D projections based on PCA transformation of the feature vectors of Graph encoding family methods of event logs from five different scenarios (1, 2, 3, 4, and 5). Each projection regards an encoding method, each point an encoded event log and each color represents a scenario.}
    \label{fig:PCA_Graph}
\end{figure}

Very high, high, and average were the levels observed when using the Graph encoding family (Figure~\ref{fig:PCA_Graph}). Average was obtained by \emph{BoostNE} (Figure~\ref{fig:PCA_Graph}.a) and \emph{role2vec} (Figure~\ref{fig:PCA_Graph}.m). PCA projections that represented high expressivity were computed from encoded vectors of \emph{DeepWalk}, \emph{diff2vec}, \emph{GLEE}, \emph{GraRep}, \emph{Hope}, \emph{Laplacian Eigenmaps}, \emph{NetMF}, \emph{NMF-ADMM}, \emph{node2vec}, NodeSketch and \emph{Walklets}, respectively, Figure~\ref{fig:PCA_Graph}.b, Figure~\ref{fig:PCA_Graph}.c, Figure~\ref{fig:PCA_Graph}.d, Figure~\ref{fig:PCA_Graph}.f, Figure~\ref{fig:PCA_Graph}.g, Figure~\ref{fig:PCA_Graph}.h, Figure~\ref{fig:PCA_Graph}.i, Figure~\ref{fig:PCA_Graph}.j, Figure~\ref{fig:PCA_Graph}.k, Figure~\ref{fig:PCA_Graph}.l, and Figure~\ref{fig:PCA_Graph}.n. Very high expressivity was observed using  \emph{GraphWave} (Figure~\ref{fig:PCA_Graph}.e), where it is possible to observe an organized gradient by scenario complexity, all different event logs are identified spread in the 2D projection. 
This result is confirmed by the measurements obtained with the T4 measure.

\begin{figure}[ht!]
    \begin{tabular}{cccc}
        \includegraphics[width = 5.0in]{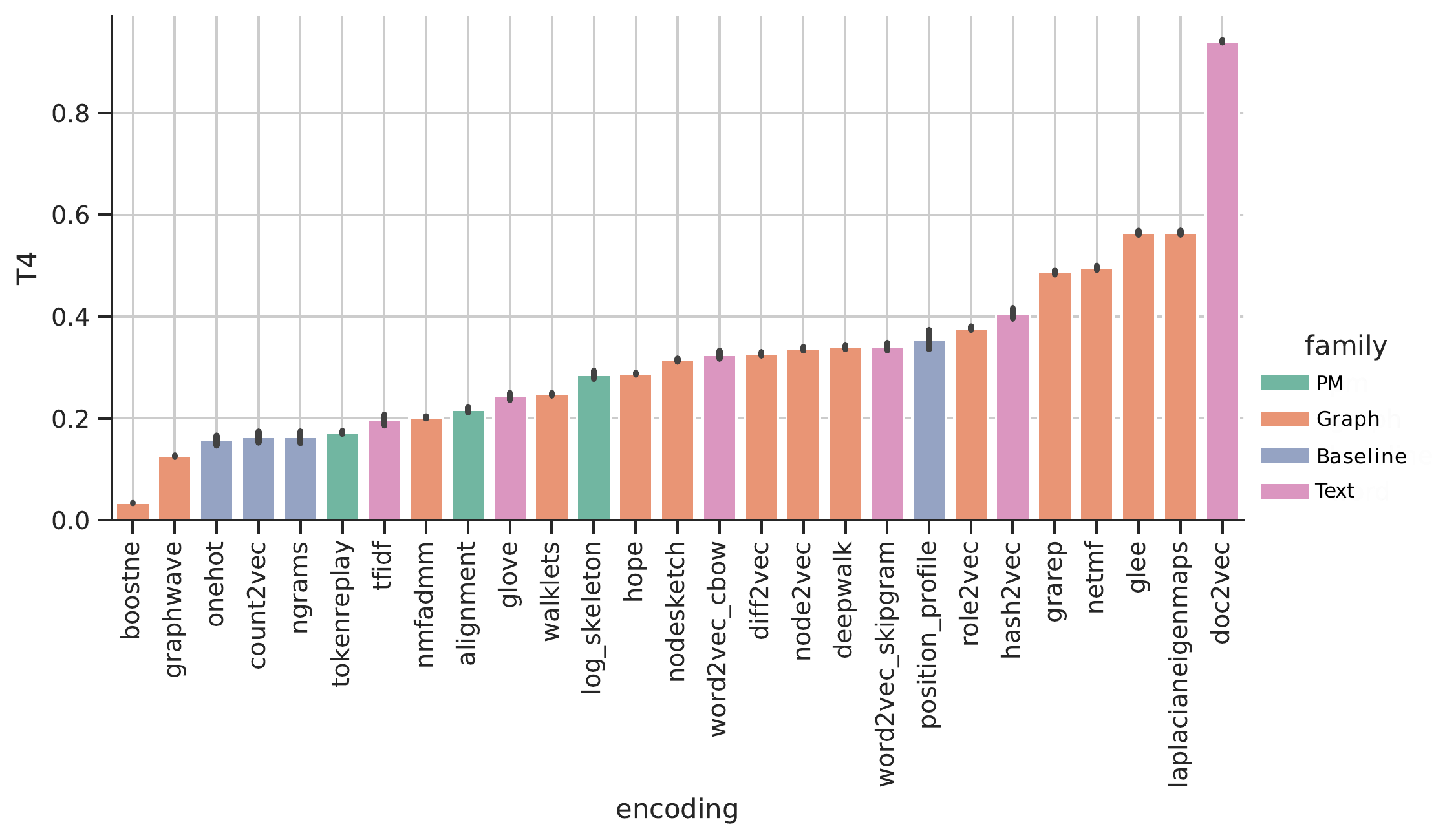}
     \end{tabular}
    \caption{T4 value obtained from encoding methods for expressivity evaluation. Low T4 values are correlated to good expressivity and high T4 values indicate poor expressivity.}
    \label{fig:T4}
\end{figure}

T4 gives a rough measure, from 0 to 1, of the proportion of relevant dimensions used by the encoding vector to map the event log \cite{barbon2020evaluating}. Relevance is determined according to the PCA criterion, which strives to describe most of the variability in the data with uncorrelated linear functions of the features \cite{lorena2019complex}. A higher T4 value indicates a more complex relationship between the input variables, indicating a larger number of original features are required to describe the data variability. Graph-based methods obtained the best expressivity, followed by the Baseline family, as shown in Figure~\ref{fig:T4}. In terms of T4 value, \emph{doc2vec} reached the lowest expressivity level, 0.92.

\subsection{Scalability}

\begin{figure}[ht!]
\centering
\begin{tabular}{ccc}
   \includegraphics[width = \textwidth]{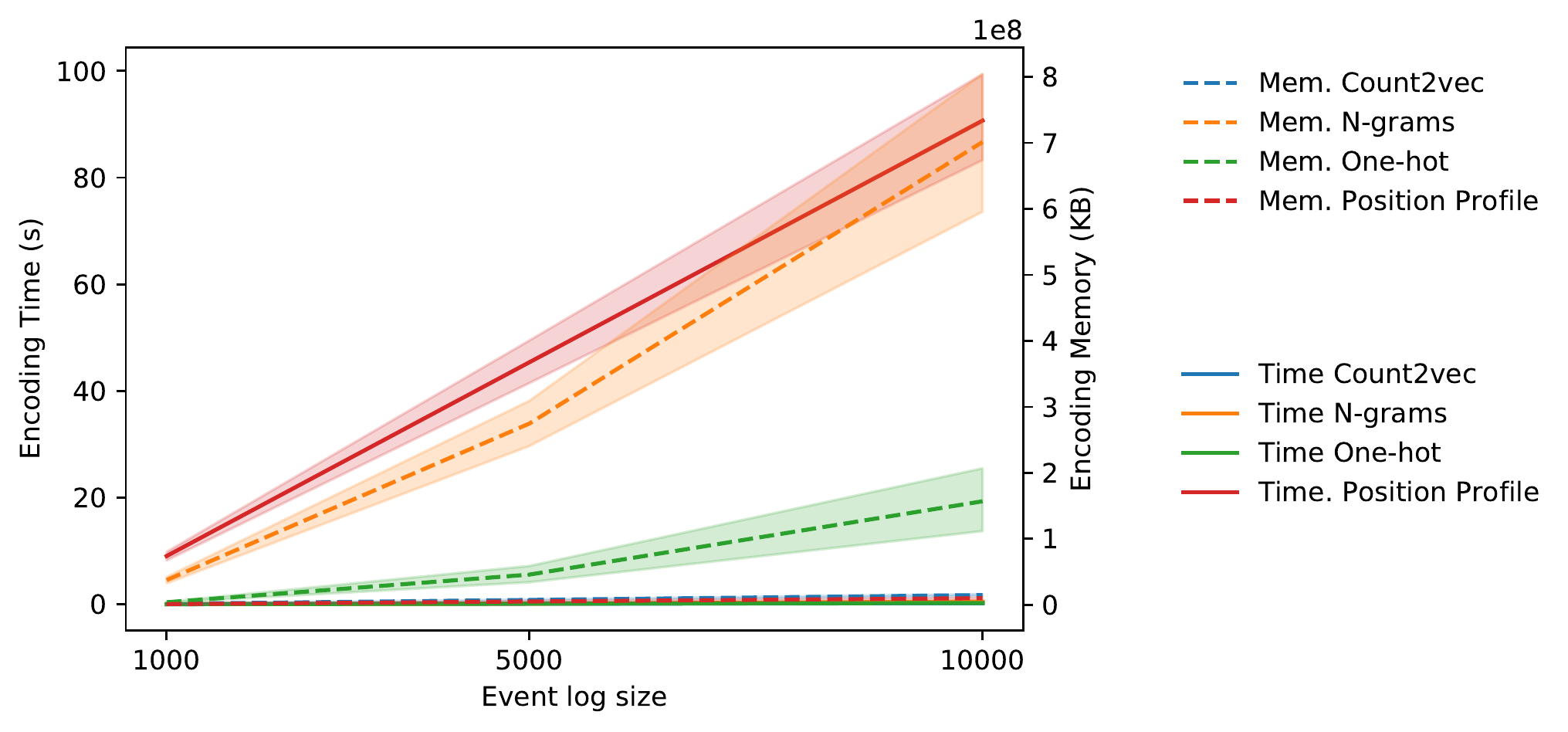}
\end{tabular}
\caption{Time and memory costs across different event log sizes (1k, 5k and 10k) when using Baseline encoding family.}
\label{fig:scal_baseline}
\end{figure}

\begin{figure}[ht!]
\centering
\begin{tabular}{ccc}
   \includegraphics[width = \textwidth]{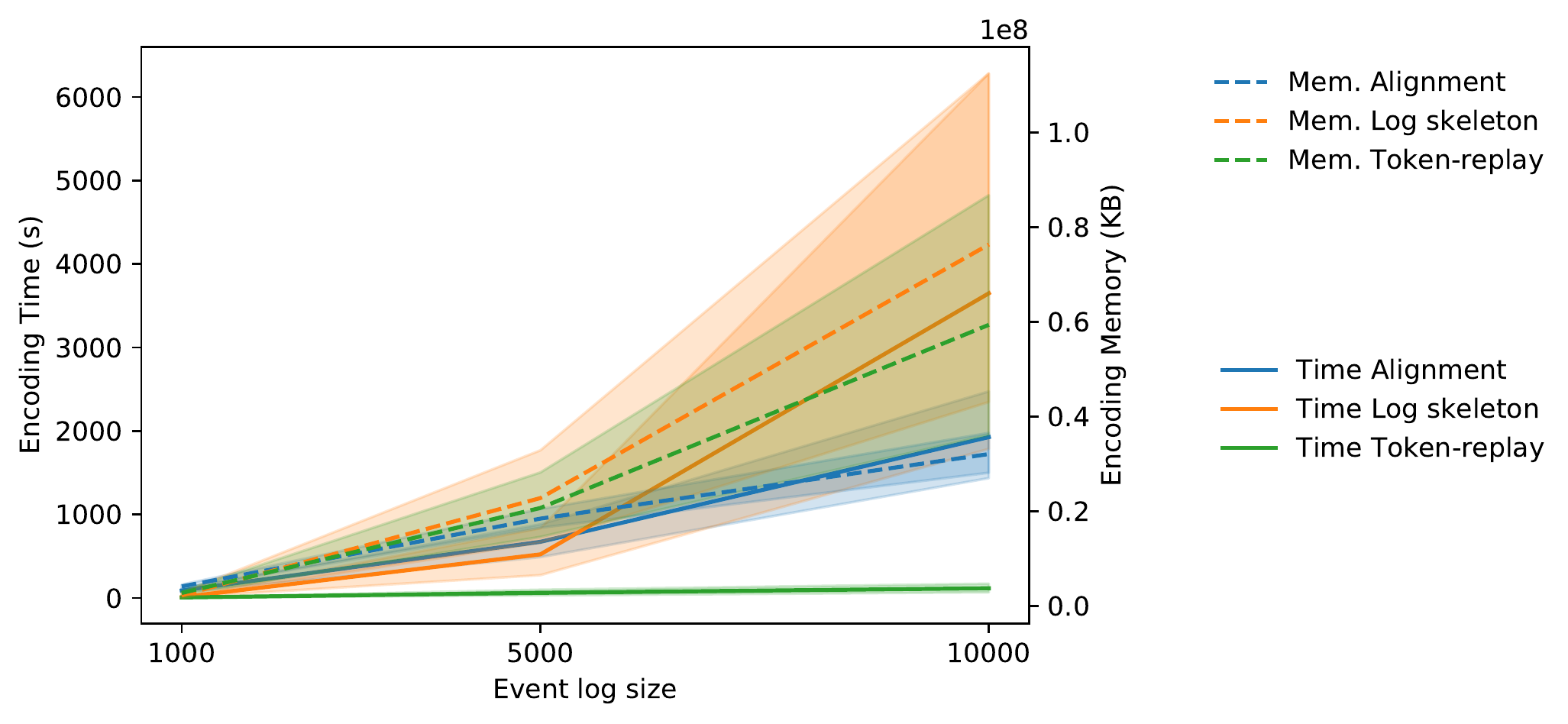}
\end{tabular}
\caption{Time and memory costs across different event log sizes (1k, 5k and 10k) when using Process Mining encoding family.}
\label{fig:scal_pm}
\end{figure}

\begin{figure}[ht!]
\centering
\begin{tabular}{ccc}
   \includegraphics[width = \textwidth]{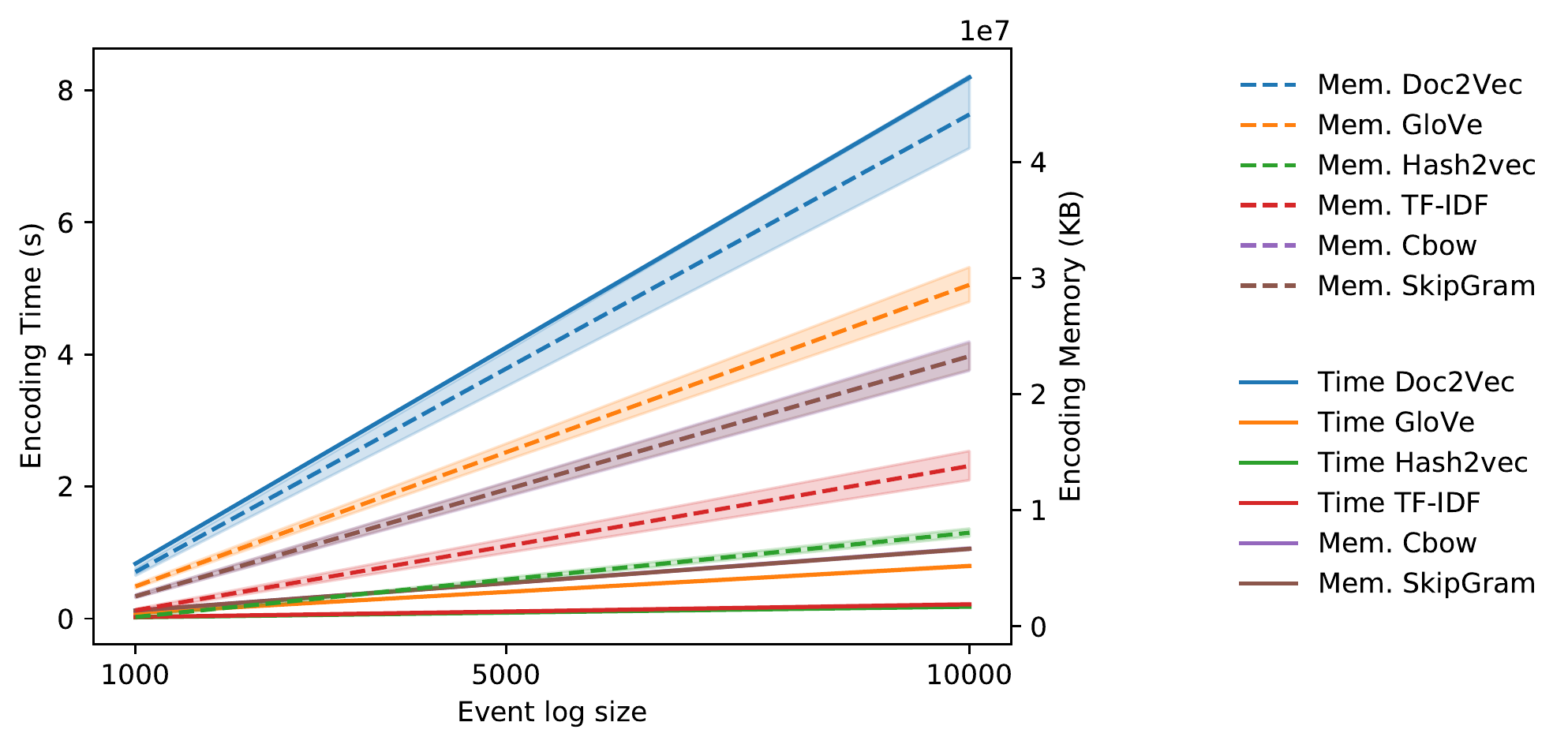}
\end{tabular}
\caption{Time and memory costs across different event log sizes (1k, 5k and 10k) when using Text encoding family.}
\label{fig:scal_text}
\end{figure}

\begin{figure}[ht!]
\centering
\begin{tabular}{ccc}
   \includegraphics[width = \textwidth]{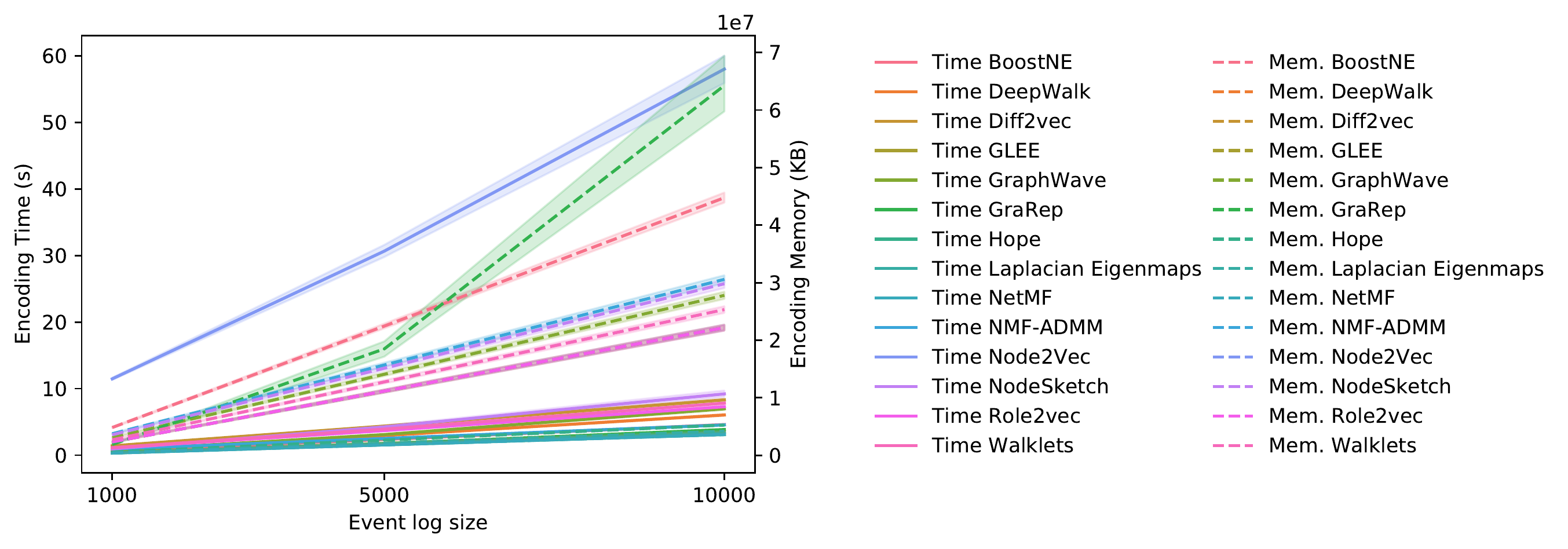}
\end{tabular}
\caption{Time and memory costs across different event log sizes (1k, 5k and 10k) when using Graph encoding family.}
\label{fig:scal_graph}
\end{figure}

We drive the discussion of scalability considering the time (seconds) and memory (KB) consumption accumulated during the whole encoding. Since costly methods are prohibitive to real-life event logs with huge volumes of data, their time and memory costs can directly influence the choice of an encoding method. In our experiments, we considered the time and memory consumed only during the encoding task. By comparing three different versions of event log size (1k, 5k, and 10k), we can observe how the costs are affected when encoding the same group of problems using different methods. Figures \ref{fig:scal_baseline}, \ref{fig:scal_pm}, \ref{fig:scal_text} and \ref{fig:scal_graph} were used to demonstrate the scalability of time and memory, limiting the y-axis according to the higher observed value (time on left-side and memory on right-side) over all experiments of each encoding method.

Baseline family analyses are supported by Figure~\ref{fig:scal_baseline}. In terms of memory cost, the Baseline encoding family showed that \emph{n-grams} was the most expensive method with a high space complexity. On the other hand, \emph{position profile} was the worst method in terms of time complexity. \emph{One-hot} demonstrated that memory costs increase more than time as the problem is scaled. In terms of scalability, \emph{count2vec} presented the best scalability results of the Baseline family. 

The PM encoding family presented high memory costs and average time costs, but with good scalability in both measures, as visible when the dataset increased and the performances slightly grew, as in Figure~\ref{fig:scal_pm}. \emph{Alignment} method was the cheapest one in terms of memory, \emph{token-replay} presented a good balance in terms of time, and \emph{Log skeleton} the most costly in both, memory and time. 


The Text encoding family presented the fastest methods with low memory consumption. However, presented time scalability issues by the majority of methods (\emph{GloVe}, \emph{hash2vec}, \emph{CBOW}, \emph{skip-gram} and \emph{doc2vec}), as represented by Figure \ref{fig:scal_text}. The least scalable was \emph{doc2vec}. A notable exception was \emph{TF-IDF}, which presented reduced memory usage even with increasing problem size. Note that the scalability was evaluated considering the ability to not suffer from data growth, and the average usage of time and memory in the Text encoding family is the lowest. 

The Graph encoding family presented a heterogeneous usage of memory and an average time cost, as Figure \ref{fig:scal_graph} shows. \emph{GLEE}, \emph{Hope} and \emph{NetMF} were the fastest methods of this family. These methods presented average scalability. The best scalability was demonstrated by \emph{NodeSketch}. The higher memory cost from the graph encoding family was achieved by \emph{GraRep} with a cost comparable to \emph{token-replay} (PM family), but with average scalability regarding time. The slowest method was \emph{node2vec}, using the smallest event log it was presented 4 times the average of the other methods of the same family. When dealing with the larger event log the time difference reached 7 times the other methods.

\begin{figure}[ht!]
     \centering
     \includegraphics[width=1.1\textwidth]{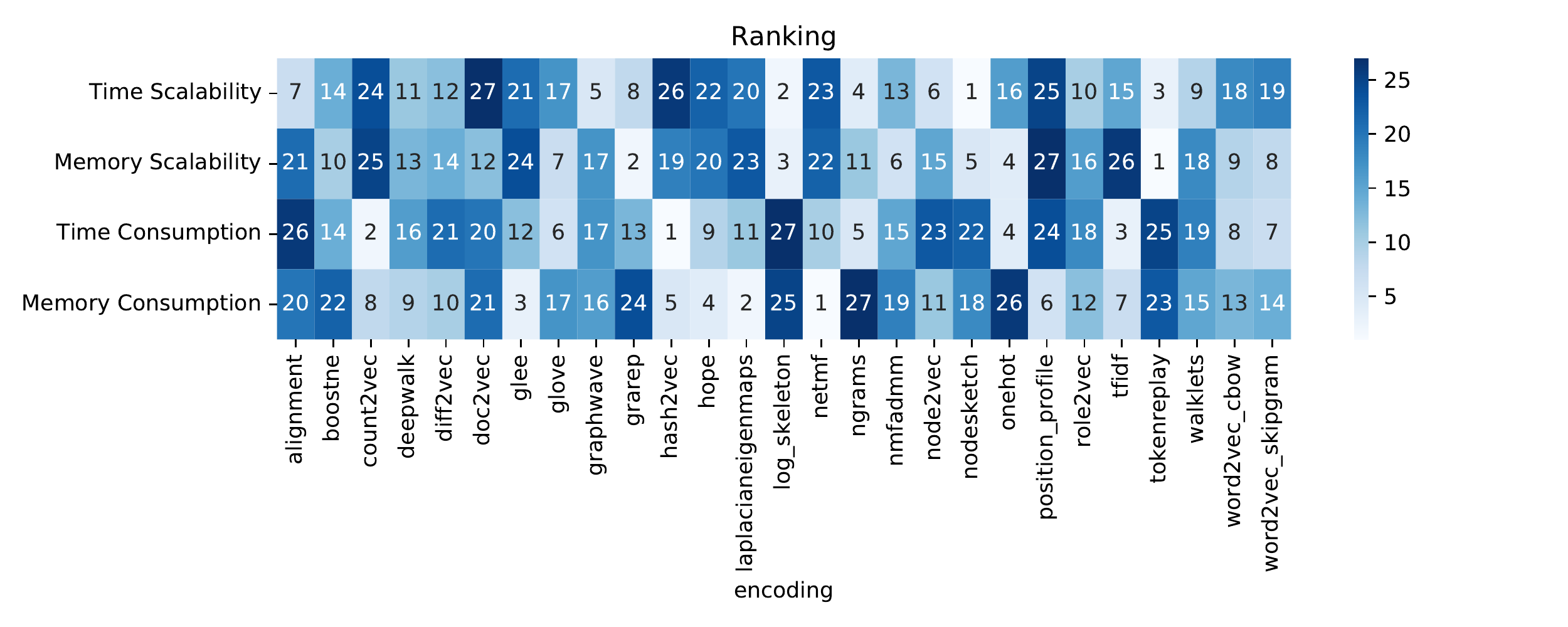}
     \caption{Ranking of Time Scalability, Memory Scalability, Time Consumption, and Memory Consumption. The better approaches are positioned in the first ranking position, colored by white. The most costly and less scalable are the last potions with dark colors.}
     \label{fig:sca_rank}
 \end{figure}

We organized the results of scalability and consumption of both time and memory analysis as a heat map (Figure~\ref{fig:sca_rank}). In the figure, it is possible to observe that general low memory and little time consumption \emph{count2vec}, do not reflects the scalability, i.e., increasing event log sizes, some methods compromise their costs with quadratic complexity costs of time and memory. Alternatively, \emph{Log skeleton}, \emph{token-replay}, and \emph{NodeSketch} are very scalable but have a high memory and time cost.

\subsection{Correlation power}

Correlation is an important analysis perspective since it reflects how correlated an encoding method is to the executed task in terms of performance. In other words, how the encoding positively contributed to the final performance. In our benchmark, we evaluated the correlation based on an anomalous trace detection task. In particular, we evaluated the F1-score obtained to detect anomalous behavior and the N2 from the mapped space. N2 is a ratio that computes distances between an example and its closest neighbor within a particular class (intra-class) and between an example and its closest neighbor from a different class (extra-class). The N2 value, which ranges from 0 to 1, is low when there is a greater distance between examples of different classes than between examples from the same class. Thus, a mapped space with a lower N2 refers to a representation that is better equipped to distinguish classes and support supervised learning. In our paper, we followed the N2 calculation as described by \cite{lorena2019complex}.

\begin{figure}[ht!]
     \centering
     \includegraphics[width=1.1\textwidth]{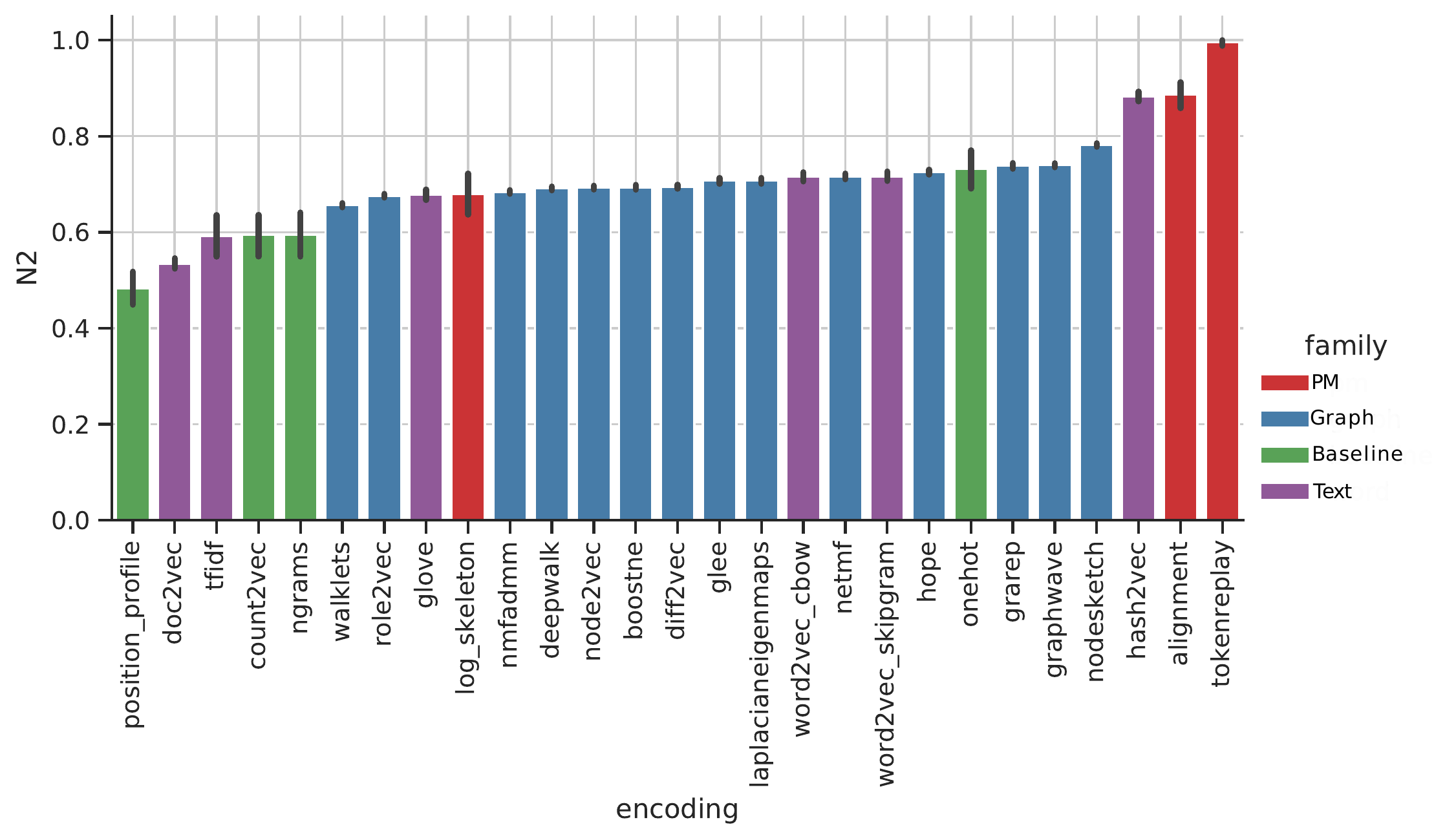}
     \caption{Ratio of Intra/extra class nearest neighbor distance (N2), provided by the same group of tasks considering 27 different encoding methods. The bars are sorted from the best score (left) to the worst one (right). Each bar is colored according to the coding family.}
     \label{fig:n2}
 \end{figure}

Figure \ref{fig:n2} represents the obtained N2 values from encoded space from all encoding methods. The figure presents each family with a particular color and results are sorted by N2, from the best one to the worst N2 value. \emph{Position profile} achieved the best N2 values, mean of 0.48, and was the only method with less than 0.50 in terms of this measure. The top 5 N2 values were obtained by Graph and Text families. The PM encoding family, particularly \emph{alignment} and \emph{token-replay} reached high N2 values, superior to 0.85. In contrast, \emph{Log skeleton} obtained less than 0.70, ranking in the bottom 10 encoding methods, but supported the best F1-score in the anomaly detection task.

\begin{figure}[ht!]
     \centering
     \includegraphics[width=1.1\textwidth]{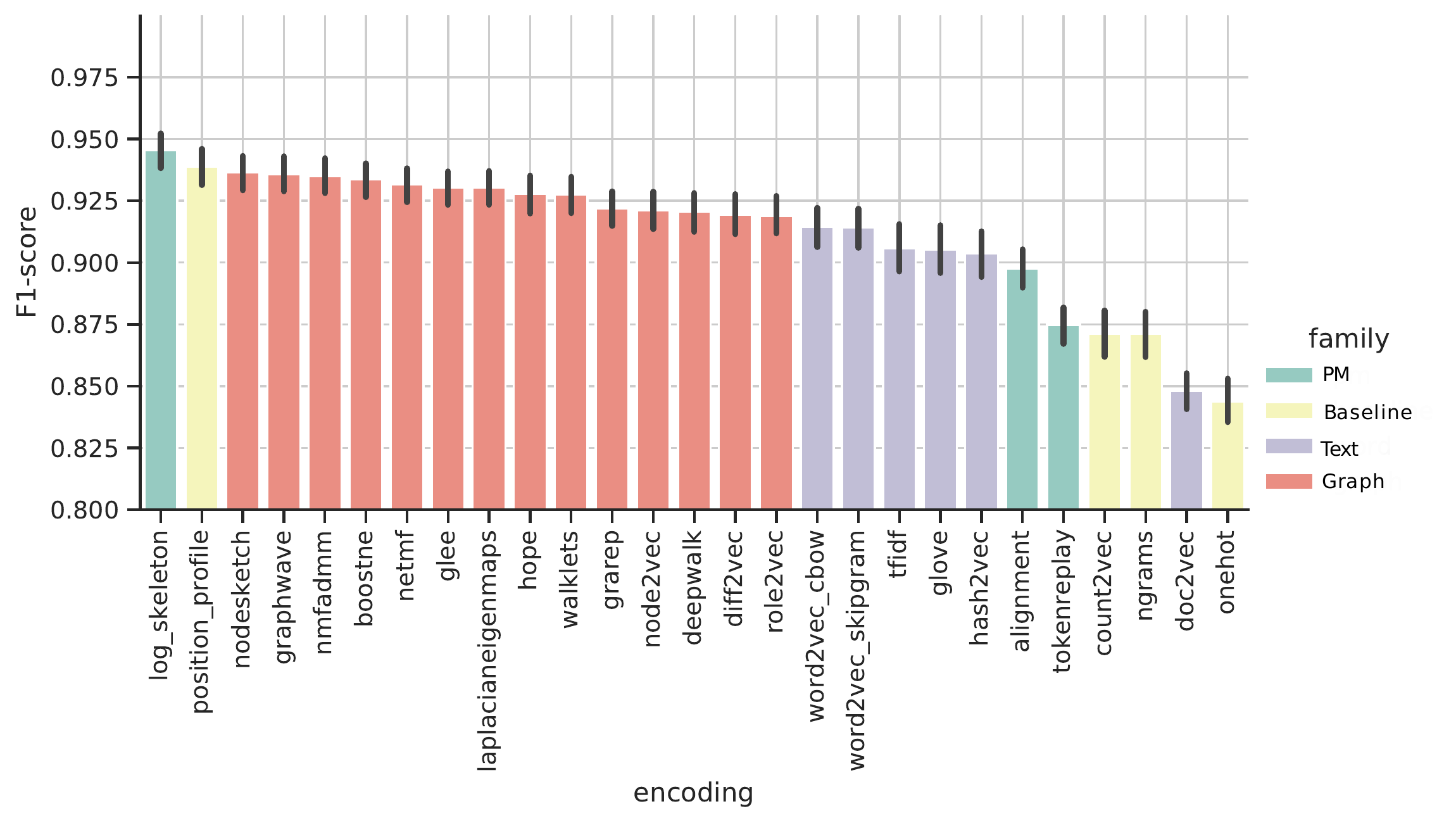}
     \caption{F1-score obtained by several classification tasks performed using a Random Forest algorithm, considering 27 different encoding tasks. The bars are sorted from the best score (left) to the worst one (right). Each bar is colored according to the coding family.}
     \label{fig:f1}
 \end{figure}

When evaluating the F1-score, \emph{Log skeleton} provided the best performance (mean of 0.942), followed by \emph{position profile} (mean of 0.935) and all encoding from the Graph encoding family (above 0.915). Text encoding family provided results above 0.903, except \emph{doc2vec} that lead to an average F1-score of about 0.845. Surprisingly, the most traditional encoding used in PM, \emph{one-hot}, obtained the worst results, i.e., inferior to 0.840 of the F1-score. The obtained F1-scores are sorted by the performance from the best to the worst one in Figure~\ref{fig:f1}.

\begin{figure}[ht!]
     \centering
     \includegraphics[trim=80 0 0 0 ,clip,width=1.1\textwidth]{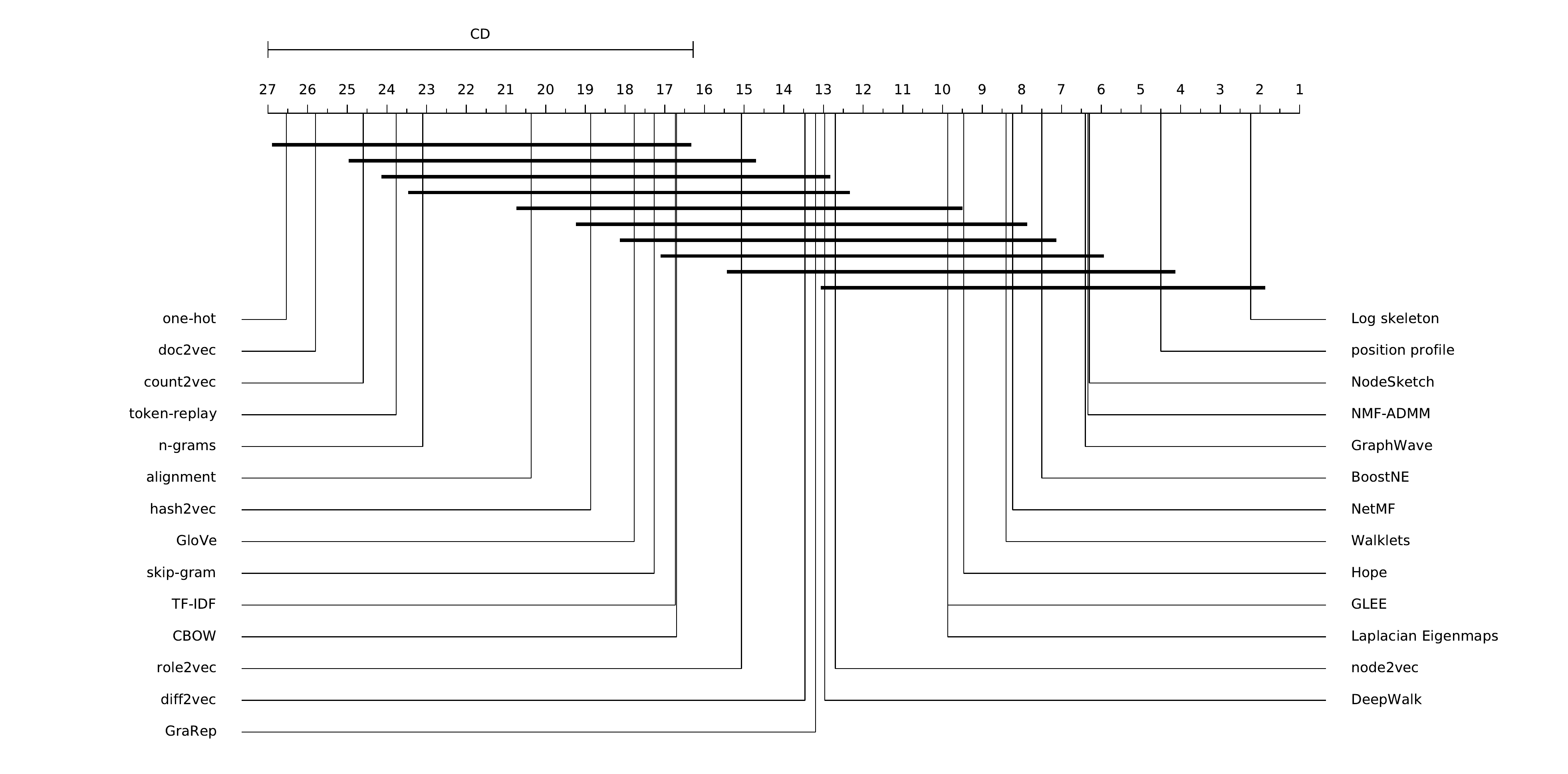}
     \caption{Nemenyi post-hoc test (significance of $\alpha=0.05$ and critical distance of 10.71)  considering the \textbf{F1-score} (predictive performance) accuracy obtained when performing the classification task using all encoding methods over 10 scenarios (1k, 5k and 10k traces from five different scenarios). Statically similar methods are linked by the solid line.}
     \label{fig:nemenyi}
 \end{figure}

In order to provide a fair and statistically grounded comparison in terms of predictive performance, we used Friedman's statistical test and the post-hoc test of Nemenyi. Both tests were employed to verify the statistically significant difference between the performance of the F1-score of each encoding method. The result of the statistical comparison can be observed as a Critical Difference chart, as illustrated in Figure~\ref{fig:nemenyi}. Critical Difference test allows checking when there were statistical differences between the segmenters, each diagram and method's average ranks are placed on the horizontal axis, with the best ranked to the right. The solid line connects encoding methods with no significant performance difference. Thus, Figure~\ref{fig:nemenyi} demonstrates no statistical difference among \emph{Log skeleton}, \emph{position profile}, \emph{NodeSketch}, \emph{NMF-ADMM},  \emph{GraphWave}, \emph{BoostNE}, \emph{NetMF}, \emph{Walklets}, \emph{Hope}, \emph{GLEE},  \emph{Laplacian Eigenmaps}, \emph{node2vec}, all of them supporting high predictive performance. On the other hand, there is no statistical difference as encoding methods that provided lesser predictive models for \emph{one-hot}, \emph{doc2vec}, \emph{n-grams}, \emph{count2vec}, \emph{token-replay}, \emph{alignment}, \emph{hash2vec}, \emph{GloVe}, \emph{skip-gram}, \emph{CBOW} and \emph{TF-IDF}.

\subsection{Domain agnosticism}

The last comparison criterion regards Domain Agnosticism. We consider this criterion as important as Expressivity, Scalability, and Correlation to comprehend the obtained results. Since a method valid for multiple applications is instrumental to the construction of adaptive data science pipelines.

The methods comprising the baseline family are traditionally used in different PM tasks. They realize straightforward transformations mapping traces into feature vectors. Their usage was not limited to PM pipelines, indeed, they were created in other data mining areas. For these reasons, we consider the Baseline family as domain agnostic. As well, the Text and Graph families have been used for a wide range of purposes. These last families are good examples of domain agnosticism and have been used with simple encoding scenarios as well as with complex and highly structured representations.

Considering our criteria, PM-based solutions are not domain agnostic. This family of encoding methods was conceived to represent event logs and has been used exclusively for this purpose. It should not be viewed as a limitation, but rather as a characteristic of specialized methods demonstrating a high correlation power, even at the cost of high computational costs, e.g., \emph{Log skeleton} provided encoded spaces to induce models that obtained high F1-scores.

\begin{figure}[ht!]
             \centering
             \includegraphics[width=1.1\textwidth]{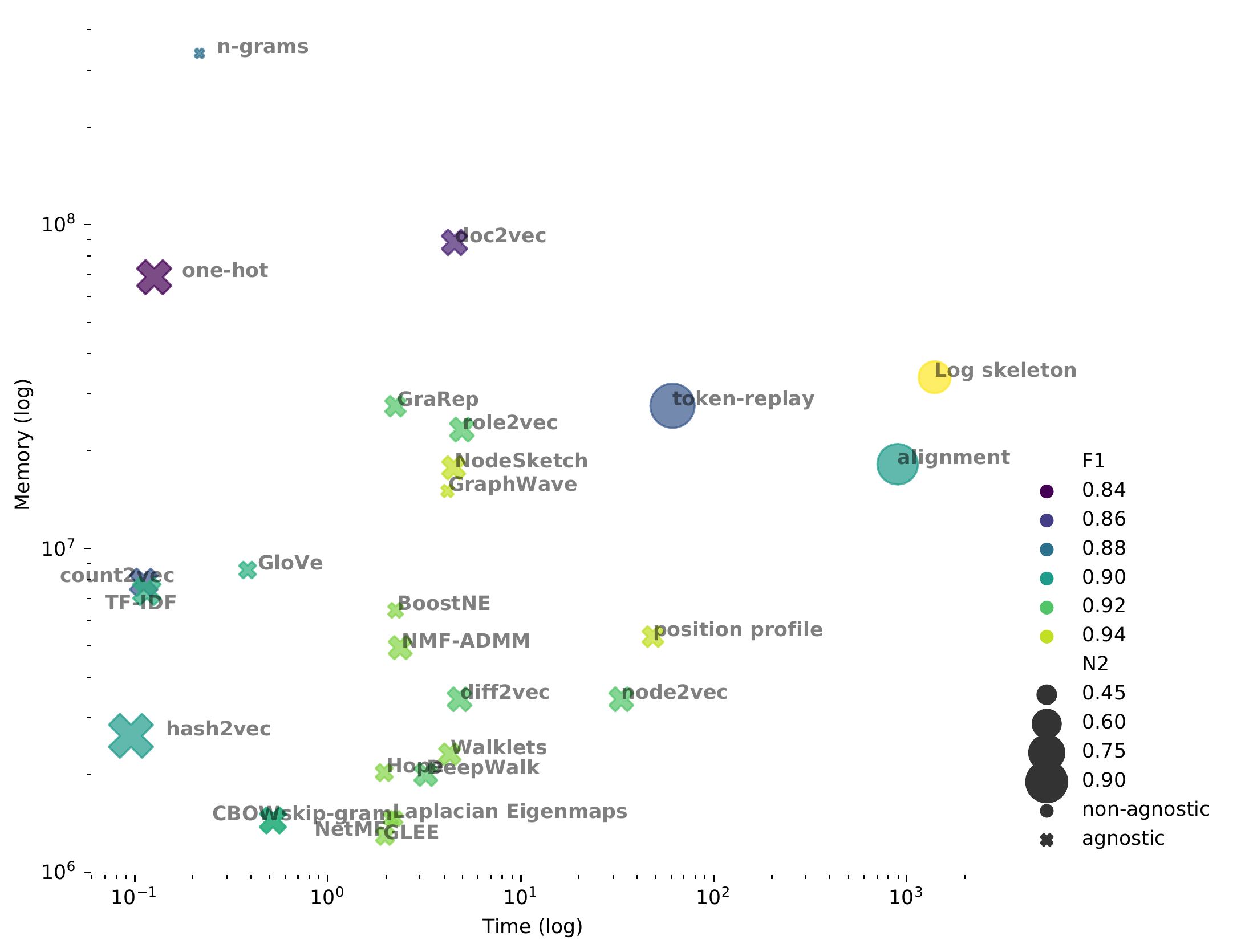}
             \caption{Mean values of time, memory, and predictive performance (F1) projected by time and memory in a logarithmic scale. The marker size represents the N2 and their color gradient performance in terms of F1. The marker symbol represents the domain agnosticism evaluation.}
             \label{fig:RQ3}
\end{figure}

An overview of domain agnosticism and some implications of encoding method performance and resource consumption could be observed in Figure~\ref{fig:RQ3}. The most notable observation is that non-agnostic methods (\emph{token-replay}, \emph{alignment}, and \emph{Log skeleton}) share high N2 values and high costs of space and time complexity.
Among the agnostic methods, the best N2 and F1 performances are obtained with average space and time complexity, e,g, with \emph{GraphWave} or \emph{BoostNE}.

\section{Issues, Concerns and Future Directions}\label{sec:discussion}
    
Research comparisons about encoding methods focused on PM are still embryonic. After classifying the papers in Section \ref{sec:encoding_methods}, we observed that the proposed solutions do not investigate the strong and weak points of each method. Thus, we have proposed a study involving a large set of methods from different families. 

Investigating the pros and cons of encoding methods based on \textit{expressivity}, \textit{scalability}, \textit{correlation power}, and \textit{domain agnosticism} over different encoding families and hundreds of event logs with various complexities, we were able to gain insights and share some assumptions for future directions and possible novel PM encoding methods. We believe the criteria employed in this work to assess the effectiveness are the most important aspects to take into consideration in order to achieve significant accomplishments for any PM task that needs to encode event logs.

Moreover, the proposed metrics can also serve as user requirements when deciding which encoding method should be employed for the specific problem. 
The \textit{expressivity} of an encoding method can measure how straightforward the representation of an event log is according to its complexity.
The \textit{scalability} is also an important concern since real-life event logs consist of a large volume of data which can lead to high computational costs regarding elapsed time and memory usage.
Understanding the \textit{correlation power} between the nature of an encoding method and the performance of the executed task is also a relevant analysis that allows practitioners to estimate the complexity of analyzing a specific event log.
Lastly, the \textit{domain agnosticism} is a novel and important discussion introduced in this work to consider if encoding methods can be adapted for different problem domains.

Addressing further issues on encoding methods in PM, online PM may introduce new challenges for the current encoding methods. As emphasized by \cite{ceravolo2020evaluation}, measures such as accuracy and memory consumption need to drive the creation of methods to match online PM goals, as the encoding methods used in such solutions. The concern of limitations posed by an online PM task was also highlighted by \cite{tavares2019overlapping}, mentioning also the demand of adapting when dealing with concept drifts and focusing on inter-activity time implications. STARDUST \cite{pasquadisceglie2022stardust} is an example of online PM, particularly on trace streams. The authors discussed issues regarding approaches to handling traces recorded without the final activity. It is therefore essential to investigate encoding methods that support online PM tasks coping with new challenges, such as reduced memory consumption and the ability to map partial traces. In our experiments, we used \textit{scalability} measure to support insights and discussions regarding this topic. A crucial indicator of scalability was the encoding families.

Currently, the predictive process monitoring problem also faces this issue regarding the lack of encoding methods specific to PM.
Representation learning or feature learning is a learning paradigm that has been recently introduced in the community by~\cite{KoninckBW18}. 
We believe this is a promising path for improving the encoding procedure in process mining tasks.
In the mentioned work, the authors derived their new proposals from the \emph{word2vec}.
However, we believe PM requires a specialized method or a sufficiently generic one regardless of the problem domain since the event data might be considered more complex than sequential text.
We claim that since the nature of event data, in general, contains sequential rules, relational information, concurrency of resources, parallel activities, etc.

Encoding regards mapping data into another representation for different goals. The new space could not allow interpretations and explainability as previously supported by the original data. Moreover, practitioners need to trust the generated mapped space, as mentioned by \cite{elkhawaga2022explainability}. In particular, the predictive model presented in a great part of predictive monitoring tasks does not explain why it provided wrong predictions, so the reason why a prediction model made a mistake cannot be understood. Shedding some light on this topic, \cite{maggiexplainability} presented post-hoc explainers and different encoding methods for identifying the important features. On a general note, our experiments using \textit{expressivity} and \textit{domain agnosticism} confirmed the wide range of representations provided by different encoding methods, even from the same family. Regarding this point of eXplainable Artificial Intelligence (XAI), as a tendency, encoding methods able to provide high explainability levels should be integrated into the PM pipeline as a key component, not a separate step follow-up effort for particular applications.

The great number of encoding methods and reduced availability of experts pose an additional challenge to selecting and properly setting the encoding method. Strategies focused on accuracy or time performance are applied when selecting a method, but the cost of testing different setups and costly tuning strategies could impact the PM pipeline conception. This problem has been addressed by promising strategies based on meta-learning \cite{tavares2021process,tavares2022selecting,TavaresBD22}, but the current solutions require creating a meta-database containing the history of possible solutions. Also, the criteria to recommend a particular algorithm are still limited to simple performance functions. Therefore, Automatic Machine Learning (AutoML)~\cite{OlsonM16,FeurerKES0H19} proves to be an important research area that impacts the aforementioned concerns regarding encoding methods used in PM tasks.
Alternatively, another learning paradigm that could be explored for this nature of data is self-supervised learning (SSL).
The general idea behind SSL is learning a set of possible outcomes given an input, instead of predicting a unique value as traditional methods.
For instance, the \emph{data2vec} was recently presented by \cite{BaevskiHXBGA22}, where the authors propose a generic framework for encoding any type of data, although only the domains of image, speech, and language have been considered.
Intuitively, this might be interesting for capturing mutual dependencies in event data.

\section{Conclusion}\label{sec:conclusion}

The main contributions presented in this work include a systematic review of process mining tasks using encoding methods, a new taxonomy to categorize each type of method into families, and an extensive experimental evaluation and benchmark assessing relevant evaluation metrics to measure the effectiveness of an encoding method.
We believe this work can support researchers and practitioners to achieve significant accomplishments in different application areas in PM.
Furthermore, we stress current challenges and issues in the literature regarding the difficulty of choosing the right algorithm and its parameters.
We also discuss how arbitrarily selecting algorithms lead to unfair evaluation and sub-optimal solutions. 
This is the first work that focuses on a detailed analysis for preprocessing event logs instead of focusing only on the task algorithm itself (i.e. a clustering or learning algorithm).

We also highlighted the need for a better understanding of how each method behaves according to different scenarios of event logs and different PM tasks.
Thus, to fill this gap we simulated such scenarios by employing the PLG2 tool to generate synthetic processes with distinct properties and presented the results as a benchmark.
In total, 27 encoding methods were evaluated throughout 420 different event logs containing different anomalies.
We considered four different evaluation criteria to measure the effectiveness of encoding methods for process mining tasks.
We limited our evaluation to only one task, anomaly detection, but the analysis and insights presented in this work can be leveraged for other applications, such as predictive monitoring and clustering.

We conclude this work by stressing the difficulty of choosing suitable algorithms and their parameters according to the user's preferences since each pipeline setting performs differently according to the event log characteristics.
This might be a direction to novel automated solutions whether for entire pipelines or preprocessing steps only. We also believe that an encoding method that specifically handles the nature of processes is essential for advancing the state-of-the-art.
This is claimed by considering that most methods are adapted or adopted from other areas. Therefore, this research line is promising and has several opportunities for work to be developed.

\bibliographystyle{unsrt}  
\bibliography{ms}

\end{document}